\documentclass{article}
\usepackage{amsmath,amsthm,amssymb}
\usepackage{enumitem}
\usepackage{microtype}
\usepackage{graphicx}
\usepackage{subcaption}
\usepackage{booktabs} 
\usepackage{amsmath,amsthm,amssymb}
\usepackage{mathtools}
\usepackage{enumitem}
\usepackage{newtxtext} 
\usepackage[smallerops]{newtxmath}
\usepackage{multirow}
\usepackage[linesnumbered,ruled]{algorithm2e}
\usepackage{xcolor}
\usepackage{wrapfig}

\usepackage[draft,inline,nomargin]{fixme}
\usepackage[round]{natbib}

\newcommand{\minimize}{\mathop{\mathrm{minimize}{}}}

\newcommand{\R}{\mathbf{R}}
\newcommand{\E}{\mathbf{E}}

\renewcommand{\P}{\mathbf{P}}
\newcommand{\Var}{\mathbf{Var}}

\newcommand{\dk}{d^k}

\newcommand{\dkp}{d^{k+1}}
\newcommand{\gk}{g^k}

\newcommand{\gkp}{g^{k+1}}

\newcommand{\xk}{x^k}

\newcommand{\xkp}{x^{k+1}}

\newcommand{\xik}{\xi^k}

\newcommand{\tr}{\mathbf{tr}}
\newcommand{\pz}[1]{\textcolor{red}{PZ: #1}}

\newcommand{\xtk}{\tilde{x}^k}

\newtheorem{theorem}{Theorem}

\usepackage[final]{style}

\usepackage[utf8]{inputenc} 
\usepackage[T1]{fontenc}    
\usepackage{hyperref}       
\usepackage{url}            
\usepackage{booktabs}       
\usepackage{amsfonts}       
\usepackage{nicefrac}       
\usepackage{microtype}      
\usepackage{newtxtext}
\usepackage[smallerops]{newtxmath}

\title{Using Statistics to Automate Stochastic Optimization}

\author{
  Hunter Lang\thanks{Work done while HL was at Microsoft Research AI.}\\
  \texttt{hunterlang1@gmail.com}
  \And
  Pengchuan Zhang\\
  Microsoft Research AI\\
  \texttt{penzhan@microsoft.com}
  \And
  Lin Xiao\\
  Microsoft Research AI\\
  \texttt{lin.xiao@microsoft.com}  
}

\let\oldmaketitle\maketitle
\renewcommand{\maketitle}{\oldmaketitle\setcounter{footnote}{0}}

\begin{document}

\maketitle

\begin{abstract}
  Despite the development of numerous adaptive optimizers, tuning the
  learning rate of stochastic gradient methods remains a major
  roadblock to obtaining good practical performance in machine learning.
  Rather than changing the learning rate at each iteration, we propose an approach
  that automates the most common hand-tuning heuristic: use a constant
  learning rate until ``progress stops'', then drop.
  We design an explicit statistical test that determines when the dynamics of stochastic
  gradient descent reach a stationary distribution. This
  test can be performed easily during training, and when it fires, we
  decrease the learning rate by a constant multiplicative factor.  Our
  experiments on several deep learning tasks demonstrate that this statistical
  adaptive stochastic approximation (SASA) method can
  automatically find good learning rate schedules and match the performance
  of hand-tuned methods using default settings of its parameters.
  The statistical testing helps to control the
  variance of this procedure and improves its robustness.

\end{abstract}

\section{Introduction}\label{sec:intro}
Stochastic approximation methods, including stochastic gradient descent (SGD)
and its many variants,
serve as the workhorses of machine learning with big data.
Many tasks in machine learning can be formulated as the stochastic
optimization problem:
\[
    \textstyle\minimize_{x\in\R^n} ~F(x) \triangleq \E_{\xi}\bigl[f(x,\xi)\bigr] ,
\]
where $\xi$ is a random variable representing data sampled from some
(unknown) probability distribution, $x\in\R^n$ represents the
parameters of the model (e.g., the weight matrices in a neural
network), and $f$ is a loss function.
In this paper, we focus on the following variant of SGD with \emph{momentum},
\begin{equation}\label{eqn:SGM}
\begin{split}
    \dkp  &= (1-\beta_k)\gk + \beta_k \dk, \\[-0.5ex]
    \xkp &= \xk - \alpha_k \dkp ,
\end{split}
\end{equation}
where $\gk=\nabla_x f(\xk,\xik)$ is a stochastic gradient,
$\alpha_k>0$ is the learning rate,
and $\beta_k\in[0,1)$ is the momentum coefficient.
This approach can be viewed as an extension of the heavy-ball method
  \citep{Polyak64heavyball} to the stochastic setting.\footnote{For fixed values of $\alpha$ and $\beta$, this ``normalized'' update formula is equivalent to the more common updates $\dkp = \gk + \beta\dk$, $\xkp = \xk - \alpha'\dkp$ with the reparametrization $\alpha = \alpha'/(1-\beta)$.} To
distinguish it from the classical SGD, we refer to the
method~\eqref{eqn:SGM} as SGM (Stochastic Gradient with Momentum).

Theoretical conditions on the convergence of stochastic approximation
methods are well established \citep[see, e.g.,][and references
  therein]{Wasan69book,KushnerYin03book}. Unfortunately, these
asymptotic conditions are insufficient in practice.
For example, the classical rule $\alpha_k=a/(k+b)^c$
where $a,b>0$ and $1/2< c\leq 1$, often gives poor performance even
when $a$, $b$, and $c$ are hand-tuned. Additionally, despite the advent of numerous adaptive variants of SGD and SGM \citep[e.g.,][and other variants]{DuchiHazanSinger2011adagrad,TielemanHinton2012lecture,KingmaBa2014adam},
achieving good performance in practice often still requires
considerable hand-tuning
\citep{Wilson2017marginal}.

\begin{wrapfigure}{r}{0.5\textwidth}
\begin{minipage}{0.495\textwidth}
\begin{algorithm}[H]
  \SetKwInput{Input}{Input}
  \Input{$\{x^0, \alpha_0, M, \beta, \zeta\}$}
  \For{$j \in \{0,1,\ldots\}$}{
    \For{$k \in \{jM,\ldots,(j+1)M-1\}$}
        {
          Sample $\xik$.\\
          Compute $\gk = \nabla_x f(\xk, \xik)$.\\
          $\dkp = (1-\beta)\gk + \beta\dk$\\
          $\xkp = \xk - \alpha\dkp$\\
          // collect statistics
        }
    \If{test(statistics)}
    {
      $\alpha \gets \zeta \alpha$\\
      // reset statistics
    }
  }
  \caption{General SASA method}
  \label{alg:const-and-cut}
\end{algorithm}
\end{minipage}
\end{wrapfigure}

\begin{figure}[t]
  \centering
  \begin{subfigure}[t]{.3\linewidth}
    \centering
    \includegraphics[width=\linewidth]{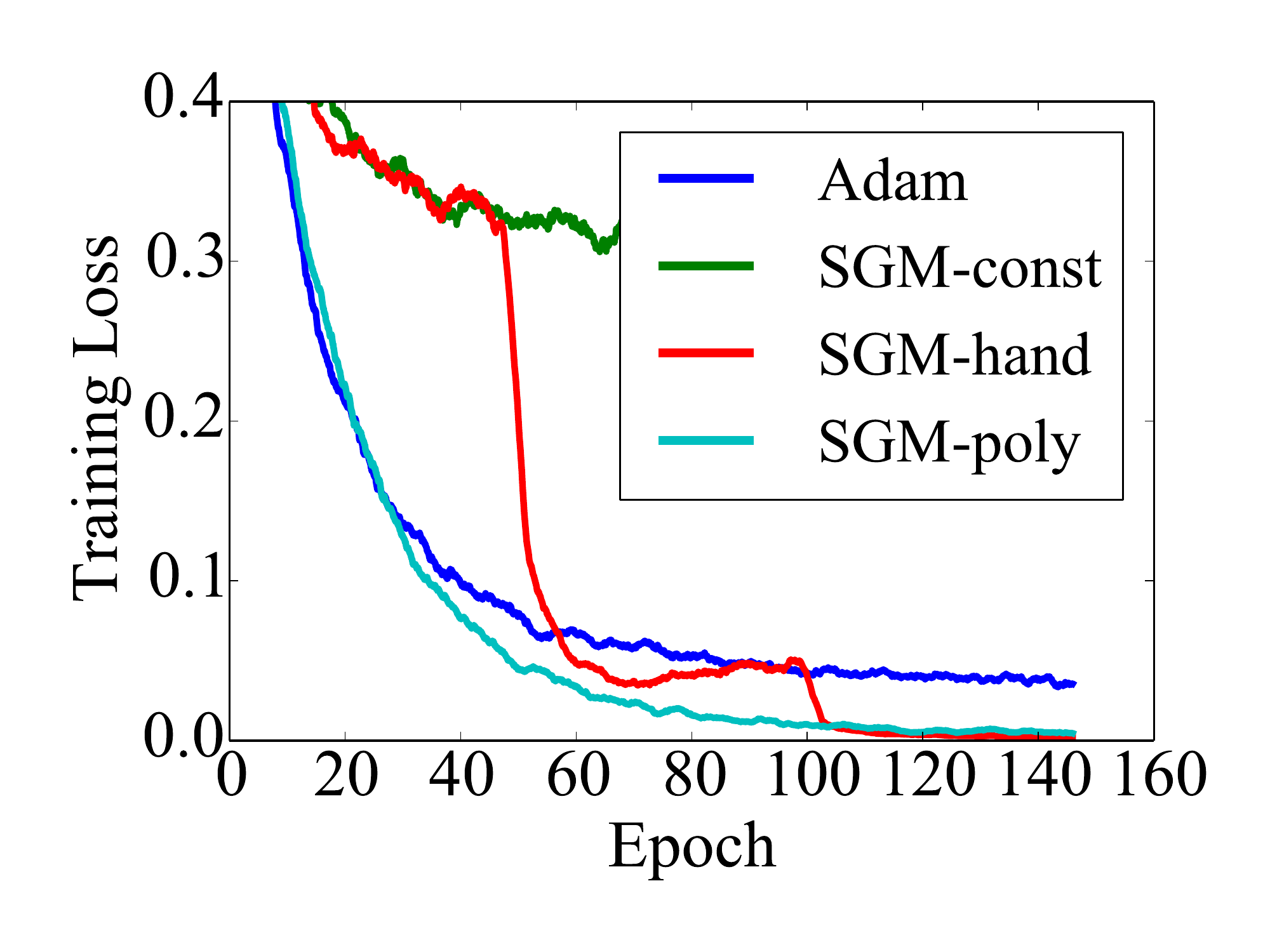}
    \subcaption{training loss}
  \end{subfigure}%
  \begin{subfigure}[t]{.3\linewidth}
    \centering
    \includegraphics[width=\linewidth]{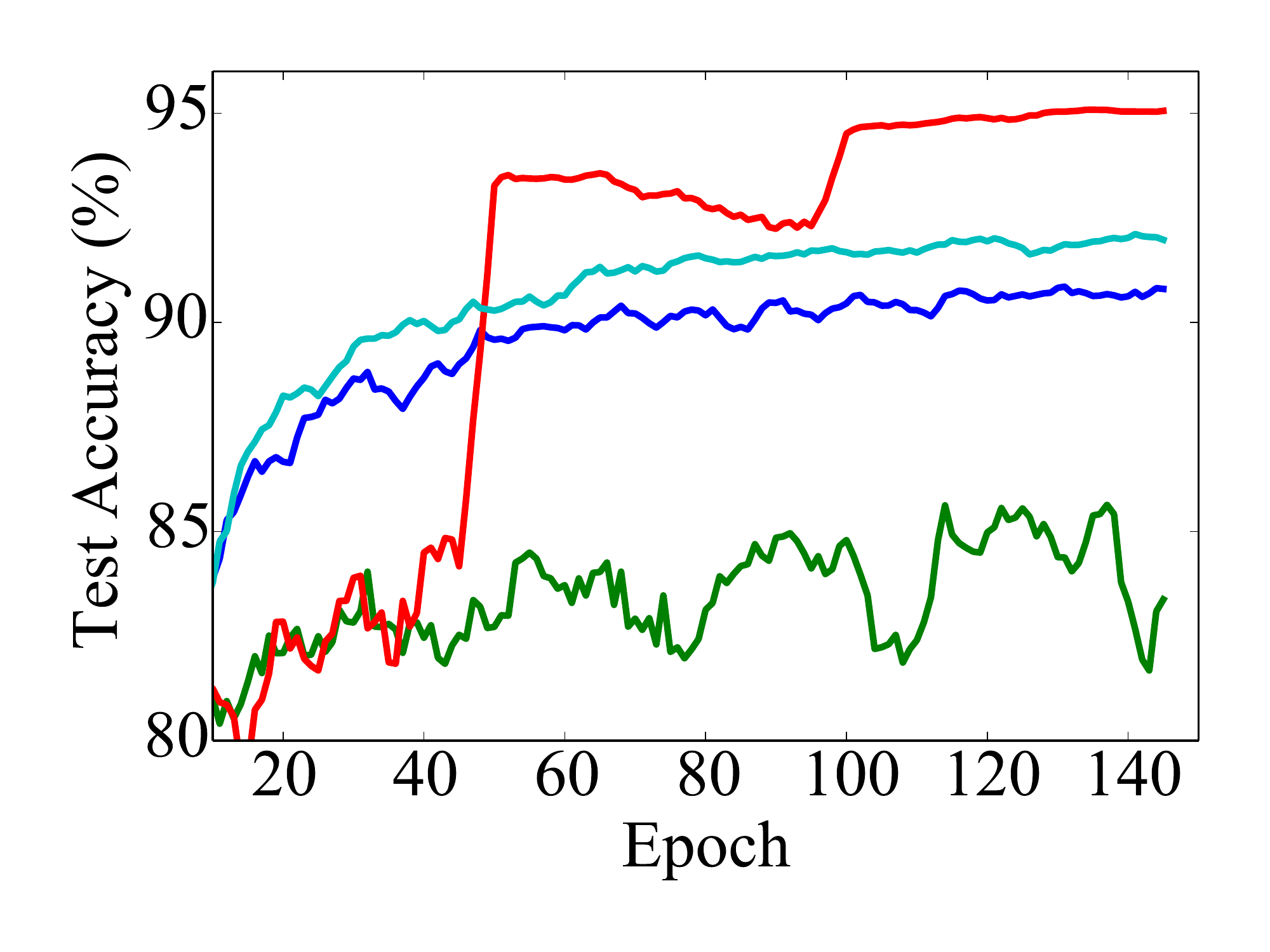}
    \subcaption{test accuracy}
  \end{subfigure}%
  \begin{subfigure}[t]{.3\linewidth}
    \centering
    \includegraphics[width=\linewidth]{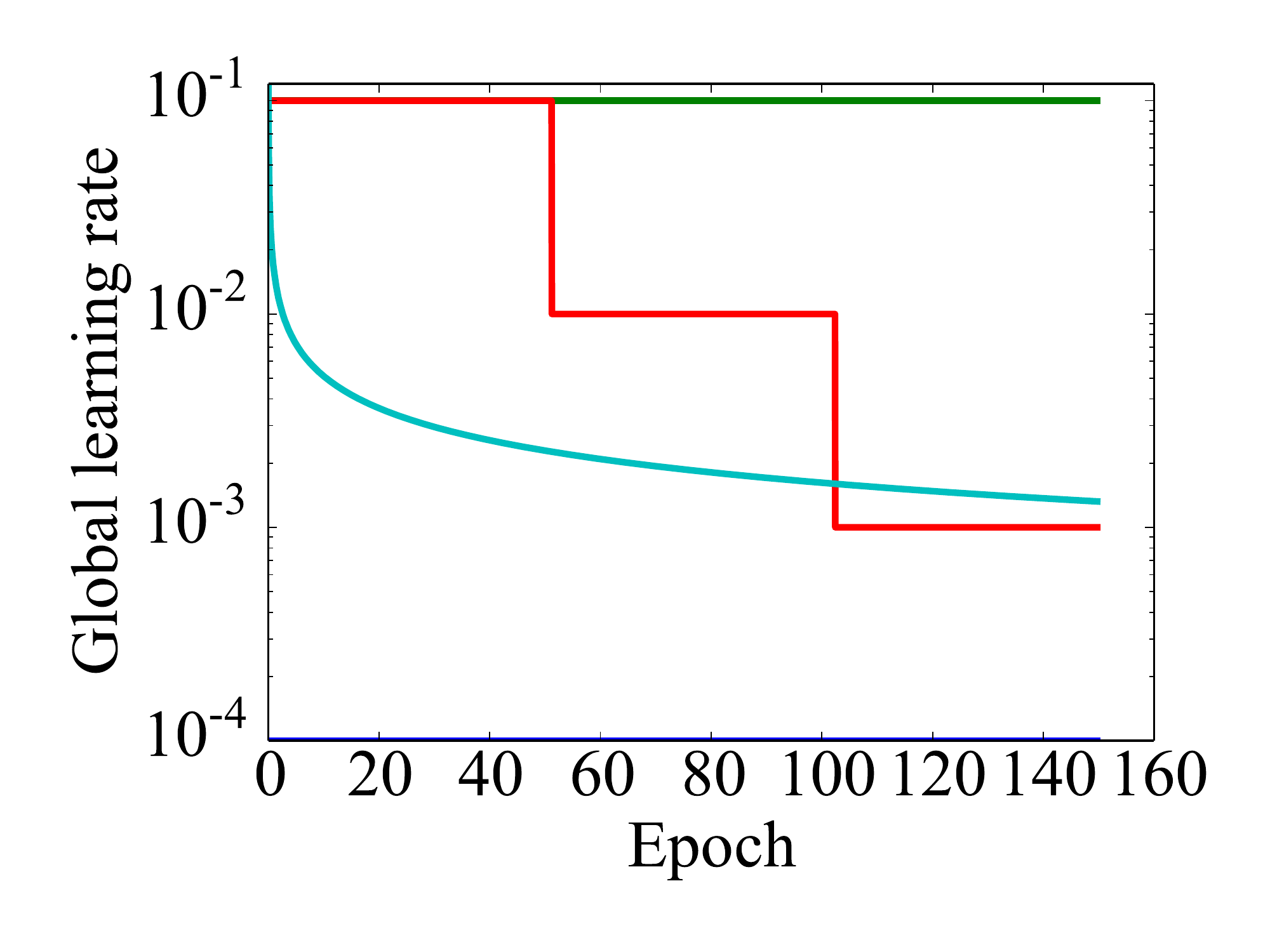}
    \subcaption{learning rate}
  \end{subfigure}%
  \caption{Smoothed training loss, test accuracy, and (global) learning rate schedule for an 18-layer ResNet
    model \citep{HeZhangRenSun2016ResNet} trained on the CIFAR-10
    dataset using four different methods (with constant momentum
    $\beta=0.9$). Adam: $\alpha=0.0001$; SGM-const: $\alpha=1.0$; SGM-poly: $a=1.0,
    b=1, c=0.5$; SGM-hand: $\alpha=1.0$, drop by 10 every 50 epochs.}
  \label{fig:lrsched}
\end{figure}

Figure~\ref{fig:lrsched} shows the training loss and test accuracy of
a typical deep learning task using four different methods: SGM with
constant step size (SGM-const), SGM with diminishing $O(1/k)$ step
size (SGM-poly), Adam \citep{KingmaBa2014adam}, and hand-tuned SGM
with learning rate scheduling (SGM-hand).  For the last method, we
decrease the step size by a multiplicative factor after a suitably
long number of epochs (``constant-and-cut''). The relative performance
depicted in Figure~\ref{fig:lrsched} is typical of many tasks in deep
learning.  In particular, SGM with a large momentum and a
constant-and-cut step-size schedule often achieves the best
performance. Many former and current state-of-the-art results use
constant-and-cut schedules during training, such as
those in image classification \citep{huang2018gpipe}, object detection
\citep{szegedy2015going}, machine translation
\citep{gehring2017convolutional}, and speech recognition
\citep{amodei2016deep}. Additionally, some recent theoretical evidence
indicates that in some (strongly convex) scenarios, the
constant-and-cut scheme has better finite-time last-iterate convergence performance
than other methods \citep{ge2019step}.

Inspired by the success of the ``constant-and-cut'' scheduling approach, we develop an algorithm
that can automatically decide when to drop $\alpha$. Most
common heuristics try to identify when
``training progress has stalled.'' We formalize stalled progress as
when the SGM dynamics in~\eqref{eqn:SGM}, with constant values of $\alpha$
and $\beta$, reach a stationary distribution. The existence of such a
distribution seems to match well with many empirical results (e.g.,
Figure \ref{fig:lrsched}), though it may not exist in general.
Since SGM generates a rich set
of information as it runs (i.e. $\{x^0,g^0, \ldots,x^k,g^k\}$), a natural approach is to collect some statistics from this information and perform certain tests on them to decide whether the
process \eqref{eqn:SGM} has reached a stationary distribution.
We call this general method SASA: statistical adaptive stochastic approximation.

Algorithm \ref{alg:const-and-cut} summarizes the general SASA method.
It performs the SGM updates \eqref{eqn:SGM}
in phases of $M$ iterations, in each iteration potentially computing some additional statistics. After $M$ iterations are
complete, the algorithm performs a statistical test to decide whether to
drop the learning rate by a factor $\zeta < 1$.
Dropping $\alpha$ after a fixed number of epochs and dropping $\alpha$ based on the loss of a held-out validation set correspond to heuristic versions of Algorithm
\ref{alg:const-and-cut}. In the rest of this work, we detail how to perform the ``test'' procedure
and evaluate SASA on a wide range of deep learning tasks.

\subsection{Related Work and Contributions}
\label{sec:related}
The idea of using statistical testing to augment stochastic
optimization methods goes back at least to \citet{Pflug83}, who
derived a stationary condition for the dynamics of SGD on quadratic
functions and designed a heuristic test to determine when the dynamics had reached stationary. He used this
test to schedule a fixed-factor learning rate drop.
\citet{chee2018convergence} recently re-investigated Pflug's method
for general convex functions.
Pflug's stationary condition relies heavily on a quadratic approximation
to $F$ and limiting noise assumptions, as do several other recent
works that derive a stationary condition
\citep[e.g.,][]{MandtHoffmanBlei2017,chee2018convergence}. Additionally, Pflug's test assumes no
correlation exists between consecutive samples in the optimization
trajectory. Neither is true in practice, which we show in Appendix \ref{apdx:comparing} can lead to poor predictivity of this condition.



\citet{yaida2018fluctuation} derived a very general stationary
condition that does not depend on any assumption about the underlying
function $F$ and applies under general noise conditions
regardless of the size of $\alpha$. Like \citet{Pflug83},
\citet{yaida2018fluctuation} used this condition to determine when to
decrease $\alpha$, and showed good performance compared to hand-tuned
SGM for two deep learning tasks with small models.
However, Yaida's method does not account for
the variance of the terms involved in the test, which we show can
cause large variance in the learning rate schedules in some cases. This variance can
in turn cause poor empirical performance.

In this work, we show how to more rigorously perform statistical hypothesis testing on samples collected from the dynamics of SGM.
We combine this statistical testing with Yaida's stationary condition to
develop an adaptive ``constant-and-cut'' optimizer (SASA) that we show is more robust than present methods.
Finally, we conduct large-scale experiments on a variety of deep learning tasks to demonstrate that
SASA is competitive with the best hand-tuned and validation-tuned methods without requiring additional tuning.

\section{Stationary Conditions}
\label{sec:stationary-condition}
To design a statistical test that fires when SGM reaches a stationary
distribution, we first need to derive a condition that holds at
stationarity and consists of terms that we can estimate during training.
To do so, we analyze the long-run behavior of SGM with
constant learning rate and momentum parameter:
\begin{equation}
  \label{eqn:sgm-fixed}
  \begin{split}
    \dkp &= (1-\beta)\gk + \beta\dk ,\\
    \xkp &= \xk - \alpha\dkp ,
  \end{split}
\end{equation}
where $\alpha>0$ and $0\leq\beta<1$. This process starts with $d^0 =
0$ and arbitrary $x^0$. Since $\alpha$ and $\beta$ are constant, the
sequence $\{\xk\}$ does not converge to a local minimum, but the
distribution of $\{\xk\}$ may converge to a stationary distribution.
Letting $\mathcal{F}_k$ denote the $\sigma$-algebra defined by the
history of the process \eqref{eqn:sgm-fixed} up to time $k$, i.e.,~$\mathcal{F}_k =
\sigma(d^0,\ldots, d^k; x^0, \ldots, \xk)$, we denote by
$\E_{k}[\cdot]:=\E[\cdot|\mathcal{F}_k]$ the expectation conditioned
on that history. Assuming that $\gk$
is Markovian and unbiased, i.e.,
\begin{equation}
\label{eqn:gmarkovian}
\P[\gk|\mathcal{F}_k] = \P[\gk|\dk, \xk], \quad \E[\gk | \dk, \xk] = \nabla F(\xk),
\end{equation}
then the SGM dynamics \eqref{eqn:sgm-fixed} form a
homogeneous\footnote{``Homogeneous'' means that the transition kernel
  is time independent.} Markov chain with continuous state $(\dk, \xk,
\gk) \in \R^{3n}$. These assumptions are always satisfied when $\gk =
\nabla_x f(\xk, \xik)$ for an i.i.d. sample $\xik$.
We further assume that the SGM process
converges to a stationary distribution, denoted as
$\pi(d,x,g)$\footnote{As stated in Section \ref{sec:intro}, this need
  not be true in general, but seems to often be the case in
  practice.}. With this notation, we need a relationship $\E_\pi[X] =
\E_\pi[Y]$ for certain functions $X$ and $Y$ of $(\xk, \dk, \gk)$
that we can compute during training. Then, if we assume the Markov chain is
ergodic, we have that:
\begin{equation}
\label{eqn:mean-conv}
\bar{z}_N = \frac{1}{N}\sum_{i=0}^{N-1}z_i = \frac{1}{N}\sum_{i=0}^{N-1}\left(X(x^i,d^i,g^i) - Y(x^i,d^i,g^i)\right) \to 0.
\end{equation}
Then we can check the magnitude of the time-average $\bar{z}_N$ to see
how close the dynamics are to reaching their stationary distribution.
Next, we consider two different stationary conditions.


\subsection{Pflug's condition}
Assuming $F(x) = (1/2)x^T A x$, where $A$ is
positive definite with maximum eigenvalue $L$, and that the stochastic gradient $\gk$ satisfies $\gk = \nabla
F(\xk) + r^k$, with $\E[r^k] = 0$ and $r^k$ independent of $\xk$, \citet{Pflug83} derived a stationary condition for the SGD dynamics. His condition can be extended to the SGM dynamics. For appropriate $\alpha$ and $\beta$, the generalized Pflug stationary condition says
\begin{equation}\label{eqn:gkp-dk-correlation}
  \E_\pi\bigl[\langle g,d\rangle\bigr]
  = -\frac{\alpha(1-\beta)}{2(1+\beta)} \tr(A\Sigma_r) + O(\alpha^2),
\end{equation}
where $\Sigma_r$ is the covariance of the noise $r$. One can estimate the left-hand-side during training by computing the inner product $\langle \gk, \dk\rangle$ in each iteration. \citet{Pflug83} also designed a clever estimator for the right-hand-side,
so it is possible to compute estimators for both sides of \eqref{eqn:gkp-dk-correlation}.

The Taylor expansion in $\alpha$ used to derive \eqref{eqn:gkp-dk-correlation} means that the relationship may
only be accurate for small $\alpha$, but $\alpha$ is typically large
in the first phase of training. This, together with the other
assumptions required for Pflug's condition, are too strong to
make the condition \eqref{eqn:gkp-dk-correlation}
useful in practice.


\subsection{Yaida's condition}
\citet{yaida2018fluctuation} showed that as long as the stationary
distribution $\pi$ exists, the following relationship holds \emph{exactly:}
\vspace{-1ex}
\begin{equation*}
\E_{\pi}[\langle x, \nabla F(x)\rangle ] = \frac{\alpha}{2}\frac{1 + \beta}{1-\beta}\E_{\pi}[ \langle d, d\rangle ]
\end{equation*}
In particular, this holds for general functions $F$ and arbitrary
values of $\alpha$. Because the
stochastic gradients $\gk$ are unbiased, one can further show that:
\begin{equation}
\label{eqn:fdr1}
\E_{\pi}[\langle x, g \rangle ] = \frac{\alpha}{2}\frac{1 + \beta}{1-\beta}\E_{\pi}[ \langle d, d\rangle ].
\end{equation}
We can estimate both sides of \eqref{eqn:fdr1} by computing $\langle \xk, \gk\rangle$ and $\langle \dk, \dk \rangle = ||\dk||^2$ at each iteration and updating the running mean $\bar{z}_N$ with their difference. That is, we let
\begin{equation}
    z_k = \langle \xk, \gk\rangle - \frac{\alpha}{2}\frac{1 + \beta}{1-\beta}\langle \dk, \dk \rangle \qquad
    \bar{z}_N = \frac{1}{N}\sum_{k=B}^{N+B-1}z_k\label{eqn:yaida-zbar}.
\end{equation}
Here $B$ is the number of samples discarded as part of a ``burn-in'' phase to reduce bias that might be caused by starting far away from the stationary distribution; we typically take $B = N/2$, so that we use the most recent $N/2$ samples.

Yaida's condition has two key advantages over Pflug's: it holds with no approximation for arbitrary functions $F$ and any learning rate $\alpha$, and both sides can be estimated with negligible cost. In Appendix \ref{apdx:comparing}, we show in  Figure \ref{fig:comparisonfig} that even on a strongly convex function, the error term in \eqref{eqn:gkp-dk-correlation} is large, whereas $\bar{z}_N$ in~\eqref{eqn:yaida-zbar} quickly converges to zero. Given these advantages, in the next section, we focus on how to test \eqref{eqn:fdr1}, i.e., that $\bar{z}_N$ defined in \eqref{eqn:yaida-zbar} is approximately zero.

\section{Testing for Stationarity}
\label{sec:testing}
By the Markov chain law of large numbers, we know that $\bar{z}_N\to0$ as $N$ grows, but there are multiple ways to determine whether $\bar{z}_N$ is ``close enough'' to zero that we should drop the learning rate.

\paragraph{Deterministic test.}
If in addition to $\bar{z}_N$ in~\eqref{eqn:yaida-zbar}, we keep track of
\begin{equation}
\label{eqn:mean-d2}
\vspace{-1em}
\bar{v}_N = \frac{1}{N}\sum_{i=B}^{N+B-1}\frac{\alpha}{2}\frac{1 + \beta}{1-\beta}\langle d^i, d^i\rangle,
\end{equation}
A natural idea is to test
\begin{equation}
\label{eqn:yaida-test}
|\bar{z}_N| < \delta\bar{v}_N \quad \text{or equivalently} \quad |\bar{z}_N/\bar{v}_N| < \delta
\end{equation}
to detect stationarity, where $\delta > 0$ is an error tolerance.
The $\bar{v}_N$ term is introduced
to make the error term $\delta$ relative to the scale of $\bar{z}$ and
$\bar{v}$ ($\bar{v}_N$ is always nonnegative). If $\bar{z}_N$ satisfies \eqref{eqn:yaida-test}, then the dynamics
\eqref{eqn:sgm-fixed} are ``close'' to stationarity.  This is precisely the method used by \citet{yaida2018fluctuation}.

However, because $\bar{z}_N$ is a random variable, there is some
potential for error in this procedure due to its variance, which is
unaccounted for by \eqref{eqn:yaida-test}. Especially when we aim to
make a critical decision based on the outcome of this test (i.e.,
dropping the learning rate), it seems important to more directly
account for this variance. To do so, we can appeal to statistical hypothesis testing.

\paragraph{I.i.d. $t$-test.}
The simplest approach to accounting for the variance in $\bar{z}_N$ is to assume each sample $z_i$ is drawn i.i.d. from the same distribution. Then by the central limit theorem, we have that $\sqrt{N}\bar{z}_N \to \mathcal{N}(0, \sigma_z^2)$, and moreover
$\hat{\sigma}_N^2 = \frac{1}{N-1}\sum_{i=1}^N(z_i - \bar{z}_N)^2 \approx \sigma_z^2$ for large $N$. So we can estimate the variance of $\bar{z}_N$'s sampling distribution using the sample variance of the $z_i$'s.
Using this variance estimate, we can form the $(1-\gamma)$ confidence interval
\[
\bar{z}_N \pm t_{1-\gamma/2}^*\frac{\hat{\sigma}_N}{\sqrt{N}},
\]
where $t_{1-\gamma/2}^*$ is the $(1-\gamma/2)$ quantile of the Student's
$t$-distribution with $N-1$ degrees of freedom. Then we can check whether
\begin{equation}
\label{eqn:our-test}
\biggl[\bar{z}_N - t_{1-\gamma/2}^*\frac{\hat{\sigma}_N}{\sqrt{N}}, ~\bar{z}_N + t_{1-\gamma/2}^*\frac{\hat{\sigma}_N}{\sqrt{N}}\biggr] \in \bigl(-\delta\bar{v}_N, ~\delta\bar{v}_N\bigr).
\end{equation}
If so, we can be confident that $\bar{z}_N$ is close to zero. The method of \citet[][Algorithm 4.2]{Pflug83} is also a kind of i.i.d. test that tries to account for the variance of $\bar{z}_N$, but in a more heuristic way than \eqref{eqn:our-test}.
The procedure \eqref{eqn:our-test} can be
thought of as a relative \emph{equivalence test} in statistical
hypothesis testing \citep[e.g.][]{streiner2003unicorns}. When $\hat{\sigma}_N = 0$ (no variance) or $\gamma=1$ ($t_{1-\gamma/2}^*=0$, no confidence), this recovers
\eqref{eqn:yaida-test}.

Unfortunately, in our case, samples $z_i$ evaluated at nearby points are highly correlated (due to the underlying Markov dynamics), which makes this procedure inappropriate. To deal with correlated samples, we appeal to a stronger Markov chain result than the Markov chain law of large numbers \eqref{eqn:mean-conv}.

\paragraph{Markov chain $t$-test}
Under suitable conditions, Markov chains admit the following analogue
of the central limit theorem:
\begin{theorem}[Markov Chain CLT (informal); \citep{jones2006fixed}]
\label{thm:markov-clt}
  Let $X = \{X_0, X_1, \ldots\}$ be a Harris ergodic Markov chain with
  state space $\mathcal{X}$, and with stationary distribution $\pi$,
  that satisfies any one of a number of additional ergodicity criteria
  (see \citet{jones2006fixed}, page 6). For suitable functions $z:
  \mathcal{X} \to \mathbb{R}$, we have that:
  \[
  \sqrt{N}\left(\bar{z}_N - \E_\pi z\right) \to \mathcal{N}(0, \sigma_z^2),
  \]
  where $\bar{z}_N = \frac{1}{N}\sum_{i=0}^{N-1}z(X_i)$ is the running
  mean over time of $z(X_i)$, and $\sigma_z^2 \ne \text{var}_\pi z$ in
  general due to correlations in the Markov chain.
\end{theorem}
This shows that in the presence of correlation, the sample variance is not the correct estimator for the variance of $\bar{z}_N$'s sampling distribution. In light of Theorem \ref{thm:markov-clt}, if we are given a consistent estimator
$\hat{\sigma}^2_N \to \sigma_z^2$, we can properly perform the test \eqref{eqn:our-test}.
All that remains is to construct such an estimator.

\paragraph{Batch mean variance estimator.}
Methods for estimating the asymptotic variance of the history average
estimator, e.g., $\bar{z}_N$ in \eqref{eqn:yaida-zbar}, on a Markov
chain are well-studied in the MCMC (Markov chain Monte Carlo) literature.
They can be used to set a stopping time for an MCMC
simulation and to determine the simulation's random error
\citep{jones2006fixed}. We present one of the simplest estimators for
$\sigma_z^2$, the \emph{batch means} estimator.

Given $N$ samples $\{z_i\}$, divide them into $b$ batches each of size
$m$, and compute the batch means:
$\bar{z}^j = \frac{1}{m}\sum_{i=jm}^{(j+1)m - 1}z_i$ for each batch $j$.
Then let
\begin{equation}
\label{eqn:bm}
\hat{\sigma}_N^2 = \frac{m}{b-1}\sum_{j=0}^{b-1}(\bar{z}^j - \bar{z}_N)^2.
\end{equation}
Here $\hat{\sigma}_N^2$ is simply the variance of the batch means around
the full mean $\bar{z}_N$. When used in the test \eqref{eqn:our-test}, it has $b-1$ degrees of freedom.
Intuitively, when $b$ and $m$ are both
large enough, these batch means are roughly independent because of the
mixing of the Markov chain, so their unbiased sample variance gives a good estimator of
$\sigma_z^2$. \citet{jones2006fixed} survey the formal conditions
under which $\hat{\sigma}_N^2$ is a strongly consistent estimator of
$\sigma_z^2$, and suggest taking $b = m = \sqrt{N}$ (the theoretically
correct sizes of $b$ and $m$ depend on the mixing of the Markov
chain). \citet{flegal2010batch} prove strong consistency for a related
method called \emph{overlapping batch means} (OLBM) that has better
asymptotic variance. The OLBM estimator is similar to \eqref{eqn:bm}, but uses
$n - b + 1$ overlapping batches of size $b$ and has $n - b$ degrees of freedom.


\begin{figure}[!t]
\centering
\begin{minipage}{0.57\textwidth}
\begin{algorithm}[H]
  \SetKwInput{Input}{Input}
  \Input{$\{x^0, \alpha_0, M, \beta, \delta, \gamma, \zeta\}$}
  zQ = HalfQueue()\\
  vQ = HalfQueue()\\
  \For{$j \in \{0,1,2,\ldots\}$}{
    \For{$k \in \{jM,\ldots,(j+1)M-1\}$}
        {
          Sample $\xik$ and compute $\gk = \nabla_x f(\xk, \xik)$\\
          $\dkp = (1-\beta)\gk + \beta\dk$\\
          $\xkp = \xk - \alpha\dkp$\\
          zQ.push($\langle \xk, \gk\rangle - \frac{\alpha}{2}\frac{1 + \beta}{1-\beta} ||\dkp||^2$)\\
          vQ.push($\frac{\alpha}{2}\frac{1 + \beta}{1-\beta} ||\dkp||^2$)
        }
    \If{$test(zQ, vQ, \delta, \gamma)$}
    {
      $\alpha \gets \zeta \alpha $\\
      zQ.reset()\\
      vQ.reset()
    }
  }
  \caption{SASA}
  \label{alg:sasa}
\end{algorithm}
\end{minipage}
\hfill
\begin{minipage}{0.41\textwidth}
\begin{algorithm}[H]
  \SetKwInput{Input}{Input}
  \SetKwInput{Output}{Output}
  \Input{$\{zQ, vQ, \delta, \gamma\}$}
  \Output{boolean (whether to drop)}
  $\bar{z}_N = \frac{1}{zQ.N}\sum_{i} zQ[i]$\\
  $\bar{v}_N = \frac{1}{vQ.N}\sum_{i} vQ[i]$\\
  $m = b = \sqrt{zQ.N}$\\
  \For{$i \in \{0, \ldots, b-1\}$} {
    $\bar{z}^i = \frac{1}{m}\sum_{t=im}^{(i+1)m - 1}zQ[t]$\\
    }
  $\hat{\sigma}_N^2 = \frac{m}{b-1}\sum_{i=0}^{b-1}(\bar{z}^i - \bar{z}_N)^2.$\\
  $L = \bar{z}_N - t_{1-\gamma/2}^*\frac{\hat{\sigma}_N}{\sqrt{zQ.N}}$\\
  $U = \bar{z}_N + t_{1-\gamma/2}^*\frac{\hat{\sigma}_N}{\sqrt{zQ.N}}$\\
  \Return{$\left[L, U\right] \in (-\delta\bar{v}_N, \delta\bar{v}_N)$}
  \caption{Test}
  \label{alg:markov-t-test}
  \vspace{16.6mm}
\end{algorithm}
\end{minipage}
\end{figure}

\subsection{Statistical adaptive stochastic approximation (SASA)}
Finally, we turn the above analysis into an adaptive algorithm for
detecting stationarity of SGM and decreasing $\alpha$, and discuss
implementation details. Algorithm \ref{alg:sasa} describes our full
SASA algorithm.

To diminish the effect of ``initialization bias'' due to starting outside of the stationary distribution,
we only keep track of the latter half of samples $z_i$ and $v_i$. That is, if $N$ total iterations of SGM have been run, the ``HalfQueues'' $zQ$ and $vQ$ contain
the most recent $N/2$ values of $z_i$ and $v_i$---these queues ``pop'' every other time they ``push.''
If we decrease the learning rate, we empty the
queues; otherwise, we keep collecting more samples. To compute the batch mean estimator, we need $O(N)$ space, but in deep learning the total number of training iterations (the worst case size of these queues) is usually small compared to the number of parameters of the model.
Collection of the samples $z_i$ and $v_i$ only requires two more inner products per iteration than SGM.

The ``test'' algorithm follows the Markov chain $t$-test procedure discussed above.
Lines 1-2 compute the running means $\bar{z}_N$ and $\bar{v}_N$;
lines 3-7 compute the variance estimator $\hat{\sigma}_N^2$ according to \eqref{eqn:bm},
and lines 8-10 determine whether the corresponding confidence interval
for $\bar{z}_N$ is within the acceptable interval $(-\delta\bar{v}_N, \delta\bar{v}_N)$.
Like the sample collection, the test procedure is computationally efficient:
the batch mean and overlapping batch mean estimators can both be computed with a 1D convolution.

For all experiments, we use default values $\delta=0.02$ and $\gamma = 0.2$. In equivalence testing, $\gamma$ is typically taken larger than usual to increase the power of the test \citep{streiner2003unicorns}. We discuss the apparent multiple testing problem of this sequential testing procedure in Appendix \ref{apdx:more-discussion}.

\begin{figure}[tb]
  \centering
  \begin{subfigure}[t]{.29\linewidth}
    \centering
    \includegraphics[width=\linewidth]{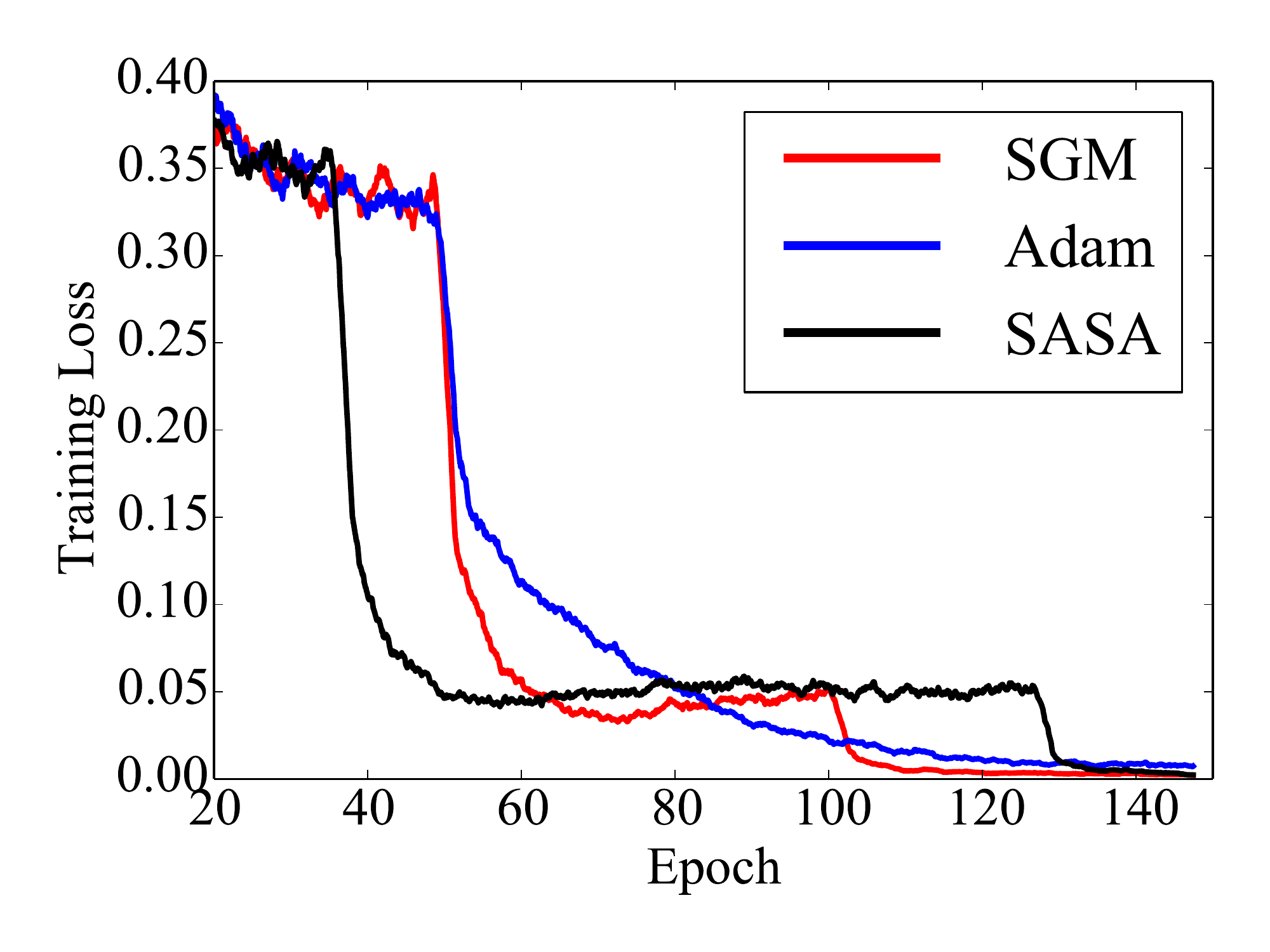}
  \end{subfigure}%
  \begin{subfigure}[t]{.29\linewidth}
    \centering
    \includegraphics[width=\linewidth]{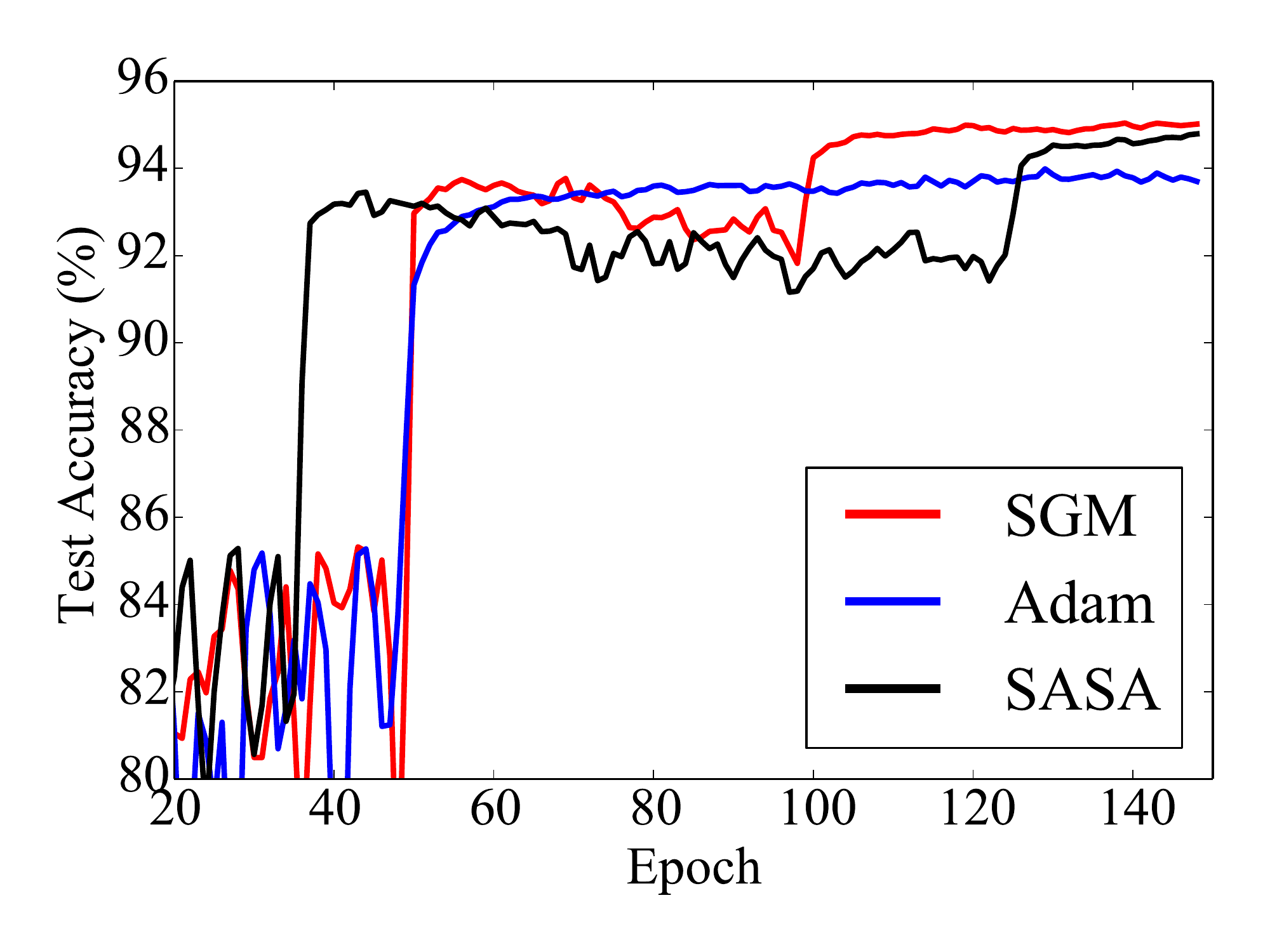}
  \end{subfigure}%
  \begin{subfigure}[t]{.29\linewidth}
    \centering
    \includegraphics[width=\linewidth]{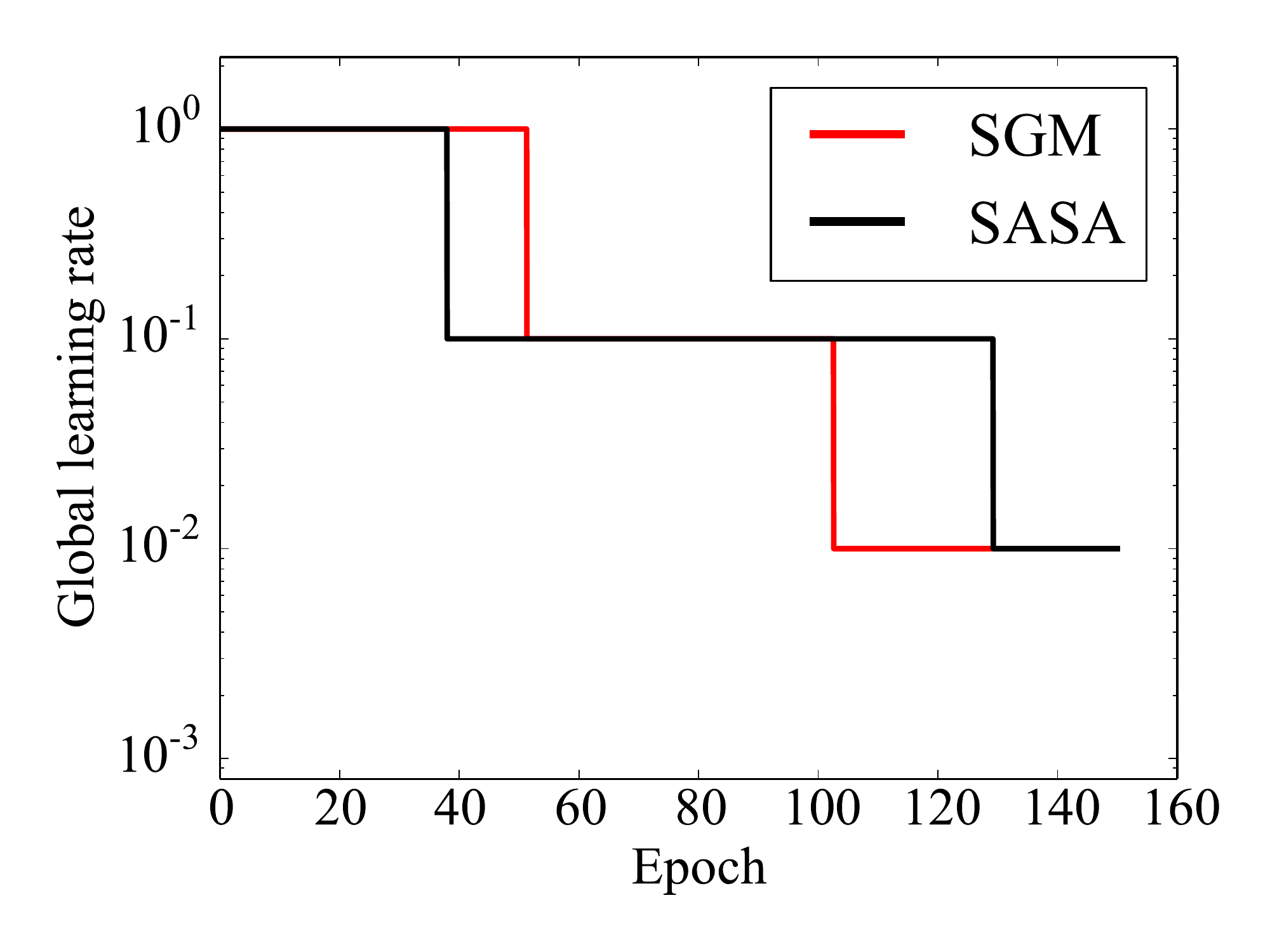}
  \end{subfigure}\\
  \begin{subfigure}[t]{.29\linewidth}
    \centering
    \includegraphics[width=\linewidth]{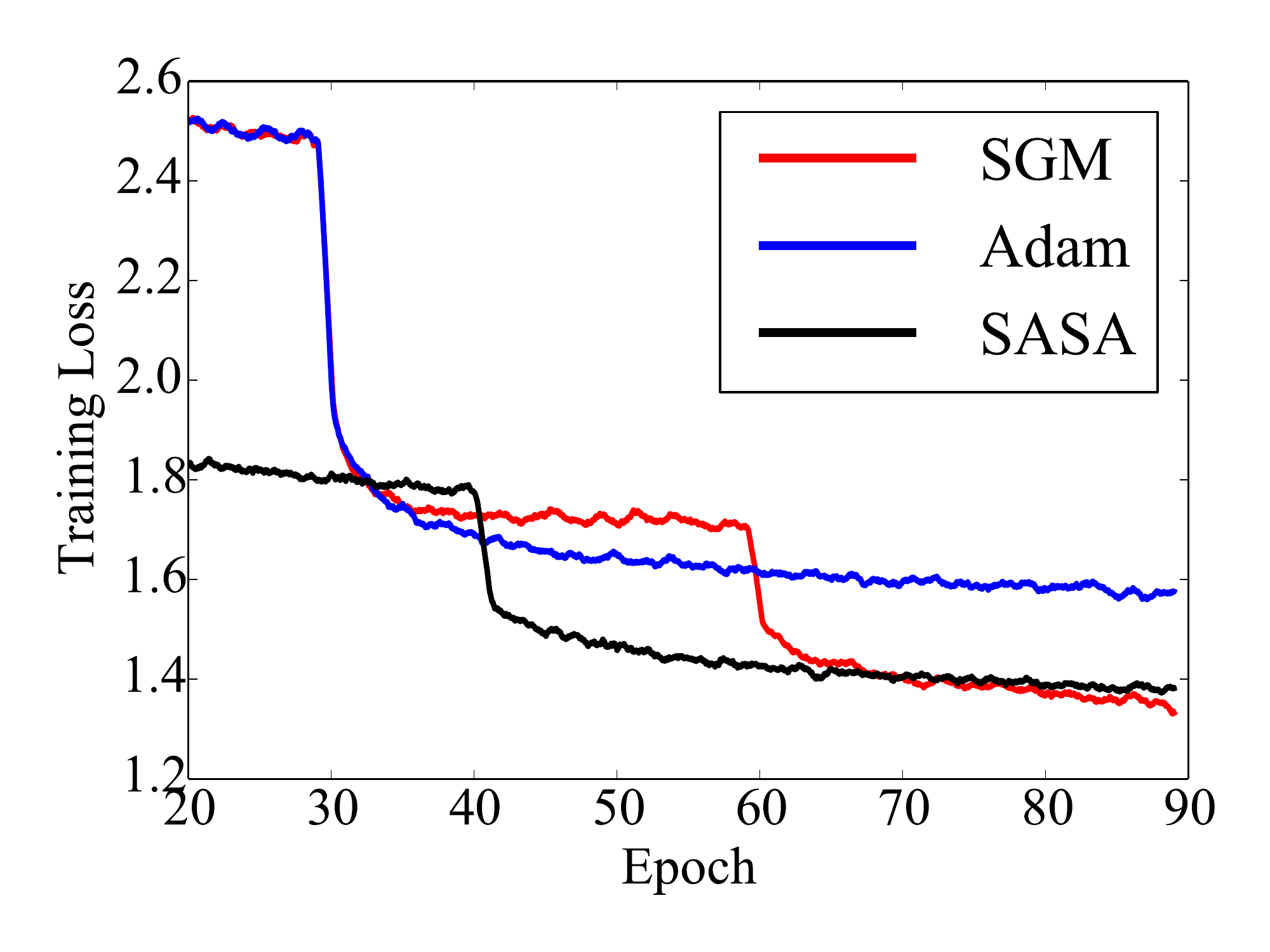}
  \end{subfigure}%
  \begin{subfigure}[t]{.29\linewidth}
    \centering
    \includegraphics[width=\linewidth]{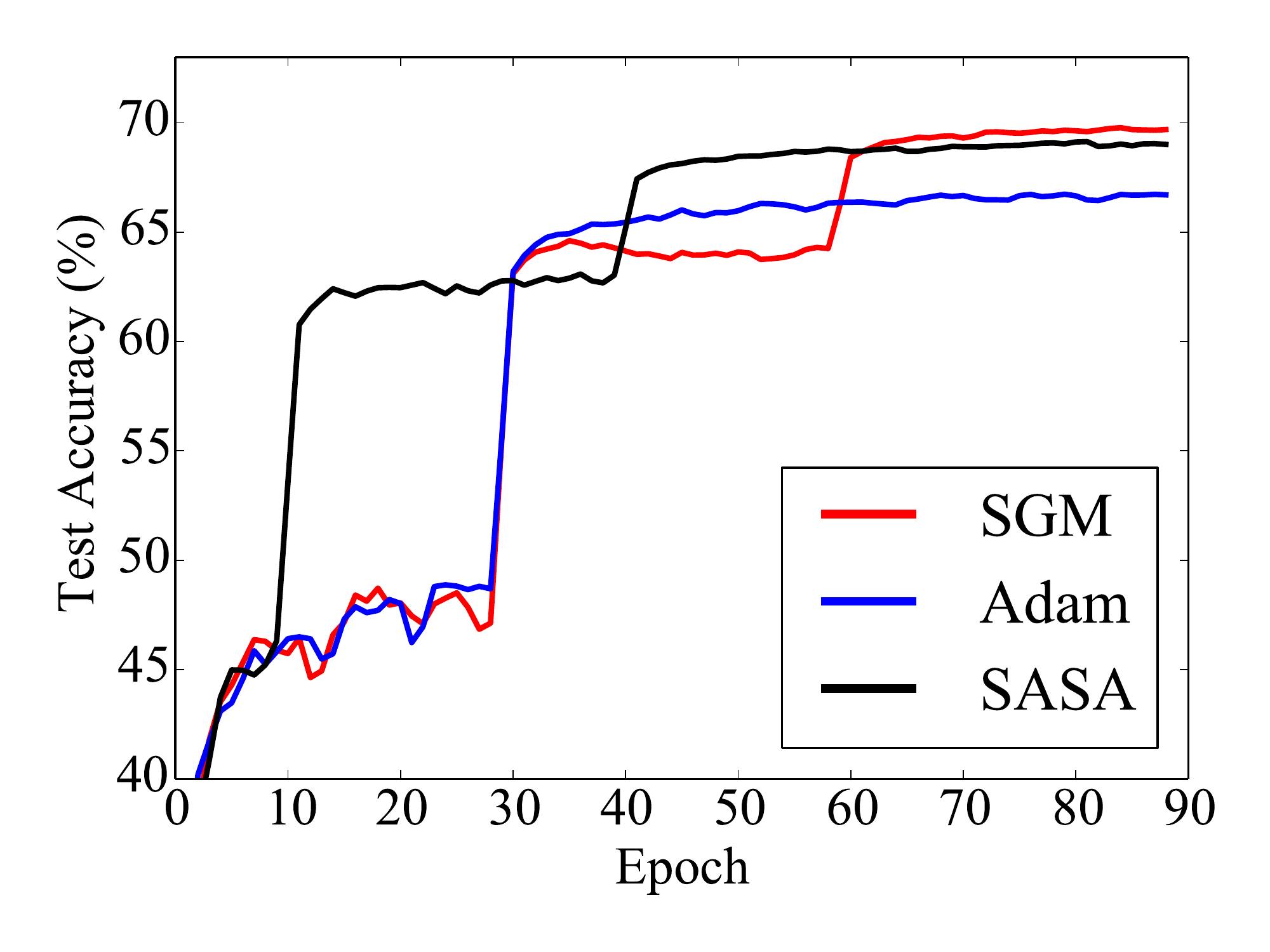}
  \end{subfigure}%
  \begin{subfigure}[t]{.29\linewidth}
    \centering
    \includegraphics[width=\linewidth]{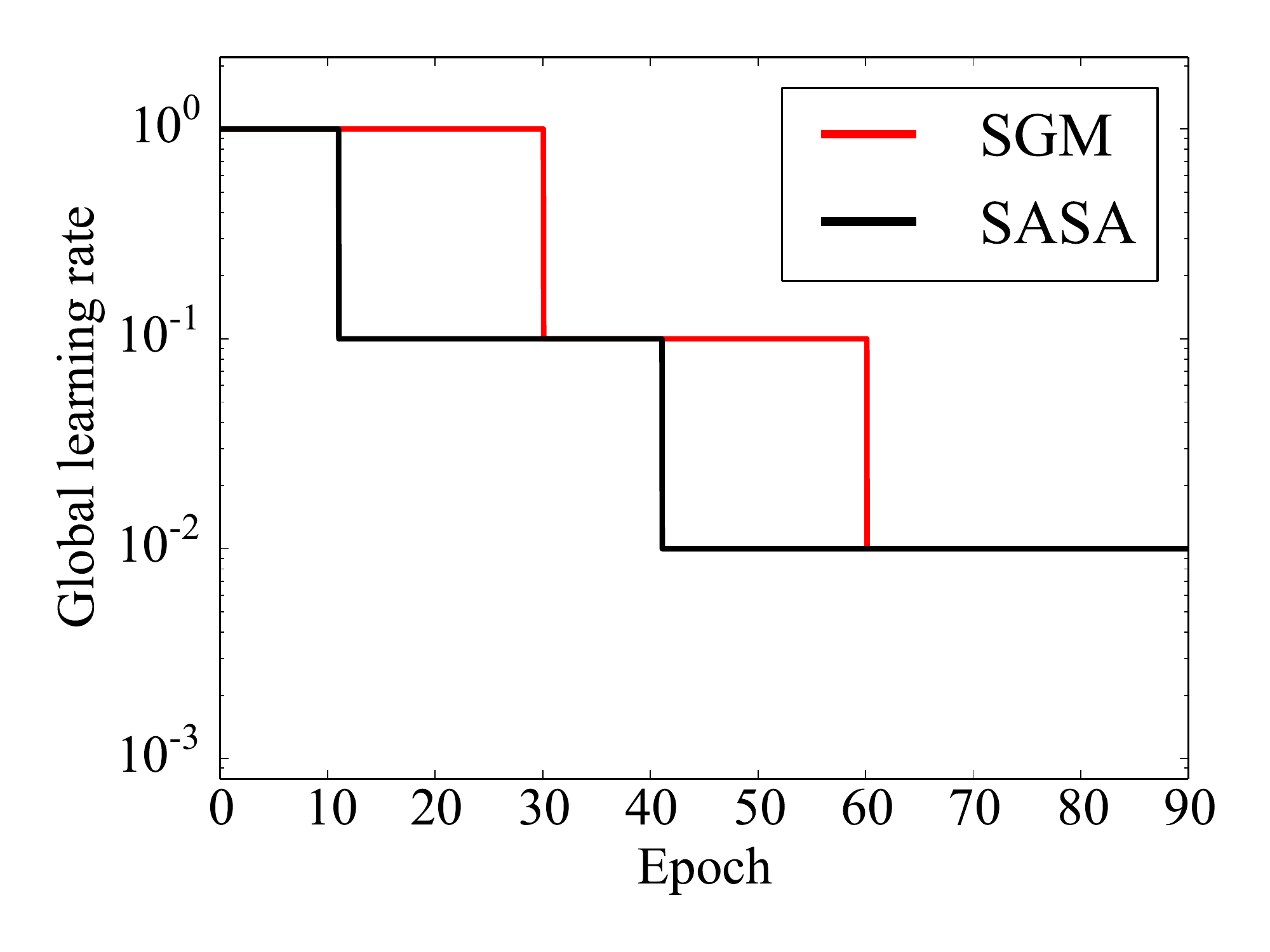}
  \end{subfigure}\\
  \begin{subfigure}[t]{.29\linewidth}
    \centering
    \includegraphics[width=\linewidth]{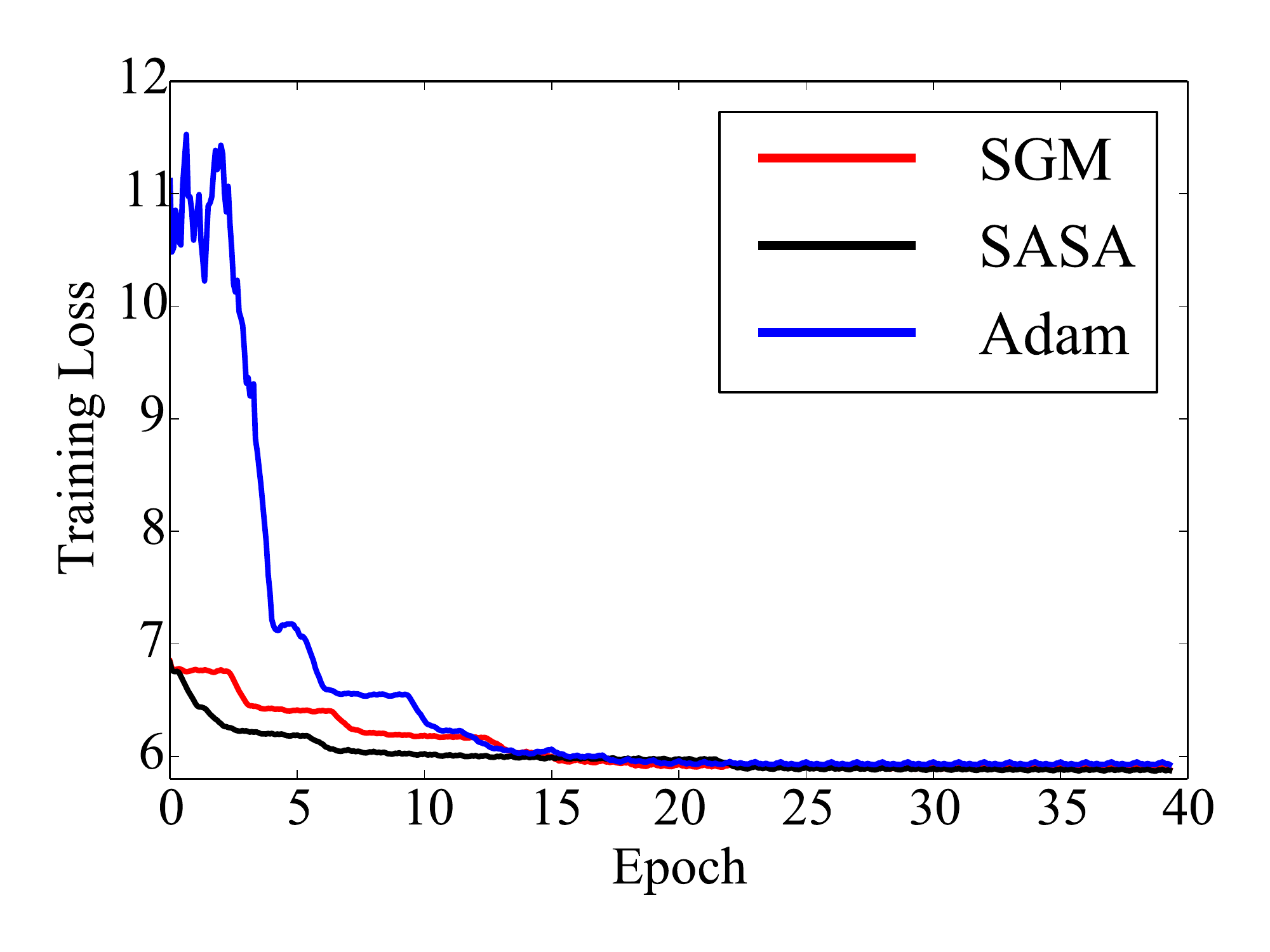}
  \end{subfigure}%
  \begin{subfigure}[t]{.29\linewidth}
    \centering
    \includegraphics[width=\linewidth]{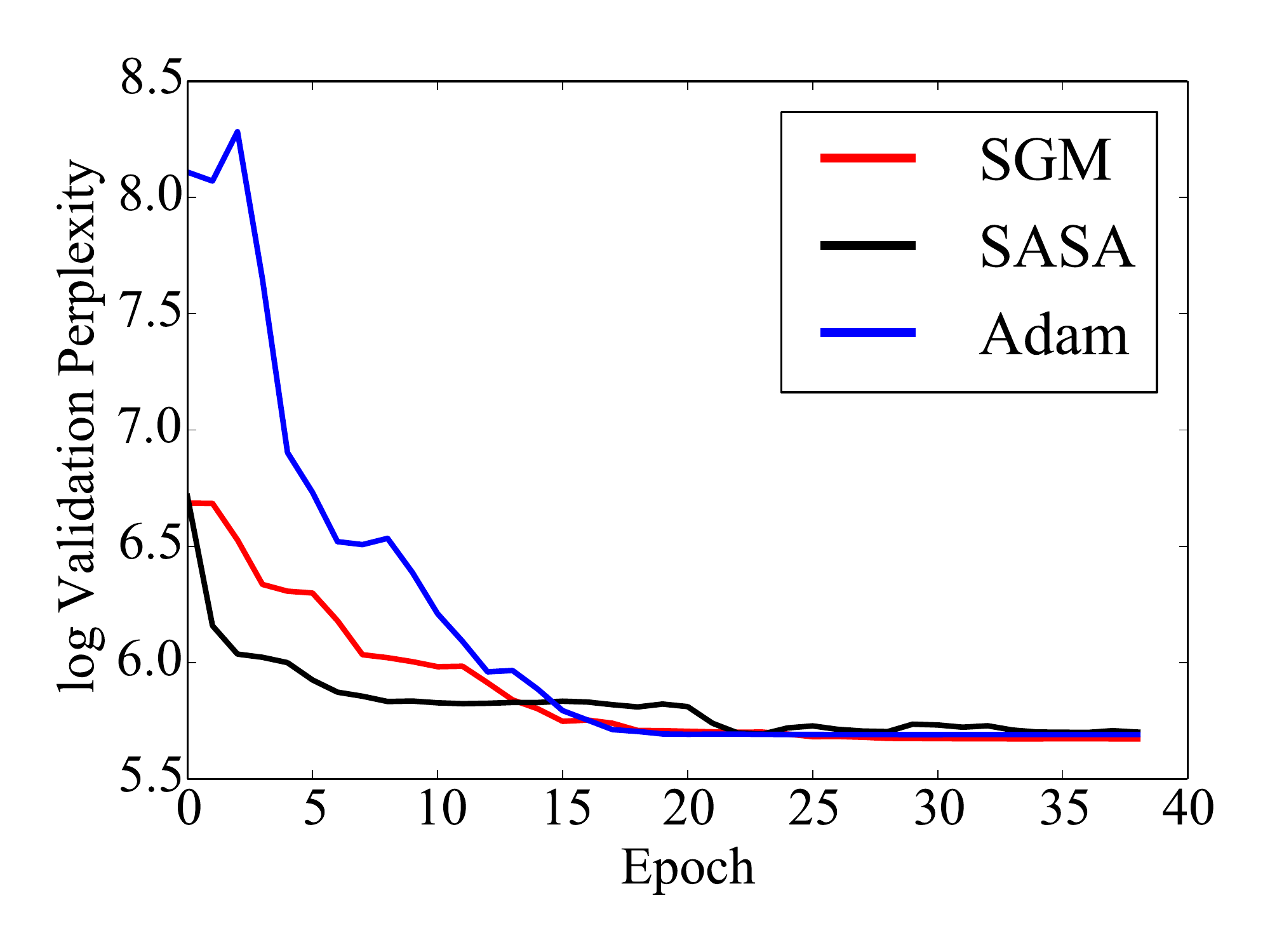}
  \end{subfigure}%
  \begin{subfigure}[t]{.29\linewidth}
    \centering
    \includegraphics[width=\linewidth]{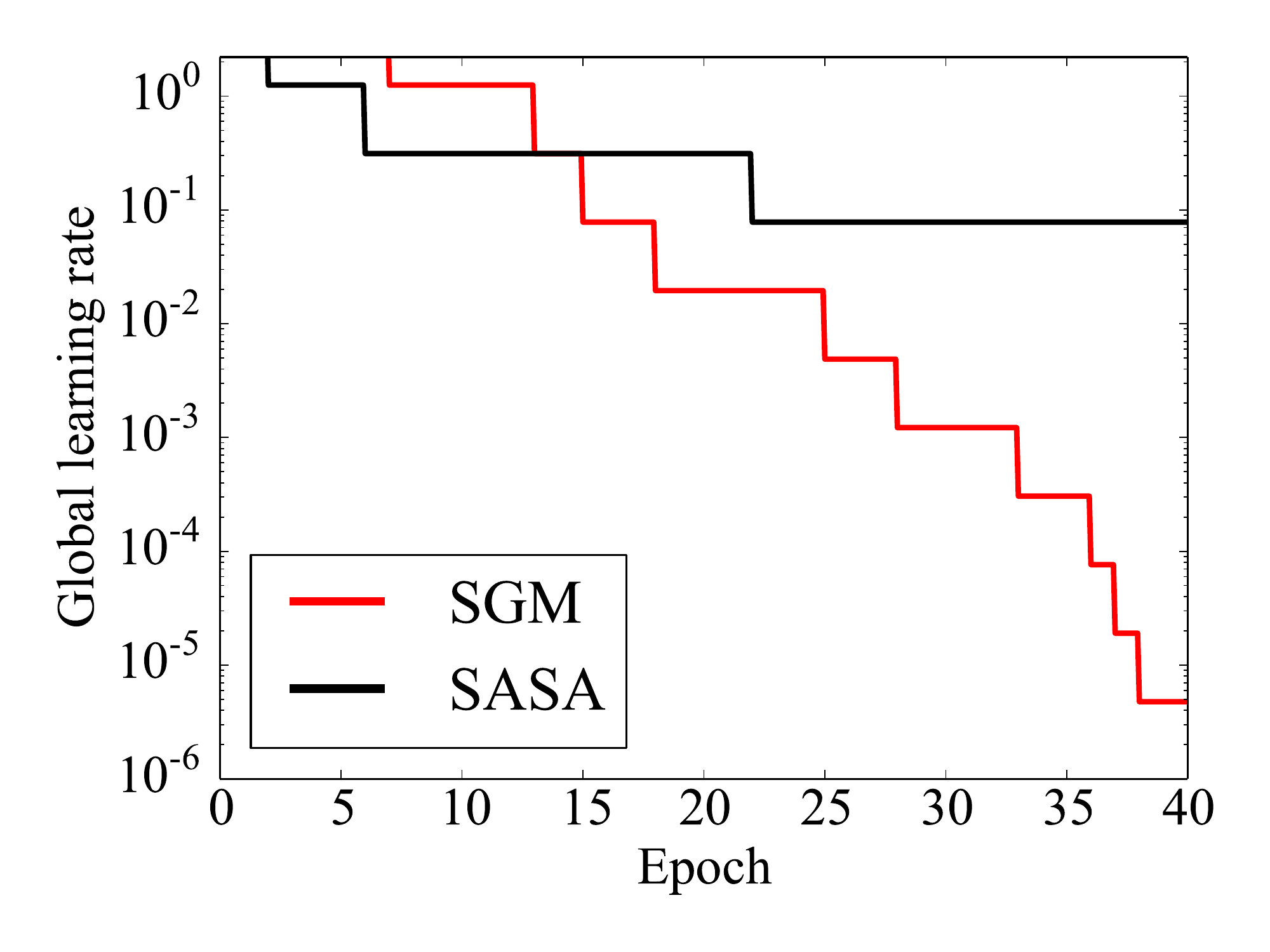}
  \end{subfigure}
  \caption{Training loss, test accuracy, and learning rate schedule for SASA, SGM, and Adam on different datasets. Top: ResNet18 on CIFAR-10. Middle: ResNet18 on ImageNet. Bottom: RNN model on WikiText-2.
  In all cases, starting with the same $\alpha_0$, SASA achieves similar performance to the best hand-tuned or validation-tuned SGM result.}
  \label{fig:results}
\end{figure}

\section{Experiments}
\label{sec:experiments}
To evaluate the performance of SASA, we run Algorithm \ref{alg:sasa}
on several models from deep
learning. We compare SASA to \emph{tuned} versions of Adam and
SGM. Many adaptive optimizers do not compare to SGM with hand-tuned
step size scheduling, \citep[e.g.,][]{schaul2013no,
  zhang2017yellowfin, Baydin18hypergradient}, and instead compare to
SGM with a fixed $\alpha$ or to SGM with tuned polynomial decay. As
detailed in Section \ref{sec:intro}, tuned constant-and-cut schedules
are typically a stronger baseline.

Throughout this section, we \emph{do not tune the SASA parameters} $\delta, \gamma, M$,
instead using the default settings of $\delta=0.02$ and $\gamma = 0.2$, and setting $M = $ one epoch (we test the statistics once per epoch).
In each experiment, we use the same $\alpha_0$ and $\zeta$ as for the best SGM baseline.
We stress that SASA is not fully automatic: it requires choices of $\alpha_0$ and $\zeta$, but we show in Appendix \ref{apdx:exps} that SASA achieves good performance for different values of $\zeta$. We use weight decay in every
experiment---without weight decay, there are simple examples where the
process \eqref{eqn:sgm-fixed} does not converge to a stationary
distribution, such as with logistic regression on separable
data. While weight decay does not guarantee convergence to a
stationary distribution, it at least rules out this simple
case. Finally, we conduct an experiment on CIFAR-10 that shows
directly accounting for the variance of the test statistic, as in
\eqref{eqn:our-test}, improves the robustness of this procedure
compared to \eqref{eqn:yaida-test}.

For hand-tuned SGM (SGM-hand), we searched over ``constant-and-cut''
schemes for each experiment by tuning $\alpha_0$, the drop
frequency, and the drop amount $\zeta$ with grid search. In all experiments, SASA and SGM use a constant $\beta=0.9$.
For Adam, we tuned the initial global learning rate as in
\citet{Wilson2017marginal} and used $\beta_1 = 0.9$, $\beta_2 = 0.999$. We also allowed Adam to have access to a
``warmup'' phase to prevent it from decreasing the learning rate too
quickly. To ``warm up'' Adam, we initialize it with the parameters obtained after running SGM with constant $\alpha_0$ for a tuned number of iterations. While the warmup phase improves Adam's performance, it still
does not match SASA or SGM on the tasks we tried.
Appendix \ref{apdx:exps} contains a
full list of the hyperparameters used in each experiment, additional results for object detection, sensitivity analysis for $\delta$ and $\gamma$, plots of the evolution of the statistic $\bar{z}_N$ over training, and plots of the different estimators for the variance $\sigma_z^2$.

\vspace{-1.5ex}
\paragraph{CIFAR-10.}
\label{sec:cifar-exp}
We trained an 18-layer ResNet model \citep{HeZhangRenSun2016ResNet} on
CIFAR-10 \citep{krizhevsky2009learning} with random cropping and
random horizontal flipping for data augmentation and weight decay
$0.0005$.  Row 1 of Figure \ref{fig:results} compares the best
performance of each method. Here SGM-hand uses $\alpha_0 = 1.0$ and
$\beta=0.9$ and drops $\alpha$ by a factor of 10 ($\zeta=0.1$) every 50
epochs. SASA uses $\gamma=0.2$ and $\delta=0.02$, as always.
Adam has a tuned global learning rate $\alpha_0=0.0001$ and a tuned ``warmup'' phase of 50 epochs, but is unable to match SASA and tuned SGM.

\vspace{-1.5ex}
\paragraph{ImageNet.}
\label{sec:imagenet-exp}
Unlike CIFAR-10, reaching a good performance level on ImageNet
\citep{imagenet_cvpr09} seems to require more gradual annealing. Even
when tuned and allowed to have a long warmup phase, Adam failed to
match the generalization performance of SGM. On the other hand, SASA
was able to match the performance of hand-tuned
SGM using the default values of its parameters. We again used an 18-layer ResNet model with random cropping,
random flipping, normalization, and weight decay 0.0001. Row 2 of
Figure \ref{fig:results} shows the performance of the different
optimizers.

\vspace{-1.5ex}
\paragraph{RNN.}
\label{sec:rnn-exp}
We also evaluate SASA on a language modeling task using an RNN. In
particular, we train the PyTorch word-level language model example
\citeyearpar{pytorchmodel} on the Wikitext-2 dataset
\citep{merity2016pointer}. We compare against SGM and Adam with (global) learning rate tuned using a validation
set. These baselines drop the learning rate
$\alpha$ by a factor of 4 when the validation loss stops
improving. Row 3 of Figure~\ref{fig:results} shows that \emph{without using
the validation set}, SASA is competitive with these baselines.

\begin{figure}[t]
    \centering
  \begin{subfigure}[t]{.29\linewidth}
    \centering
    \includegraphics[width=\linewidth]{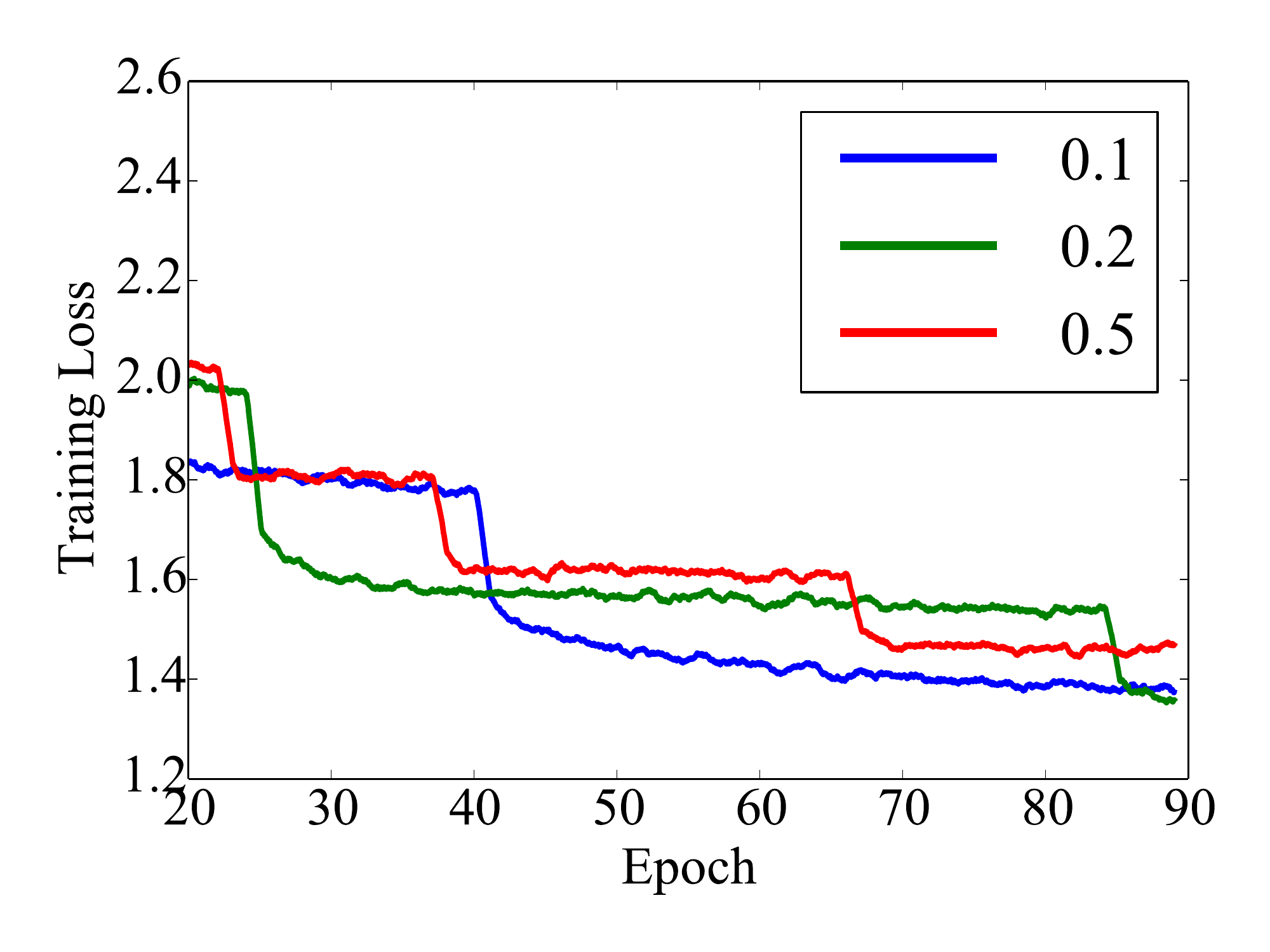}
  \end{subfigure}%
  \begin{subfigure}[t]{.29\linewidth}
    \centering
    \includegraphics[width=\linewidth]{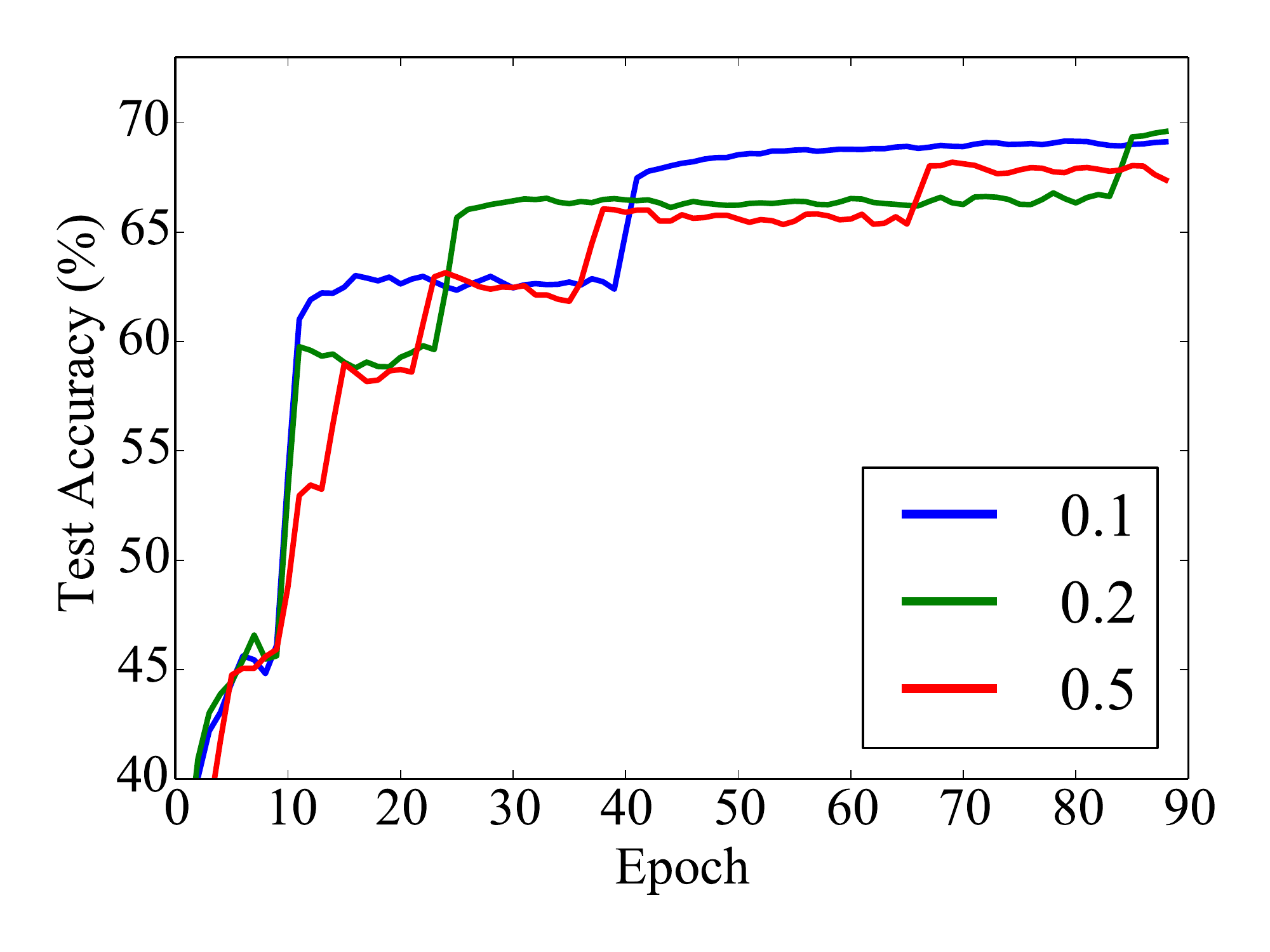}
  \end{subfigure}%
  \begin{subfigure}[t]{.29\linewidth}
    \centering
    \includegraphics[width=\linewidth]{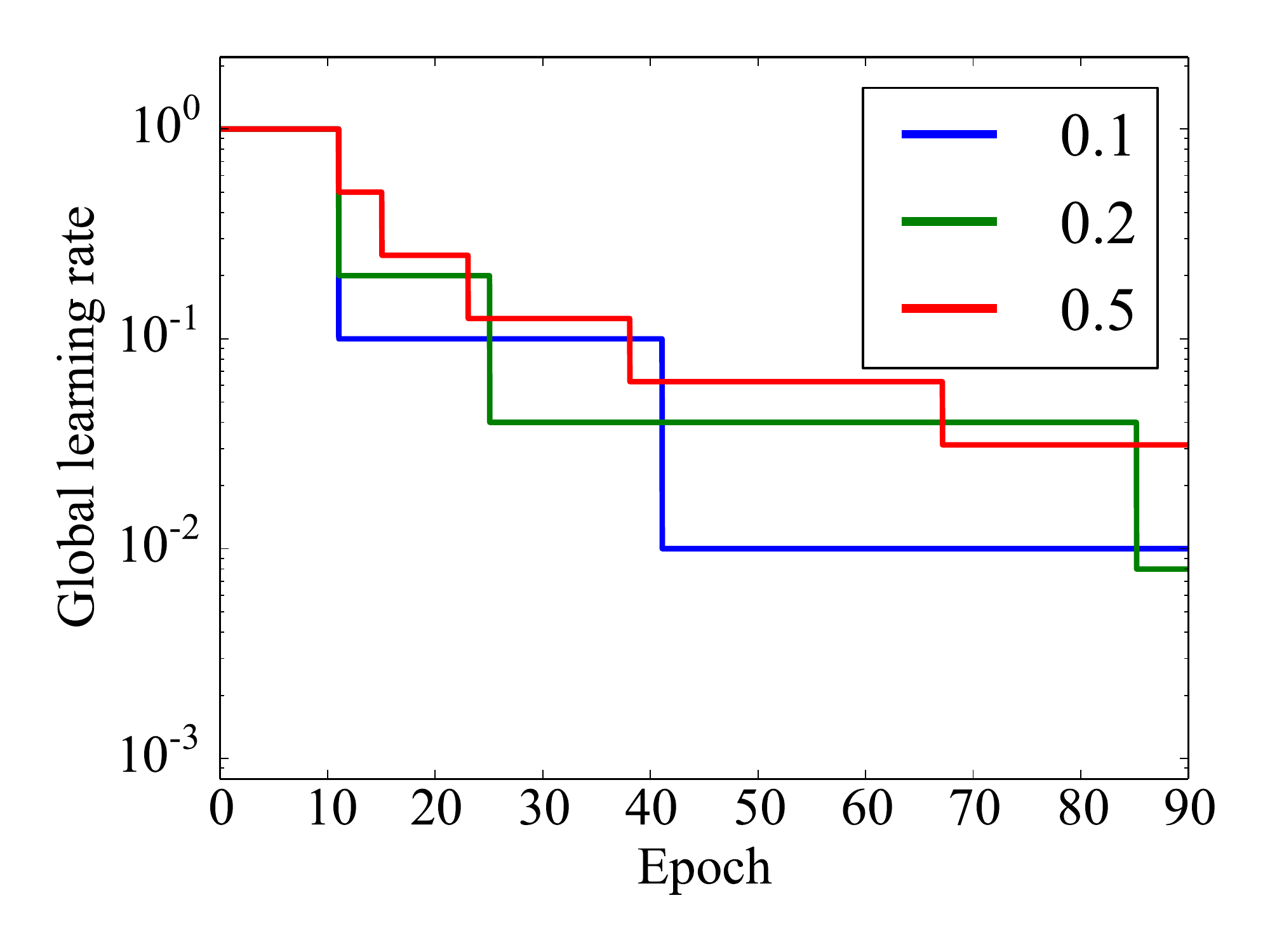}
  \end{subfigure}
  \caption{Smoothed training loss, test accuracy and learning rate schedule for ResNet18 trained on ImageNet using SASA with different values of $\zeta$. SASA automatically adapts the drop frequency.
  }
  \label{fig:changing-zeta}
  \vspace{-1ex}
\end{figure}

\begin{figure}[t]
  \centering
  \begin{subfigure}[t]{.23\linewidth}
    \centering
    \includegraphics[width=\linewidth]{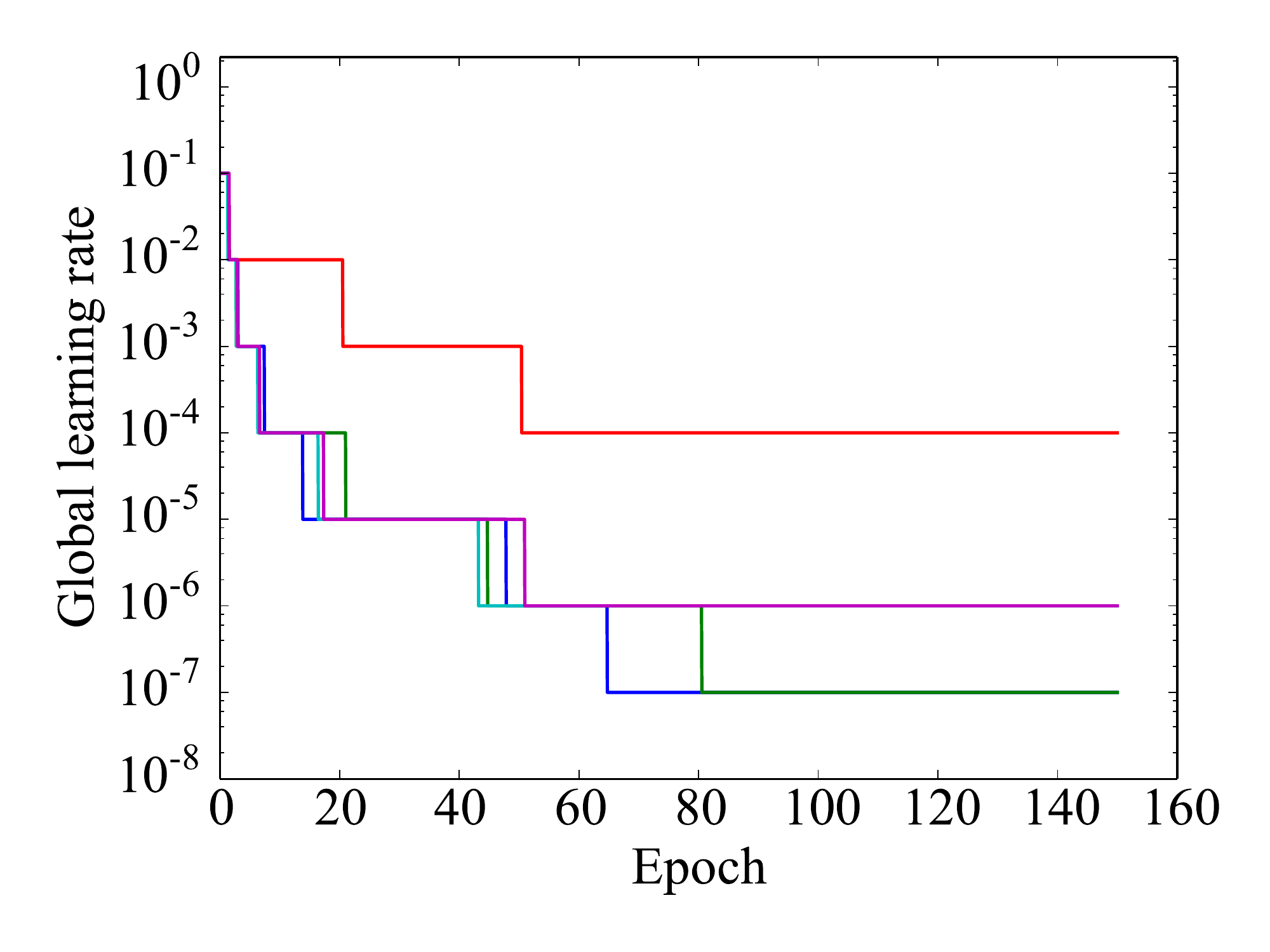}
  \end{subfigure}%
  \begin{subfigure}[t]{.23\linewidth}
    \centering
    \includegraphics[width=\linewidth]{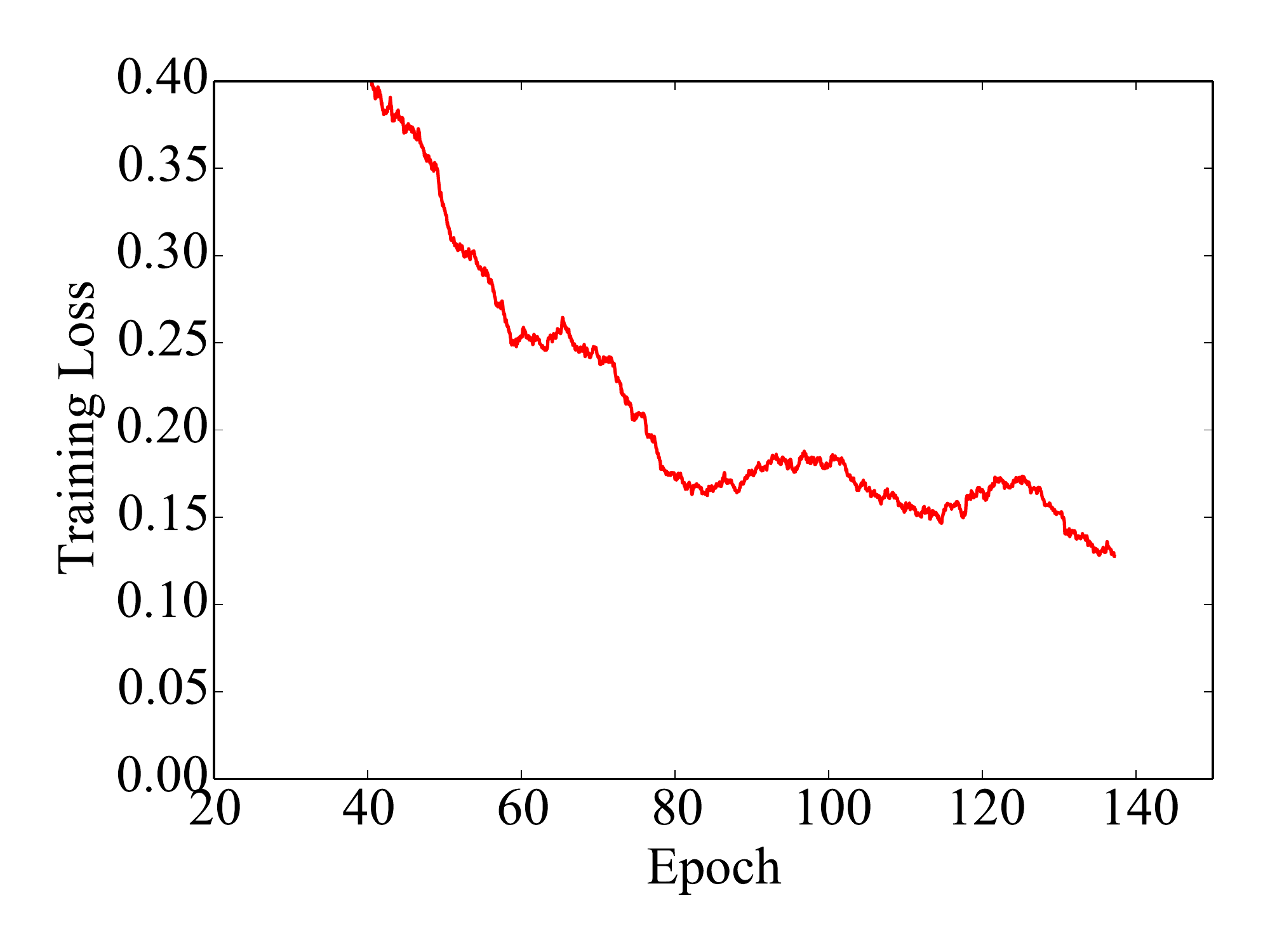}
  \end{subfigure}%
  \begin{subfigure}[t]{.23\linewidth}
    \centering
    \includegraphics[width=\linewidth]{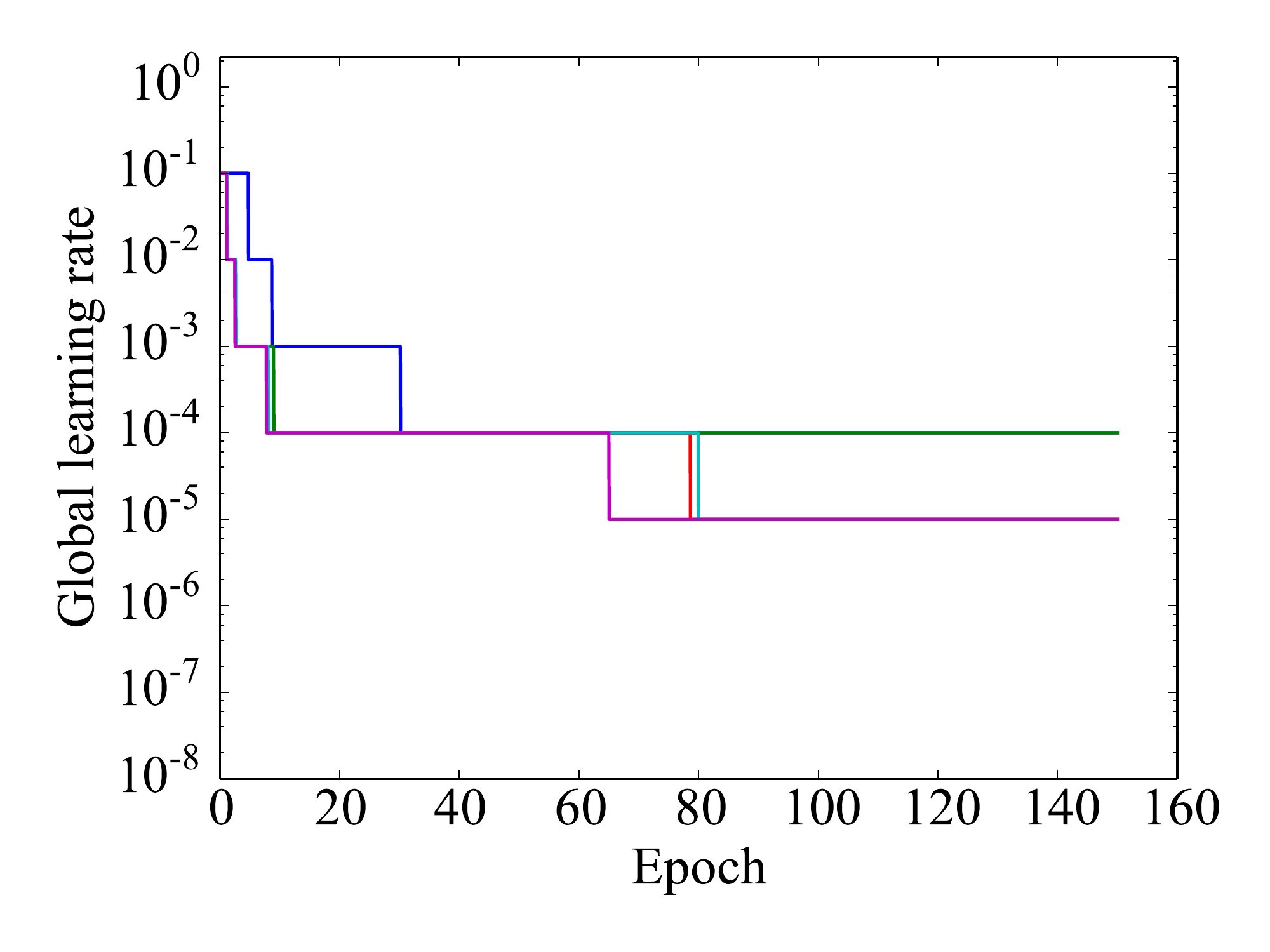}
  \end{subfigure}%
  \begin{subfigure}[t]{.23\linewidth}
    \centering
    \includegraphics[width=\linewidth]{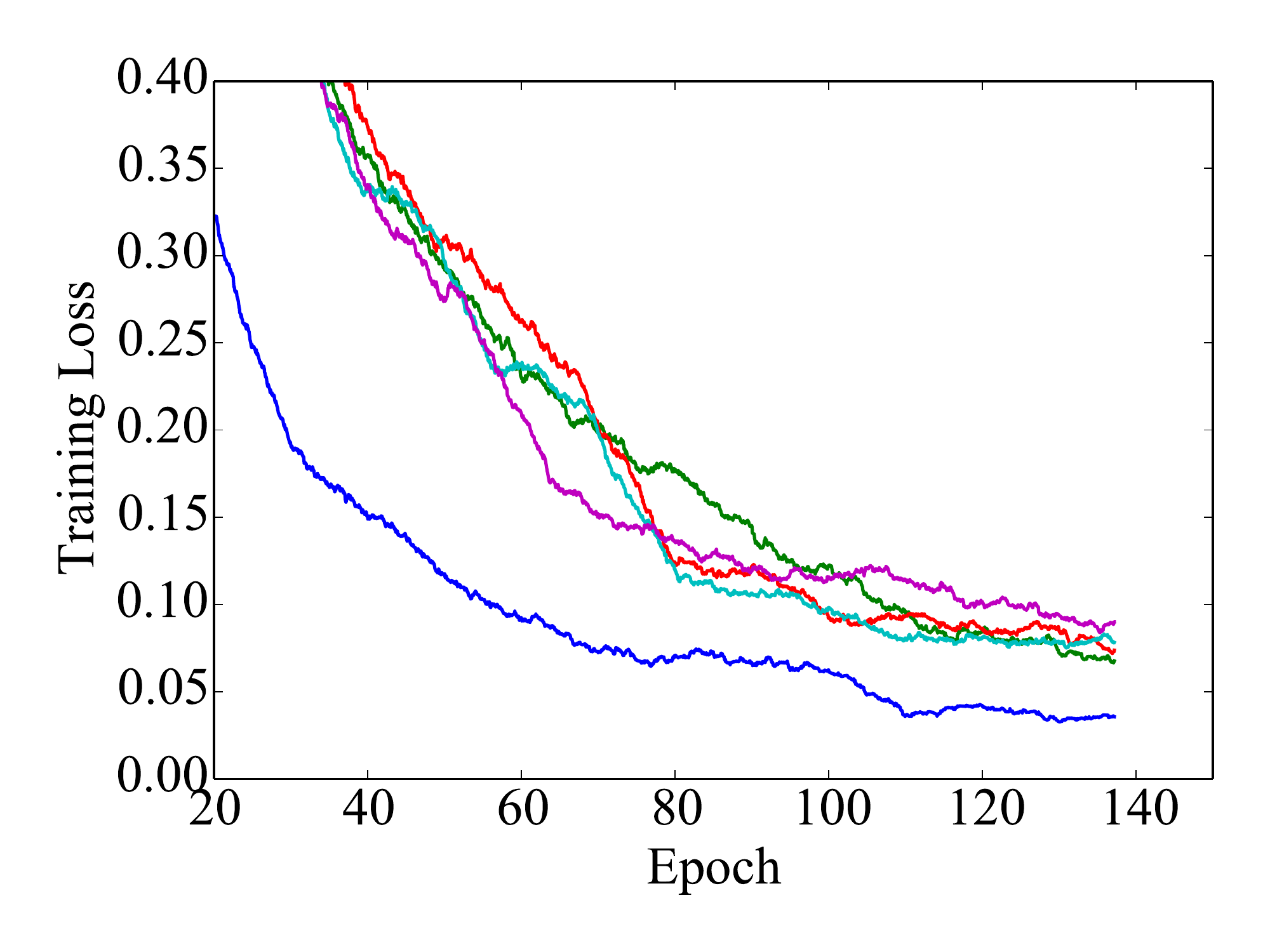}
  \end{subfigure}\\
  \begin{subfigure}[t]{.23\linewidth}
    \centering
    \includegraphics[width=\linewidth]{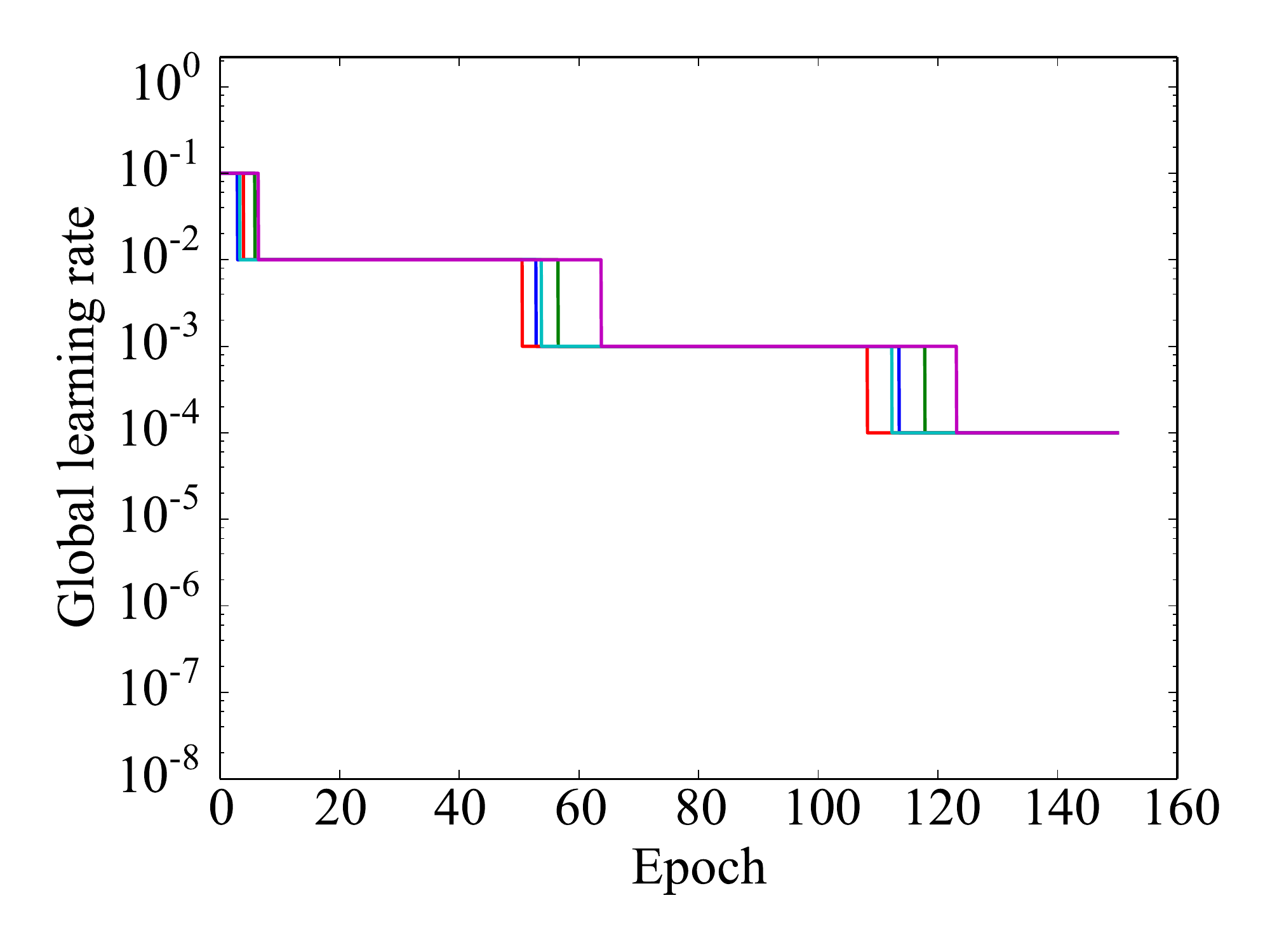}
  \end{subfigure}%
  \begin{subfigure}[t]{.23\linewidth}
    \centering
    \includegraphics[width=\linewidth]{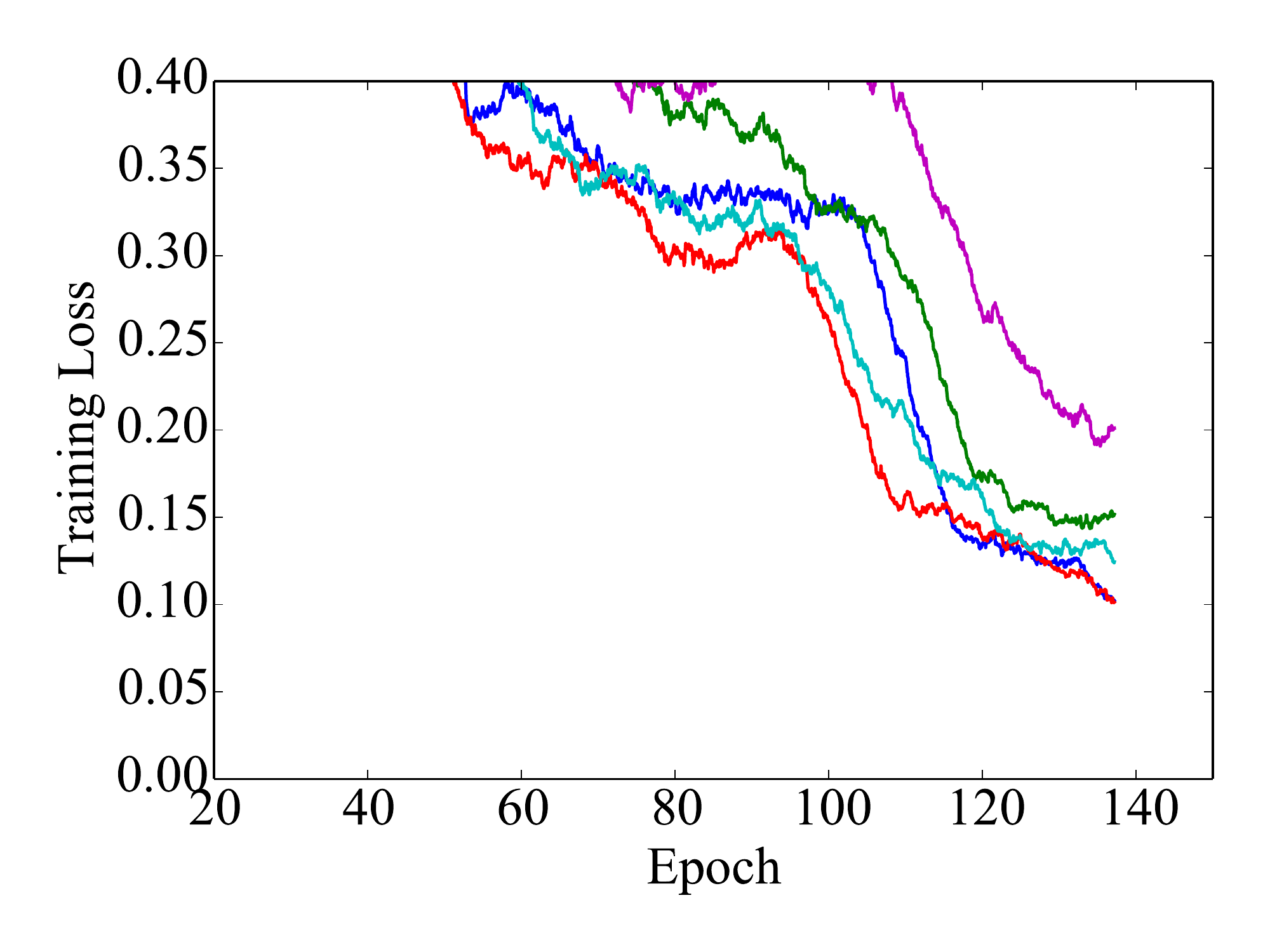}
  \end{subfigure}%
  \begin{subfigure}[t]{.23\linewidth}
    \centering
    \includegraphics[width=\linewidth]{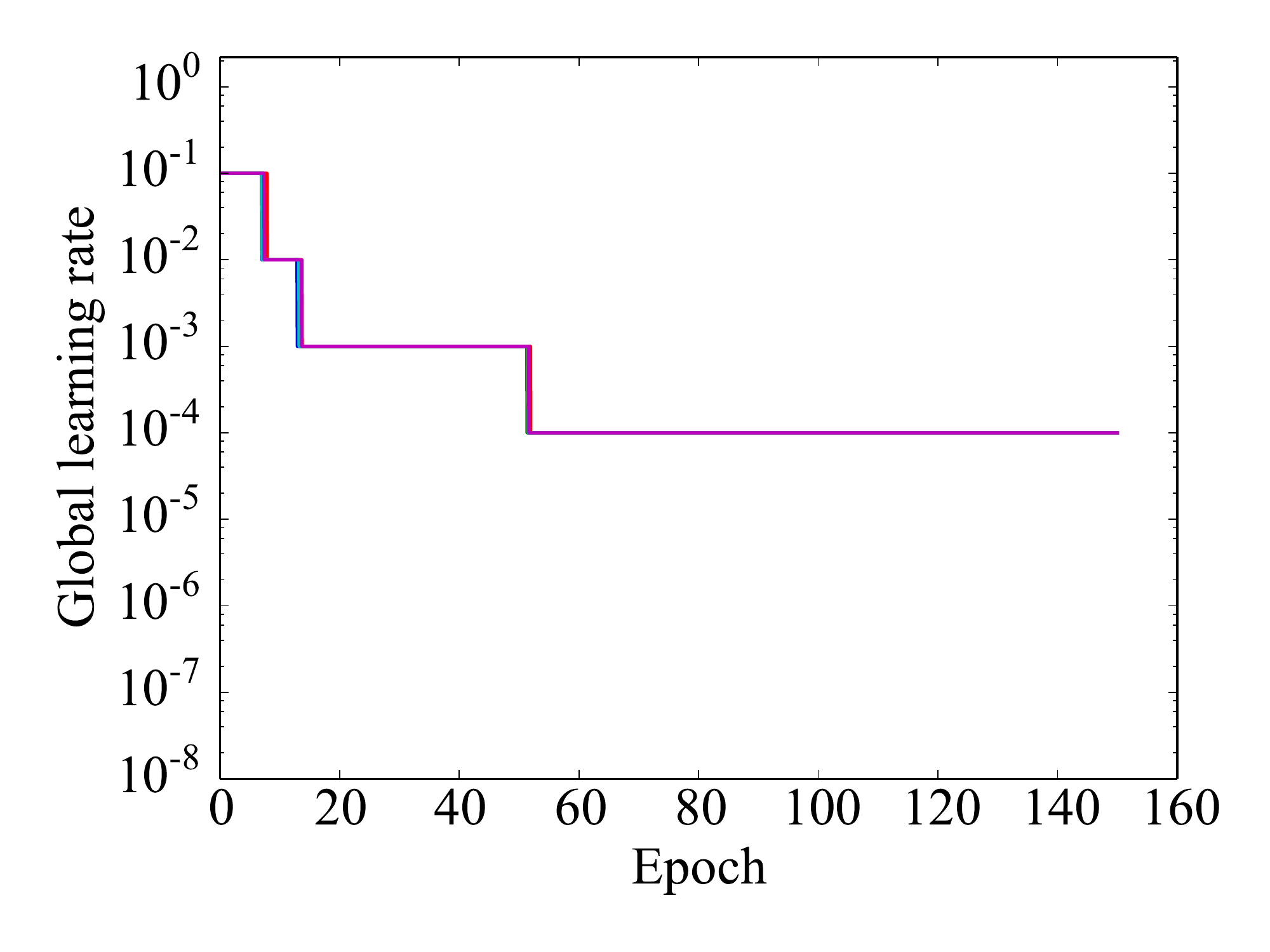}
  \end{subfigure}%
  \begin{subfigure}[t]{.23\linewidth}
    \centering
    \includegraphics[width=\linewidth]{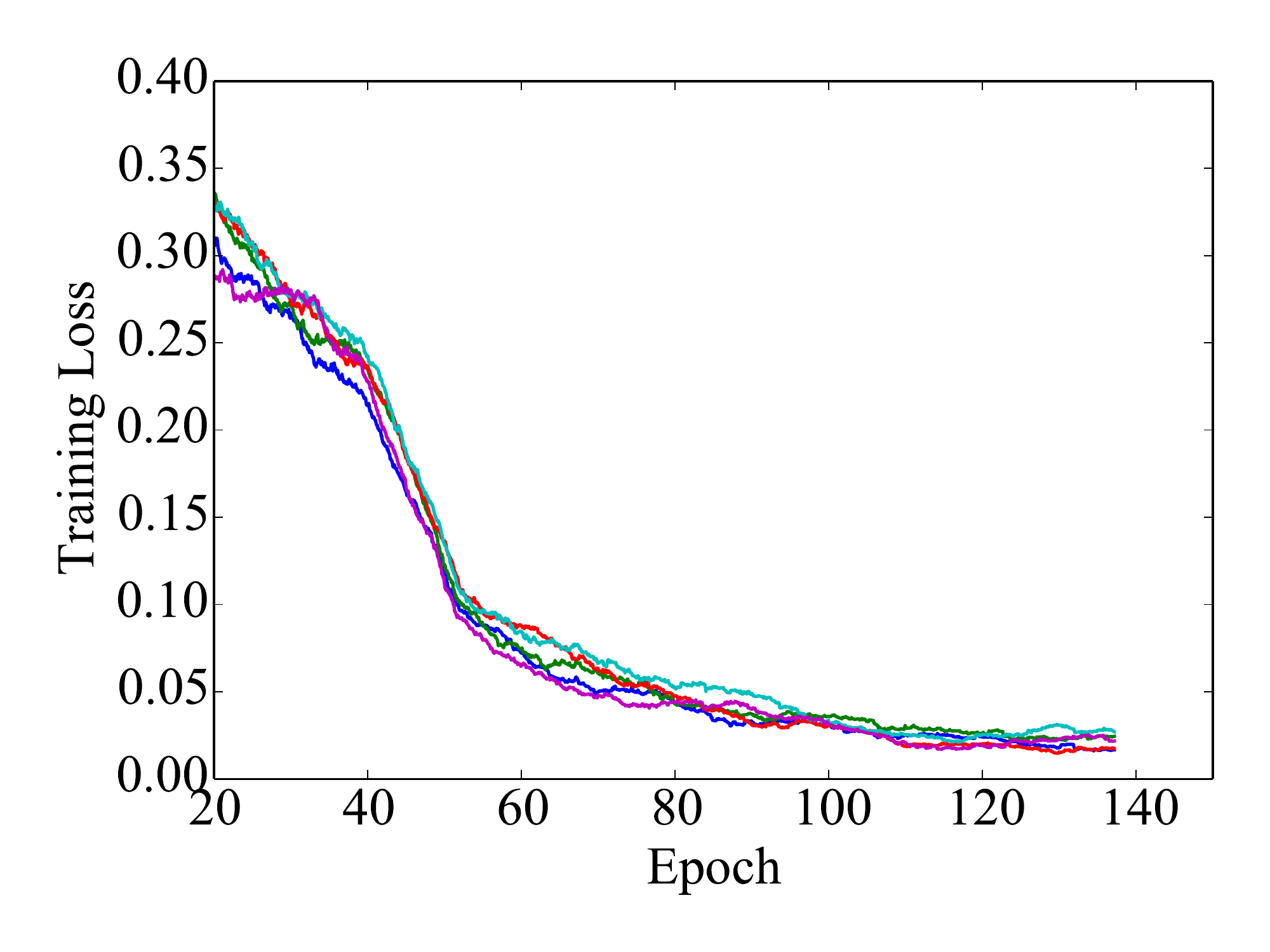}
   \end{subfigure}
  \caption{Variance in learning rate schedule and training loss for the two tests \eqref{eqn:yaida-test} (top row) and \eqref{eqn:our-test} (bottom row) with \emph{fixed} sample size $N$, across five independent runs. With the same number of samples and the same value of $\delta$ (0.02), the test \eqref{eqn:yaida-test} is much more sensitive to the level of noise.}
  \label{fig:variancefig}
  \end{figure}

\vspace{-1.5ex}
\paragraph{Adaptation to the drop factor.}
At first glance, the choice of the drop factor $\zeta$ seems critical. However, Figure \ref{fig:changing-zeta} shows that SASA automatically adapts to different values of $\zeta$. When $\zeta$ is larger, so $\alpha$ decreases slower, the dynamics converge more quickly to the stationary distribution, so the overall rate of decrease stays roughly constant across different values of $\zeta$. Aside from the different choices of $\zeta$, all other hyperparameters were the same as in the ImageNet experiment of Figure \ref{fig:results}.

\vspace{-1.5ex}
\paragraph{Variance.}
Figure \ref{fig:variancefig} shows the variance in learning rate schedule and training loss for the two tests in~\eqref{eqn:yaida-test} (top row) and~\eqref{eqn:our-test} (bottom row) with a fixed testing frequency $M=400$ iterations, across five independent runs. The model is ResNet18 
trained on CIFAR-10 using the same procedure as in the previous CIFAR experiment, but with different batch sizes. The left two columns use batch size four, and the right two use batch size eight. With the same number of samples and the same value of $\delta=0.02$, the test~\eqref{eqn:yaida-test} is much more sensitive to the level of noise in these small-batch examples. When the batch size is four, only one of the training runs using the test \eqref{eqn:yaida-test} achieves training loss on the same scale as the others. Appendix \ref{apdx:vs-yaida} contains additional discussion comparing 
these two tests.

\section{Conclusion}
We provide a theoretically grounded statistical procedure for automatically determining when to decrease the learning rate $\alpha$ in constant-and-cut methods. On the tasks we tried, SASA was competitive
with the best hand-tuned schedules for SGM, and it
came close to the performance of SGM and Adam when they were tuned using a validation set.
The statistical testing procedure controls the variance of the method and makes it more robust than other more heuristic tests.
Our experiments across several different tasks and datasets did not require any adjustment to the parameters~$\gamma$,~$\delta$, or $M$.

We believe these practical results indicate that automatic ``constant-and-cut'' algorithms
are a promising direction for future research in adaptive optimization.
We used a simple statistical test to check Yaida's stationary condition \eqref{eqn:fdr1}.
However, there may be better tests that more properly control the false discovery rate \citep{blanchard2009adaptive, lindquist2015zen}, or more sophisticated conditions that
also account for non-stationary dynamics like overfitting or limit cycles \citep{yaida2018fluctuation}.
Such techniques could make the SASA approach more broadly useful.

\bibliography{sasa}
\bibliographystyle{plainnat}

\clearpage
\appendix
\section{Details and Additional Experiments}
\label{apdx:exps}

\begin{figure}[tb]
    \centering
  \begin{subfigure}[t]{.245\linewidth}
    \centering
    \includegraphics[width=\linewidth]{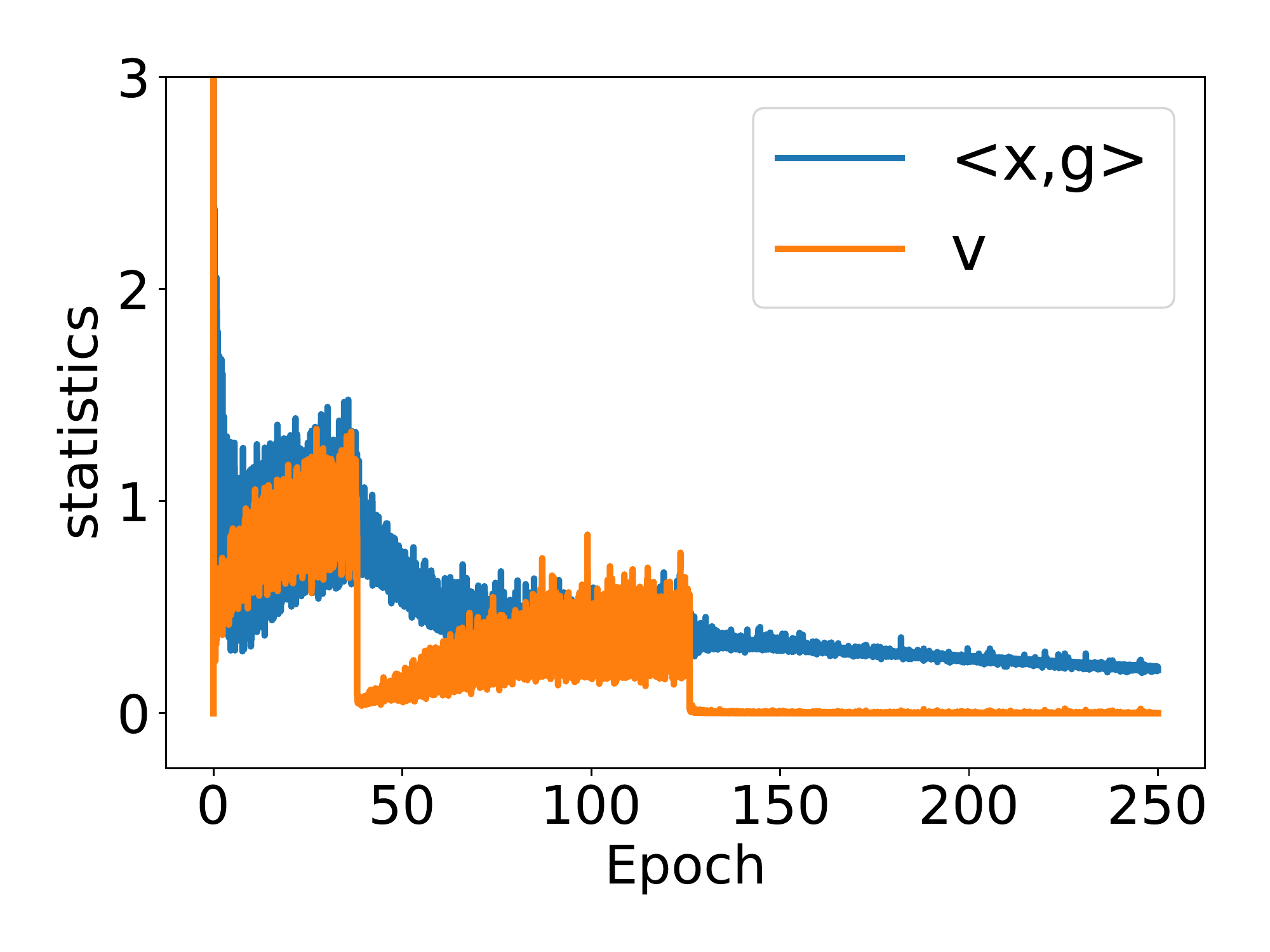}
    \subcaption{}
  \end{subfigure}%
  \begin{subfigure}[t]{.245\linewidth}
    \centering
    \includegraphics[width=\linewidth]{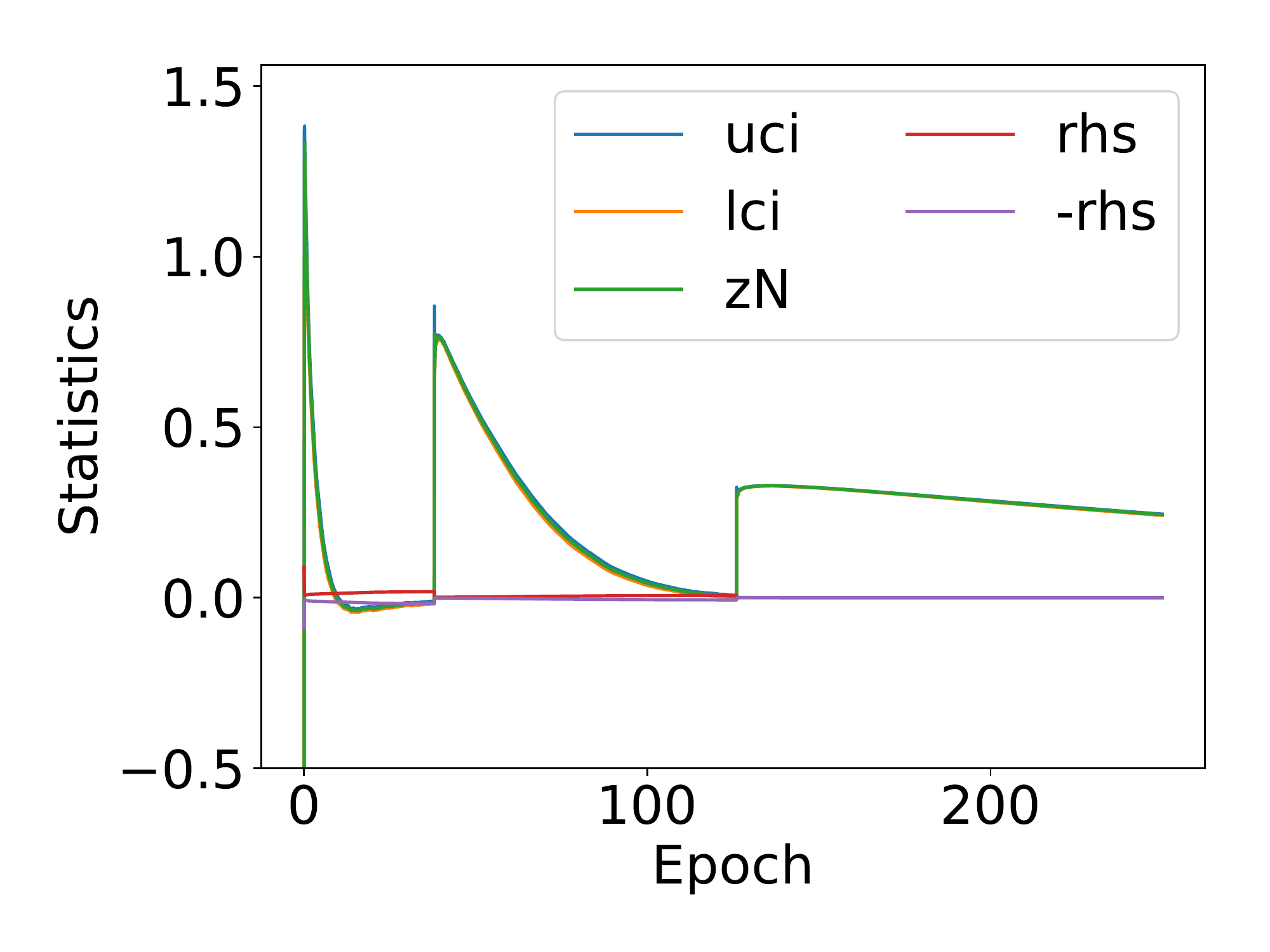}
    \subcaption{}
  \end{subfigure}%
  \begin{subfigure}[t]{.245\linewidth}
    \centering
    \includegraphics[width=\linewidth]{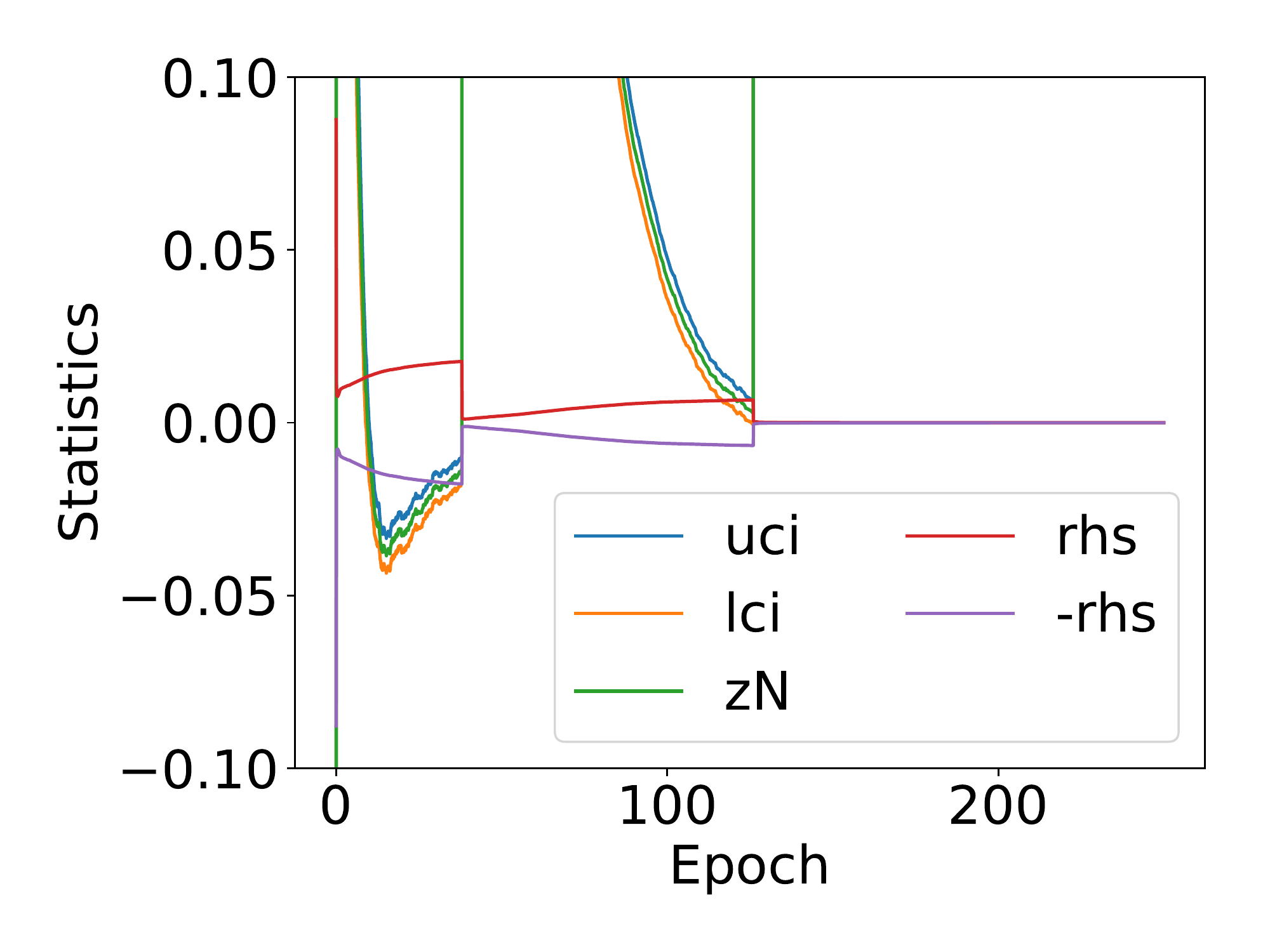}
    \subcaption{}
  \end{subfigure}%
  \begin{subfigure}[t]{.245\linewidth}
    \centering
    \includegraphics[width=\linewidth]{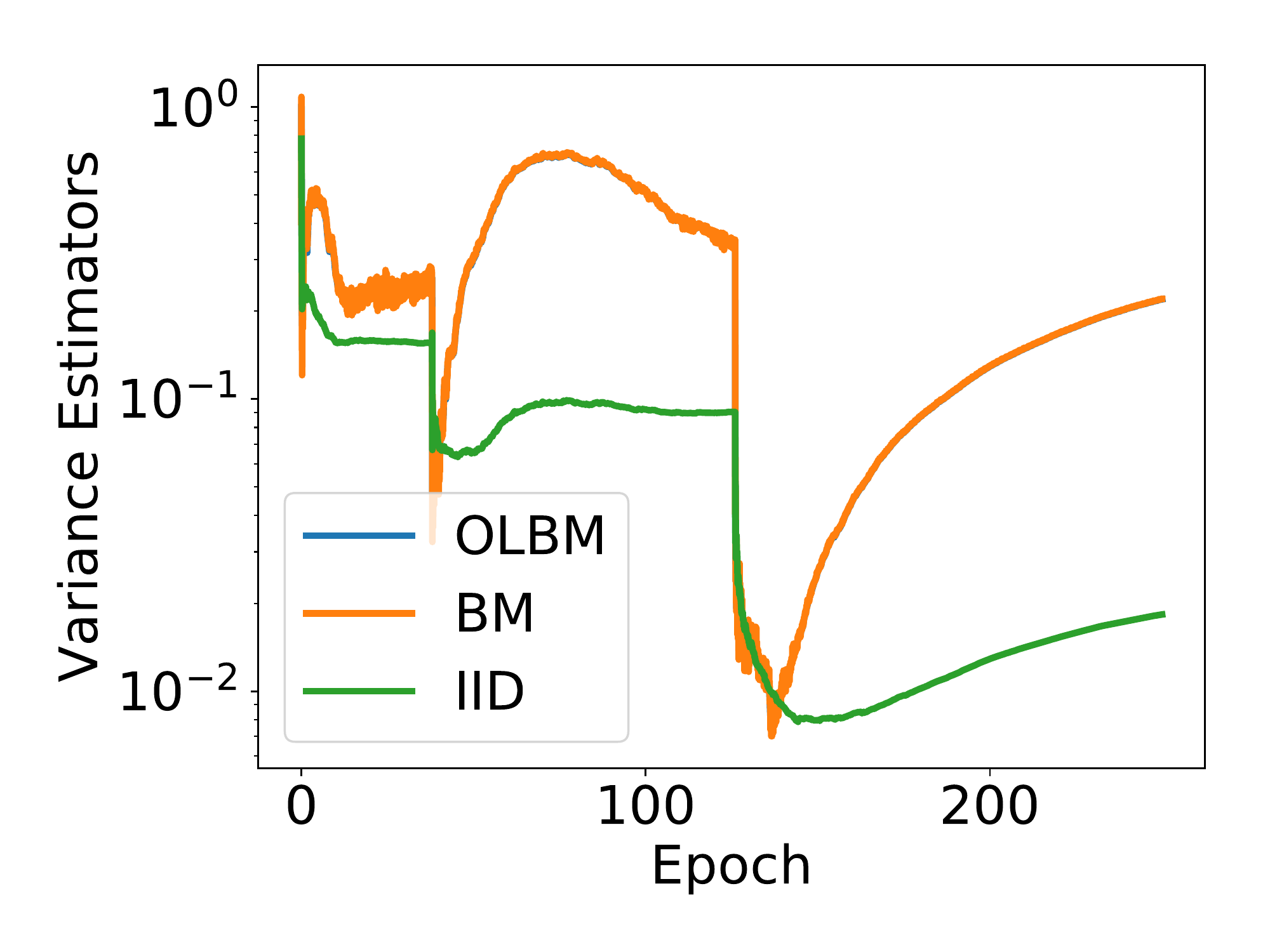}
    \subcaption{}
  \end{subfigure}
  \caption{Evolution of the different statistics for SASA over the course of training ResNet18 on CIFAR-10 using the default parameters $\delta=0.02, \gamma=0.2, \zeta=0.1$. Panel (a) shows the raw data for both sides of condition \eqref{eqn:fdr1}. That is, it shows the values of $\langle \xk, \gk\rangle$ and $\frac{\alpha}{2}\frac{1+\beta}{1-\beta}\langle \dk, \dk\rangle$ at each iteration. Panel (1) shows $\bar{z}_N$ with its lower and upper confidence interval $[lci, uci]$ and the "right hand side" (rhs) $(-\delta \bar{v}_N, \delta\bar{v}_N)$ (see Eqn.~\eqref{eqn:our-test}). Panel (c) shows a zoomed-in version of (b) to show the drop points in more detail. Panel (d) depicts the different variance estimators (i.i.d., batch means, overlapping batch means) over the course of training. The i.i.d. variance (green) is a poor estimate of $\sigma_z^2$.}
  \label{fig:cifar_stats}
\end{figure}

\begin{figure}[tb]
    \centering
  \begin{subfigure}[t]{.33\linewidth}
    \centering
    \includegraphics[width=\linewidth]{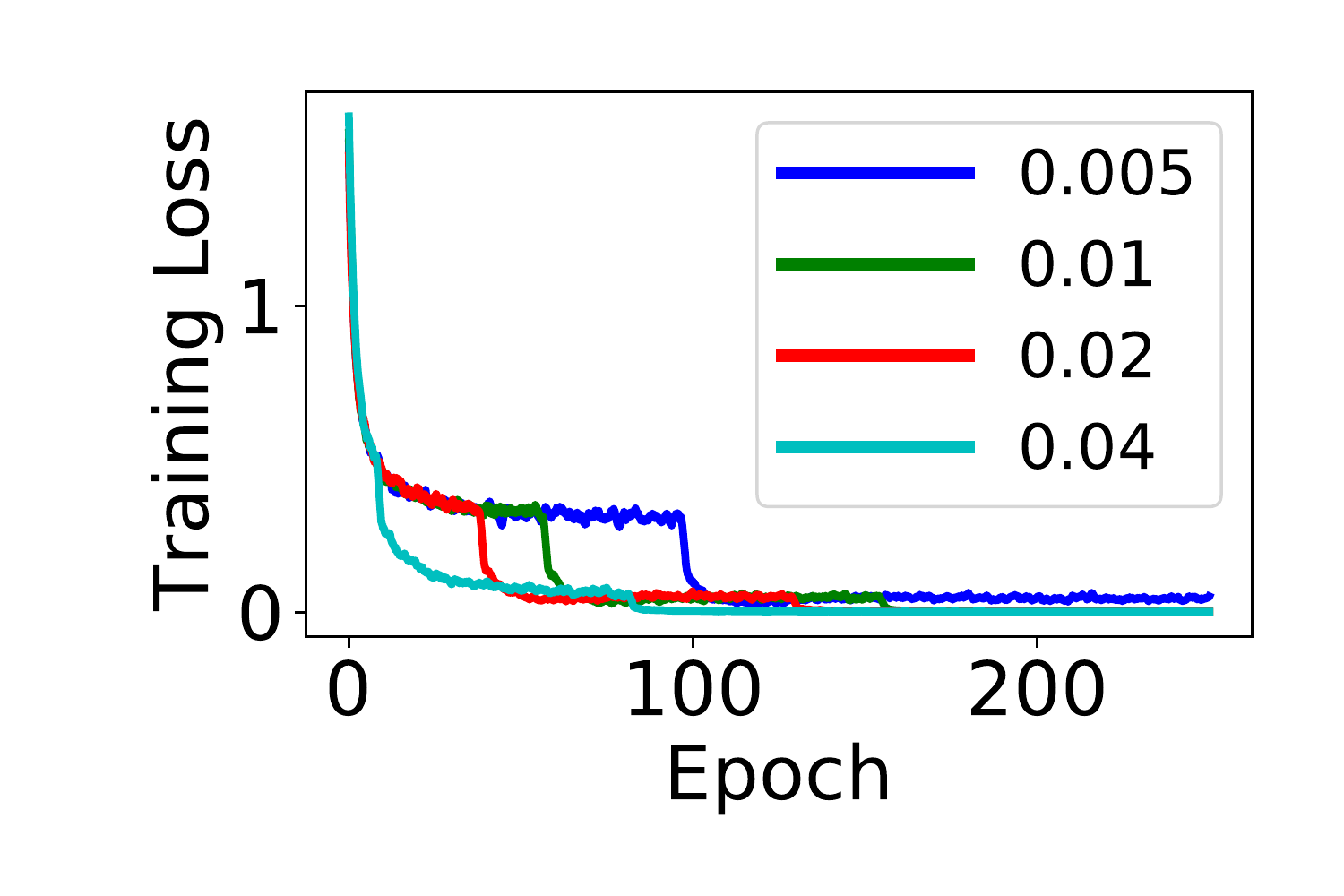}
  \end{subfigure}%
  \begin{subfigure}[t]{.33\linewidth}
    \centering
    \includegraphics[width=\linewidth]{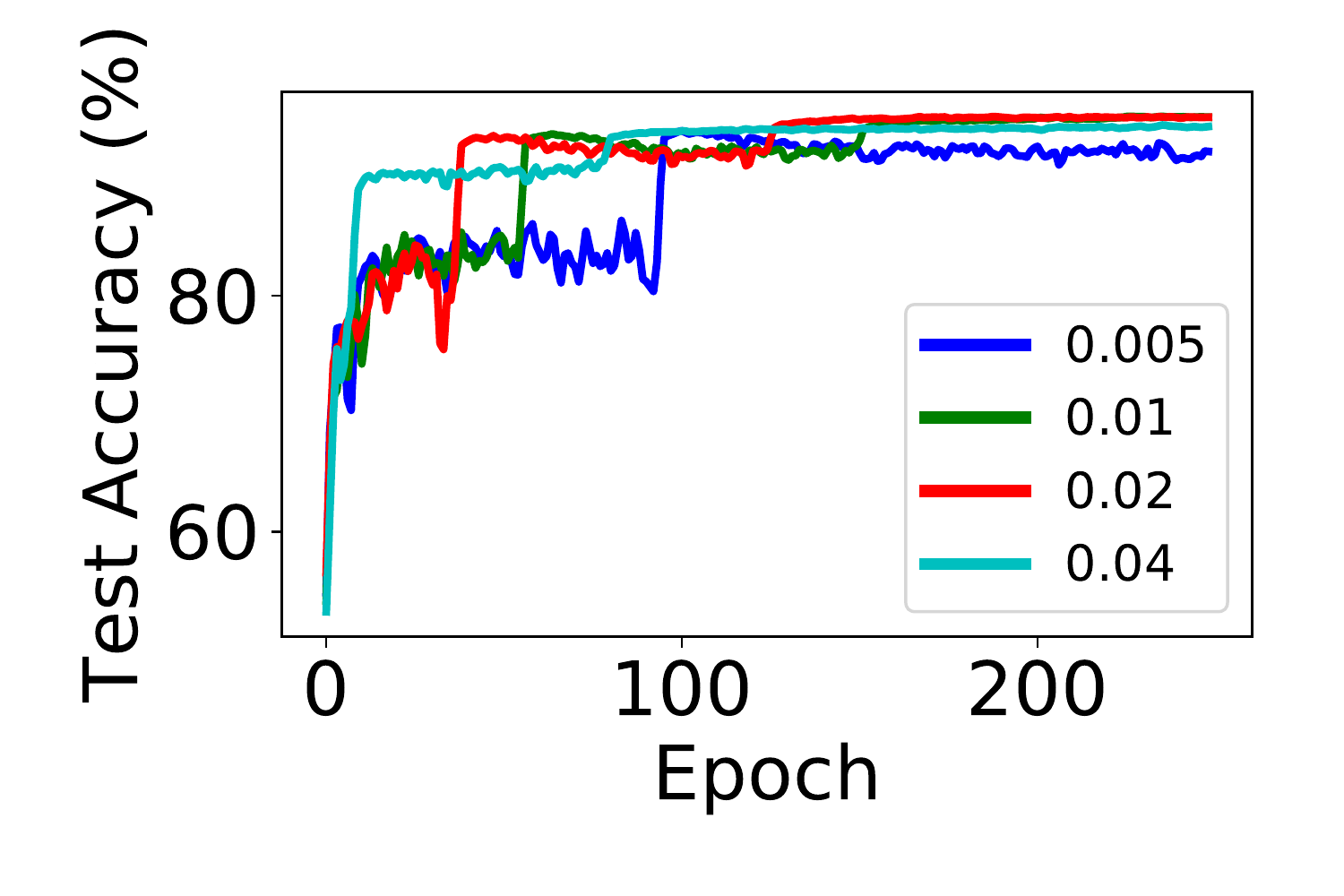}
  \end{subfigure}%
  \begin{subfigure}[t]{.33\linewidth}
    \centering
    \includegraphics[width=\linewidth]{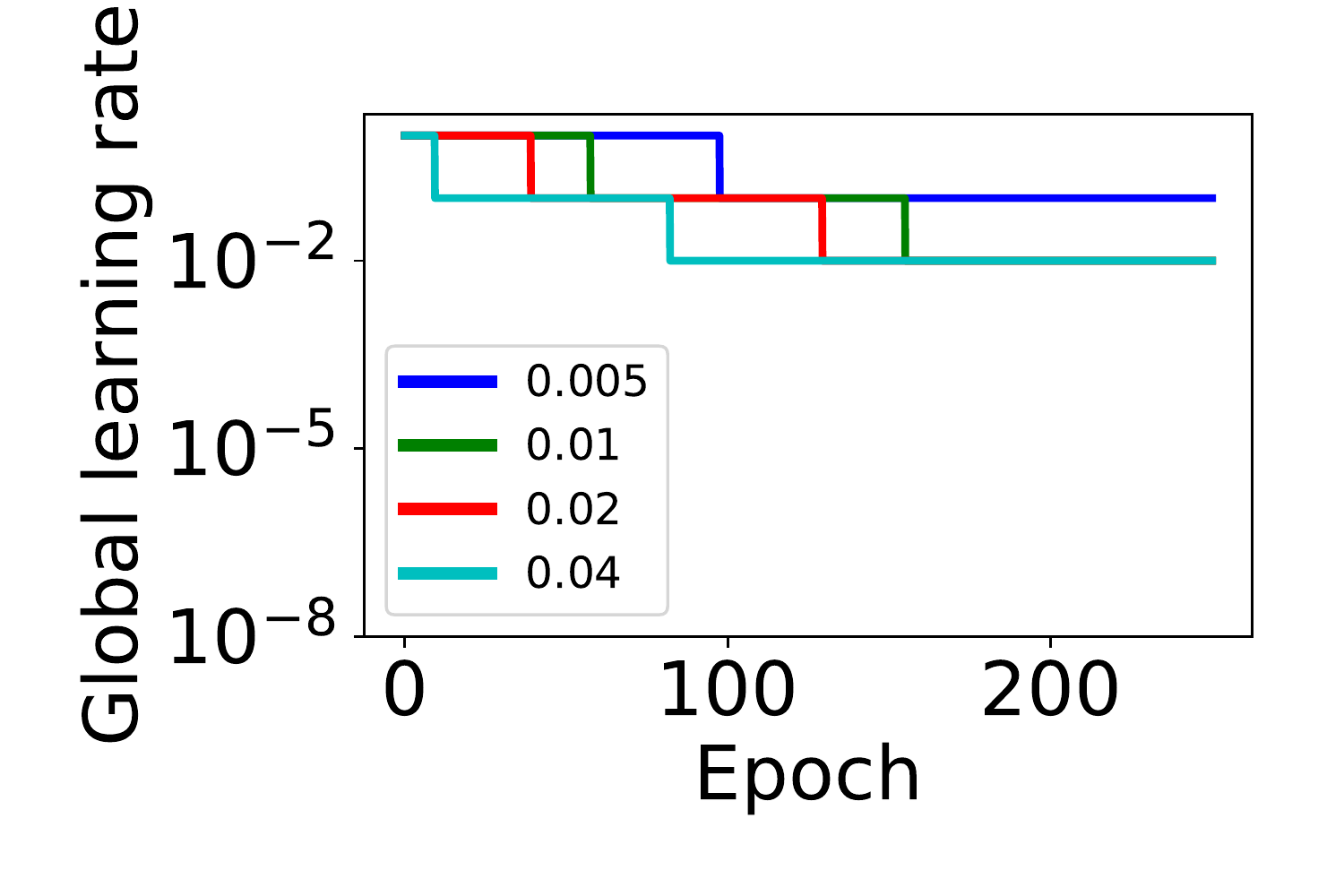}
  \end{subfigure}\\
  \begin{subfigure}[t]{.33\linewidth}
    \centering
    \includegraphics[width=\linewidth]{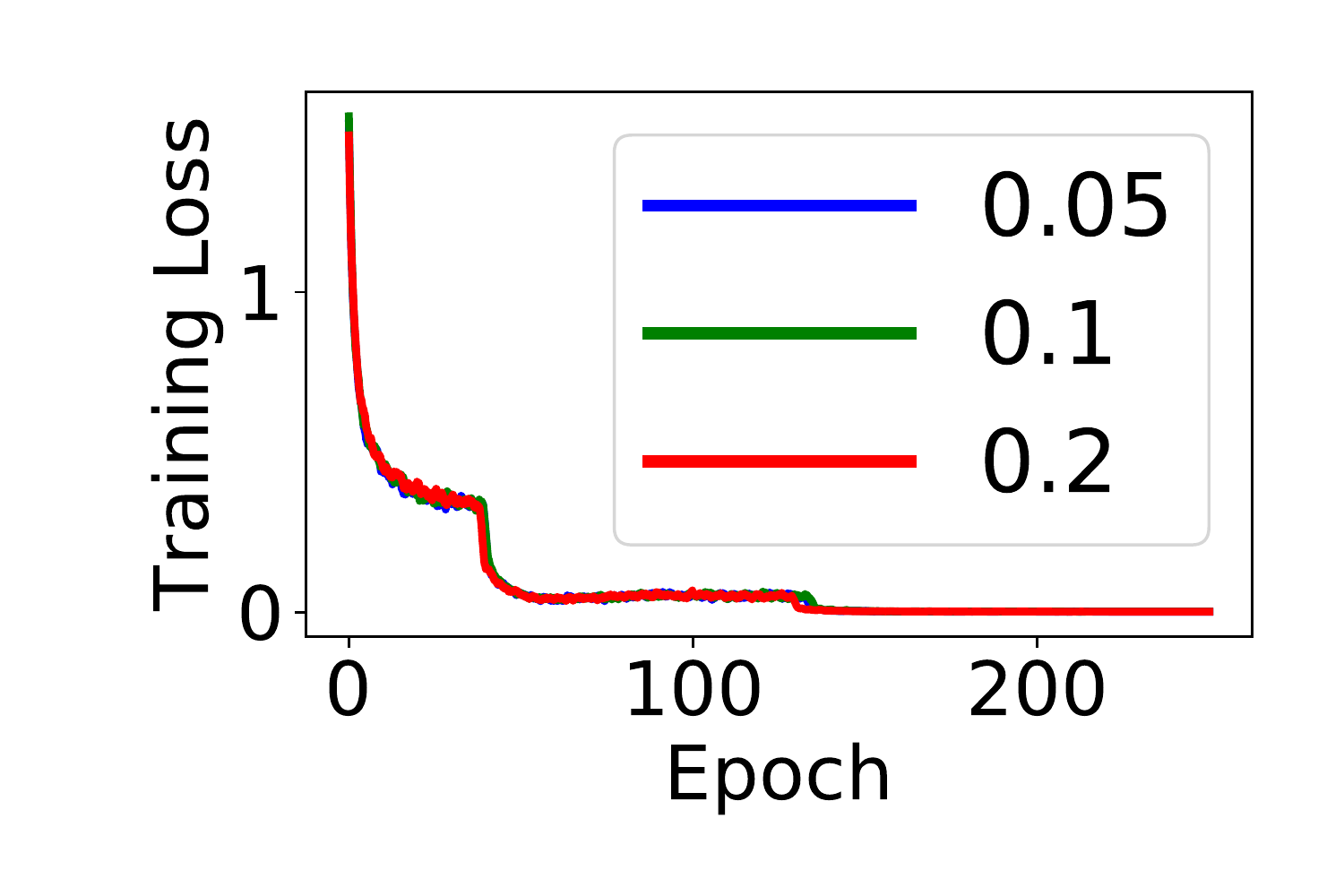}
  \end{subfigure}%
  \begin{subfigure}[t]{.33\linewidth}
    \centering
    \includegraphics[width=\linewidth]{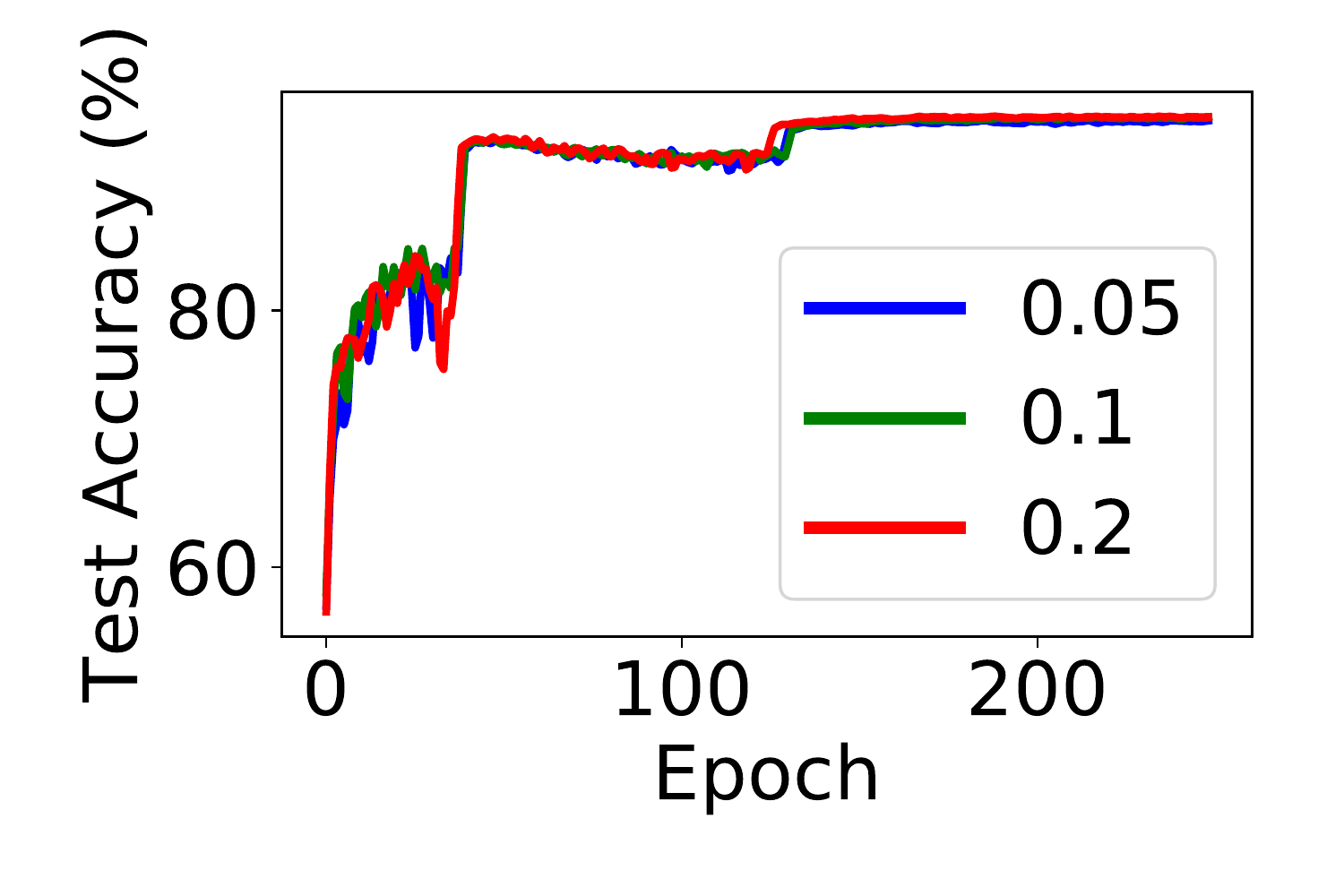}
  \end{subfigure}%
  \begin{subfigure}[t]{.33\linewidth}
    \centering
    \includegraphics[width=\linewidth]{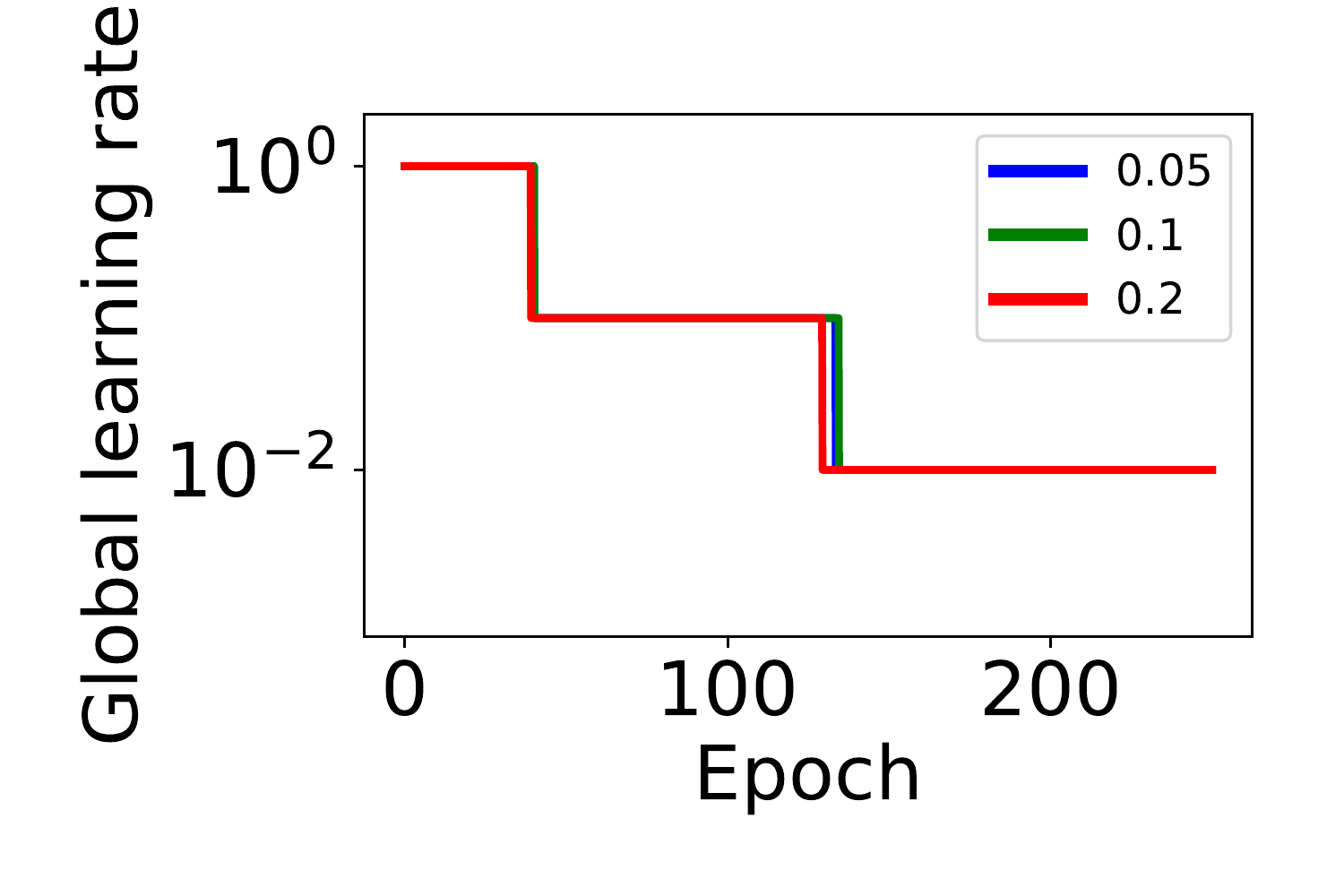}
  \end{subfigure}  \\
  \begin{subfigure}[t]{.33\linewidth}
    \centering
    \includegraphics[width=\linewidth]{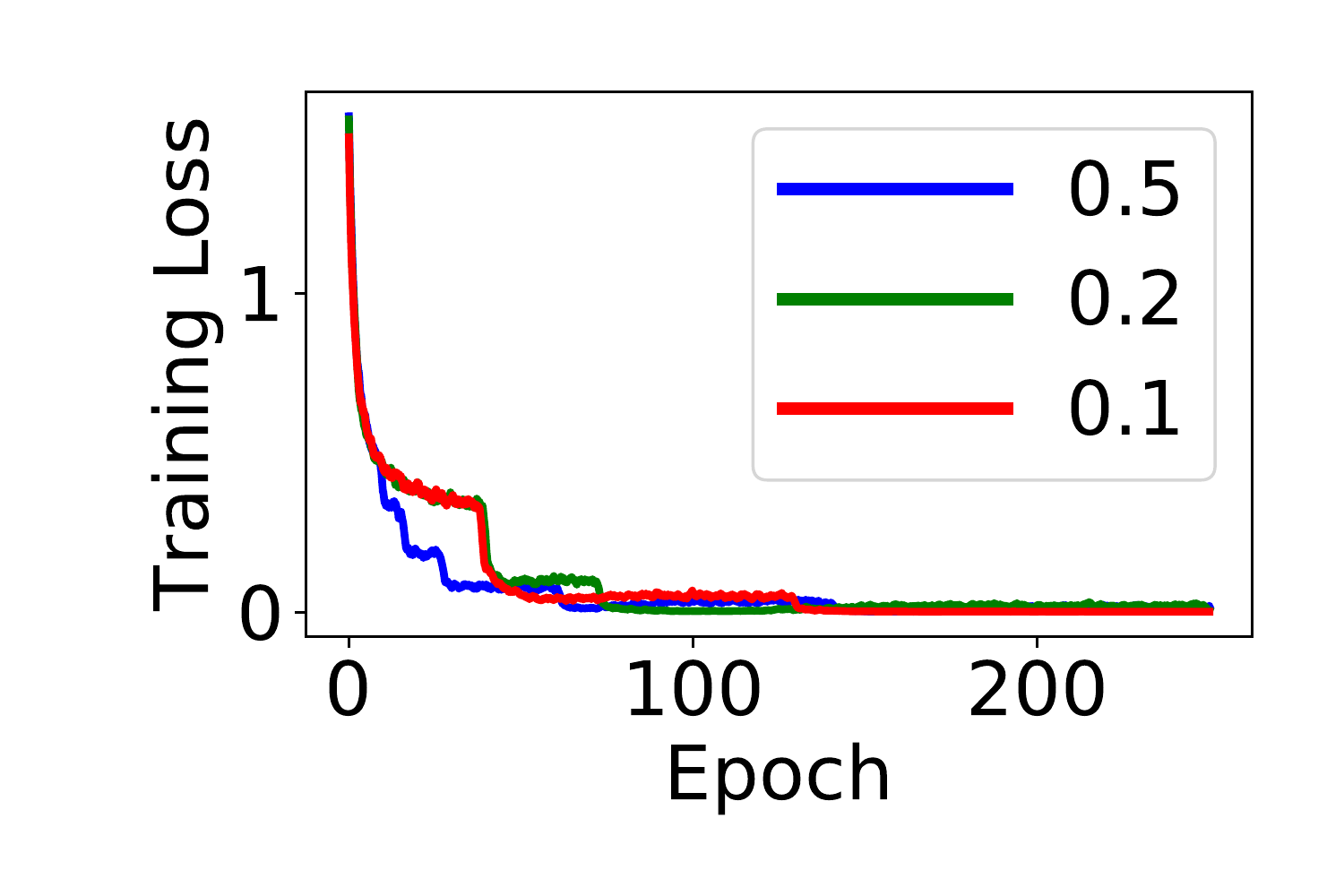}
  \end{subfigure}%
  \begin{subfigure}[t]{.33\linewidth}
    \centering
    \includegraphics[width=\linewidth]{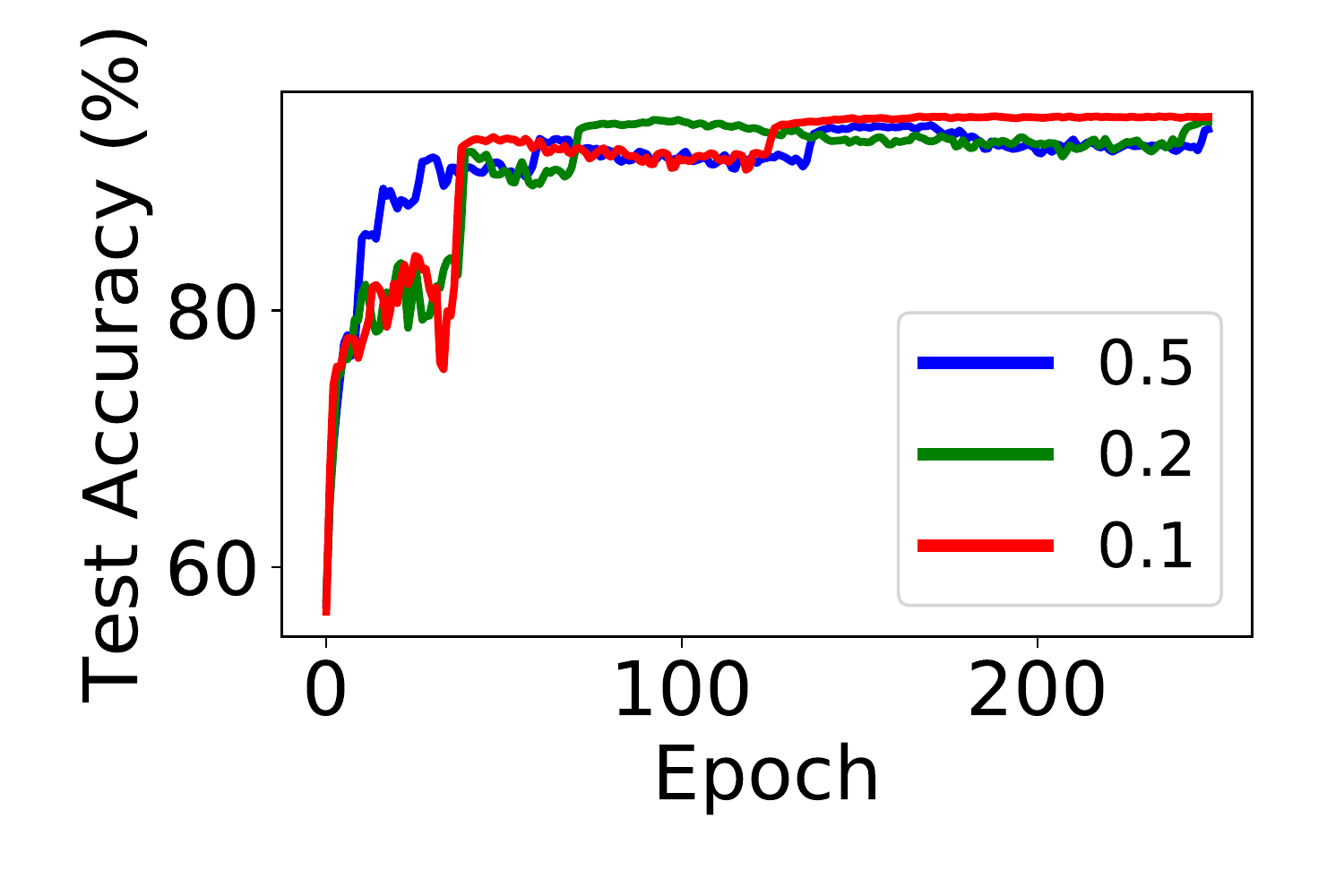}
  \end{subfigure}%
  \begin{subfigure}[t]{.33\linewidth}
    \centering
    \includegraphics[width=\linewidth]{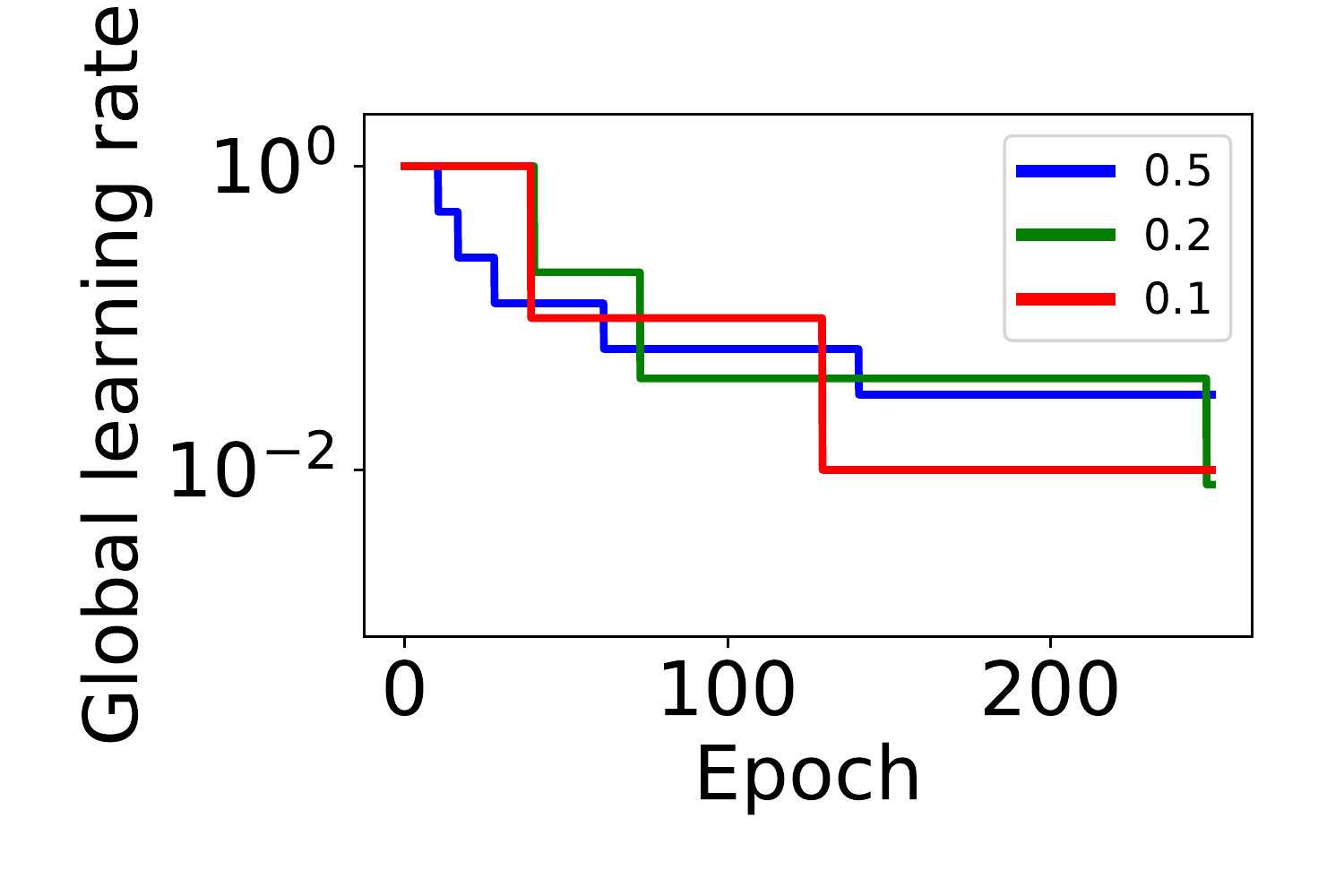}
  \end{subfigure}  
  \caption{Training loss, test accuracy, and learning rate schedule for SASA using different values of $\gamma$, $\delta$ and $\zeta$ around the defaults $0.2$, $0.02$ and $0.1$. The model is ResNet18 trained on CIFAR-10, with the procedure the same as in Section \ref{sec:experiments}. Top row: performance for fixed $\gamma=0.2, \zeta=0.1$, and $\delta \in \{0.005, 0.01, 0.02, 0.04\}$. Middle row: performance for fixed $\delta=0.02, \zeta=0.1$, and $\gamma \in \{0.05, 0.1, 0.2\}$. Bottom row: performance for fixed $\gamma=0.2, \delta=0.02$, and $\zeta \in \{0.5, 0.2, 0.1\}$. Qualitatively, increasing $\delta$ and increasing $\gamma$ both cause the algorithm to drop sooner. The value of $\zeta$ does not influence the final performance, as long as the learning rate finally decays to the same level.}
  \label{fig:cifarsens}
 \end{figure}

In this section, we provide more experimental details and discussion, and examine the sensitivity of SASA's performance with respect to its parameters. 

\subsection{CIFAR-10 experiments in Section \ref{sec:experiments}}
For the CIFAR-10 experiment, we set SGM to use $\alpha_0 = 1.0$ and $\zeta=0.1$, and drop every 50 epochs. We used the same values of $\alpha_0$ and $\zeta$ for SASA. For Adam, we used a warmup phase of 50 epochs with $\alpha=1.0$, and then set $\alpha_0=0.0001$, the optimal value in our grid search of $\{0.00001, 0.0001, 0.01, 0.1\}$. The weight decay parameter for all models was set to 0.0005. All methods used batch size 128.

\paragraph{Evolution of statistics.} Figure \ref{fig:cifar_stats} shows the evolution of SASA's different statistics over the course of training the ResNet18 model on CIFAR-10 using the default parameter settings $\delta=0.02, \gamma=0.2, \zeta=0.1$. In each phase, the running average of the difference between the statistics, $\bar{z}_N$, decays toward zero. The learning rate $\alpha$ drops once $\bar{z}_N$ and its confidence interval are contained in $(-\delta \bar{v}_N, \delta \bar{v}_N)$; see Eqn~\eqref{eqn:our-test}. After the drop, the statistics increase in value and enter another phase of convergence. The batch means variance estimator (BM) and overlapping batch means variance estimator (OLBM) give very similar estimates of the variance, while the i.i.d. variance estimator, as expected, gives quite different values.

\paragraph{Sensitivity analysis.} We perturb the relative equivalence threshold $\delta$, the confidence level $\gamma$ and the decay rate $\zeta$ around their default values $(0.2, 0.02, 0.1)$ and repeat the CIFAR-10 experiment from the previous section, using the same values for the other hyperparameters. In Figure~\ref{fig:cifarsens}, the top row shows the performance for fixed $(\gamma, \zeta)=(0.2, 0.1)$ and changing $\delta$. The middle row shows the performance for fixed $(\delta, \zeta)=(0.02, 0.1)$ and changing $\gamma$. The bottom row shows the performance for fixed $(\delta, \gamma)=(0.02, 0.2)$ and changing $\zeta$. Increases in both $\delta$ and $\gamma$ tend to cause the algorithm to drop sooner; this behavior is intuitive from the testing procedure \eqref{eqn:our-test}. For values of the parameters close to the defaults, SASA still obtains good performance. The value of $\zeta$ does not influence the final performance, as long as the learning rate finally decays to the same level.

\subsection{ImageNet experiments in Section \ref{sec:experiments}}
For the ImageNet experiment, we again used $\alpha_0=1.0$ and $\zeta=0.1$ for SGM and SASA, and dropped the SGM learning rate every 30 epochs. We let Adam have a warmup phase of 30 epochs, initializing it with the parameters obtained from running SGM with $\alpha=1.0$. After this phase, we used $\alpha_0=0.0001$, the optimal  value from a grid $\{0.00001, 0.0001, 0.001, 0.01\}$. The weight decay for all models was set to 0.0001. All methods used batch size 256.

\begin{figure}[tb]
    \centering
  \begin{subfigure}[t]{.245\linewidth}
    \centering
    \includegraphics[width=\linewidth]{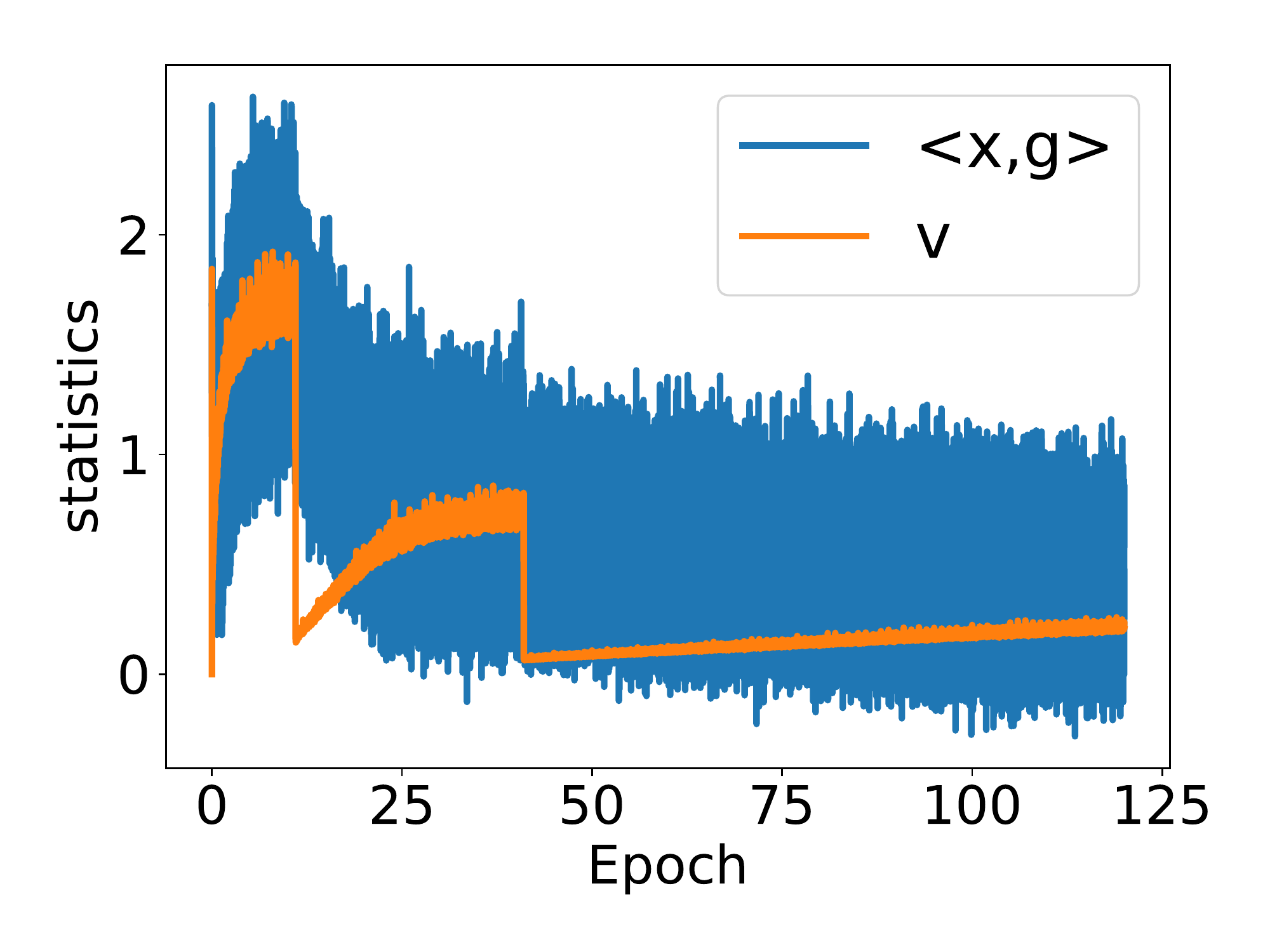}
    \subcaption{}
  \end{subfigure}%
  \begin{subfigure}[t]{.245\linewidth}
    \centering
    \includegraphics[width=\linewidth]{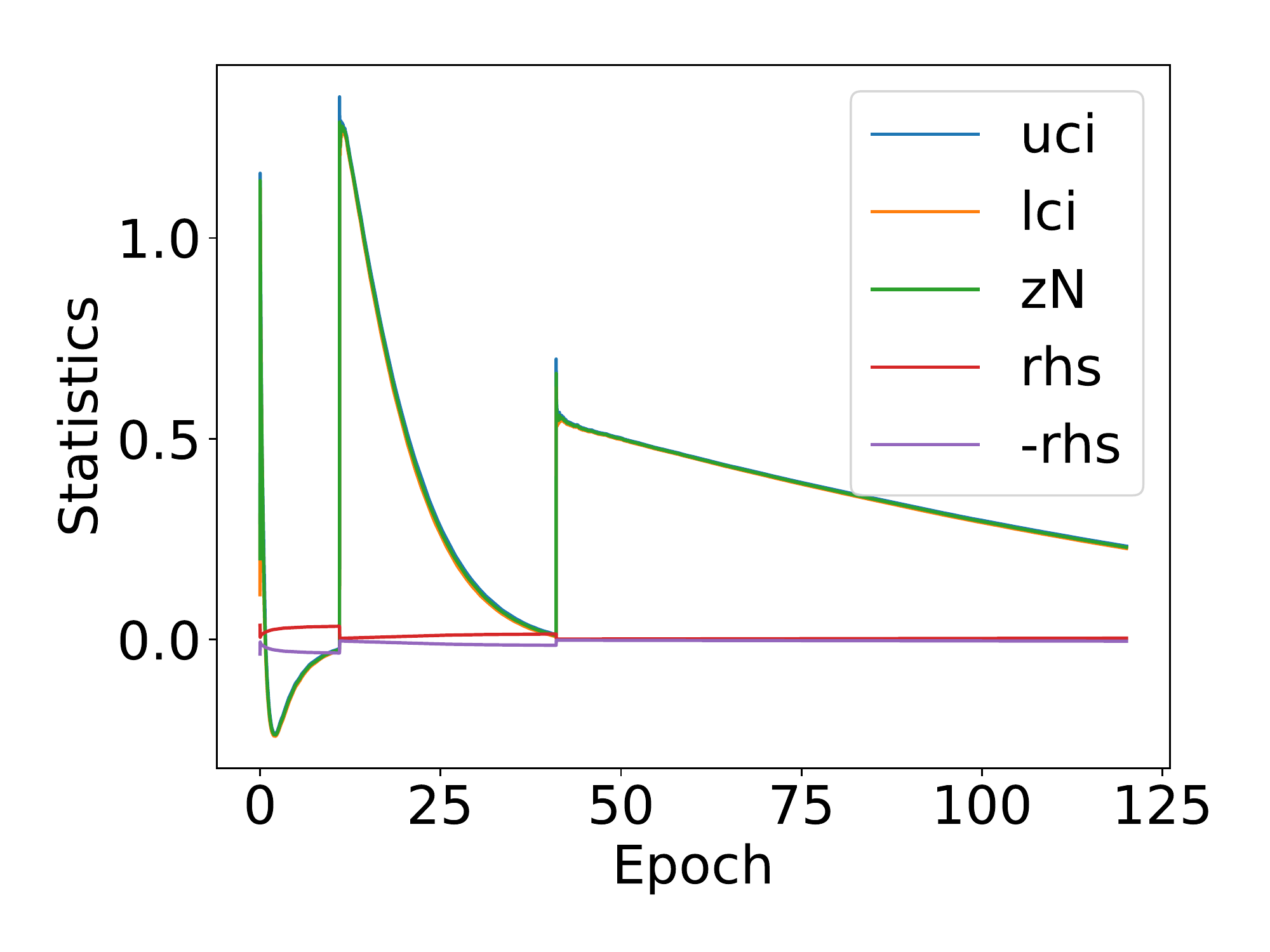}
    \subcaption{}
  \end{subfigure}%
  \begin{subfigure}[t]{.245\linewidth}
    \centering
    \includegraphics[width=\linewidth]{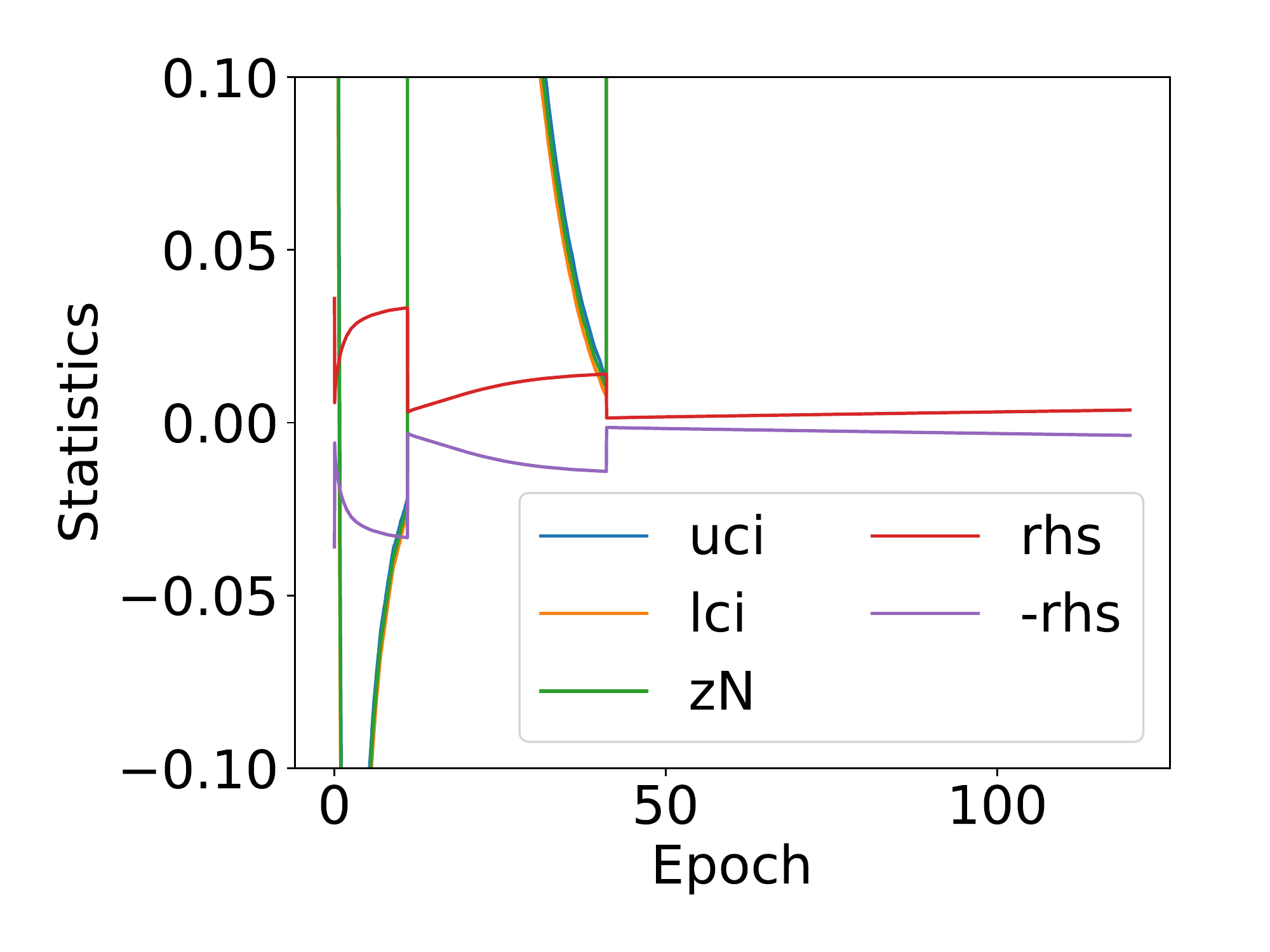}
    \subcaption{}
  \end{subfigure}%
  \begin{subfigure}[t]{.245\linewidth}
    \centering
    \includegraphics[width=\linewidth]{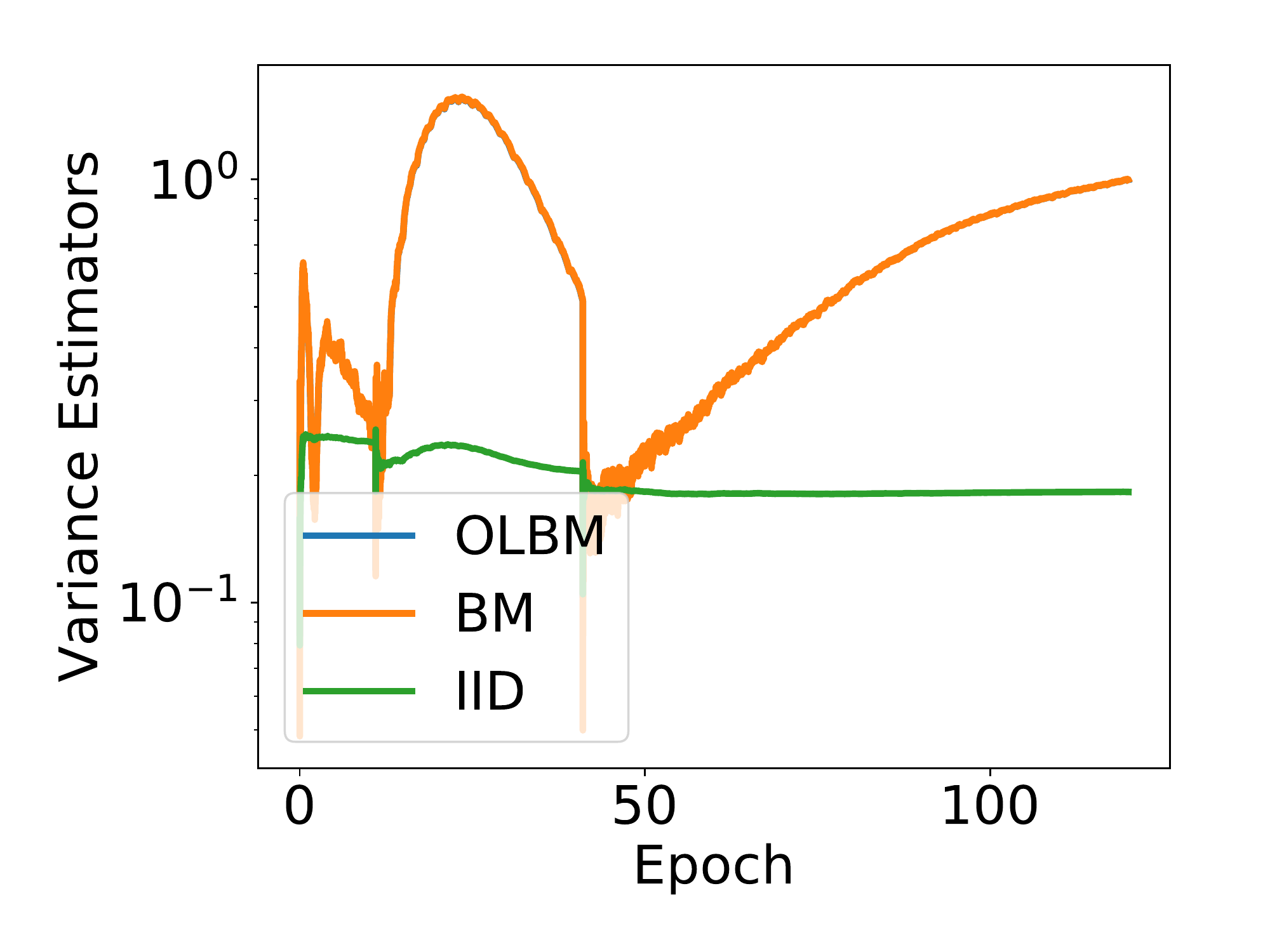}
    \subcaption{}
  \end{subfigure}
  \caption{Evolution of the different statistics for SASA over the course of training ResNet18 on ImageNet using the default parameters $\delta=0.02, \gamma=0.2, \zeta=0.1$. Panel (a) shows the raw data for both sides of condition \eqref{eqn:fdr1}. That is, it shows the values of $\langle \xk, \gk\rangle$ and $\frac{\alpha}{2}\frac{1+\beta}{1-\beta}\langle \dk, \dk\rangle$ at each iteration. Panel (b) shows $\bar{z}_N$ with its lower and upper confidence interval $[lci, uci]$ and the "right hand side" (rhs) $(-\delta \bar{v}_N, \delta\bar{v}_N)$ (see Eqn.~\eqref{eqn:our-test}). Panel (c) shows a zoomed-in version of (b) to show the drop points in more detail. Panel (d) depicts the different variance estimators (i.i.d., batch means, overlapping batch means) over the course of training. The i.i.d. variance (green) is a poor estimate of $\sigma_z^2$.}
  \label{fig:imagenet_stats}
\end{figure}

\paragraph{Evolution of statistics.} Figure \ref{fig:imagenet_stats} shows the evolution of SASA's different statistics over the course of training the ResNet18 model on CIFAR-10, under the default parameter setting $\delta=0.02, \gamma=0.2, \zeta=0.1$. In each phase, $z$ and $v$ get close two easy other as predicted by \eqref{eqn:fdr1}. Together with its confidence interval, the statistics $\bar{z}_N$ decay toward zero. The learning rate is dropped as long as the confidence interval is contained in $(-\delta \bar{v}_N, \delta \bar{v}_N)$; see Eqn~\eqref{eqn:our-test}. The batch mean variance estimator (bm) and overlapping batch mean variance estimator (olbm) give very close variance estimates, while the i.i.d. variance estimator is clearly much different from the batch mean and overlapping batch mean estimators.

\begin{figure}[tb]
    \centering
  \begin{subfigure}[t]{.33\linewidth}
    \centering
    \includegraphics[width=\linewidth]{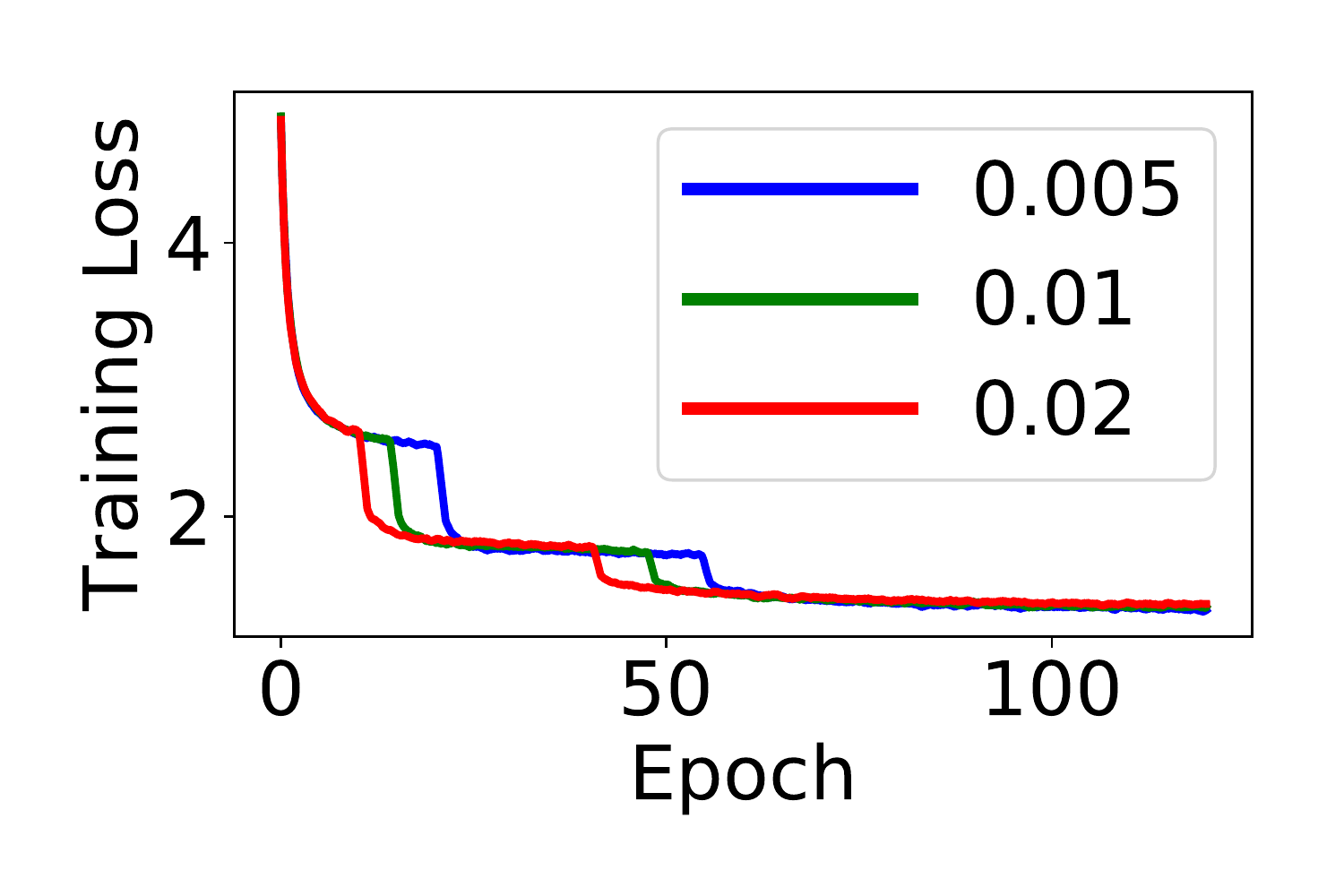}
  \end{subfigure}%
  \begin{subfigure}[t]{.33\linewidth}
    \centering
    \includegraphics[width=\linewidth]{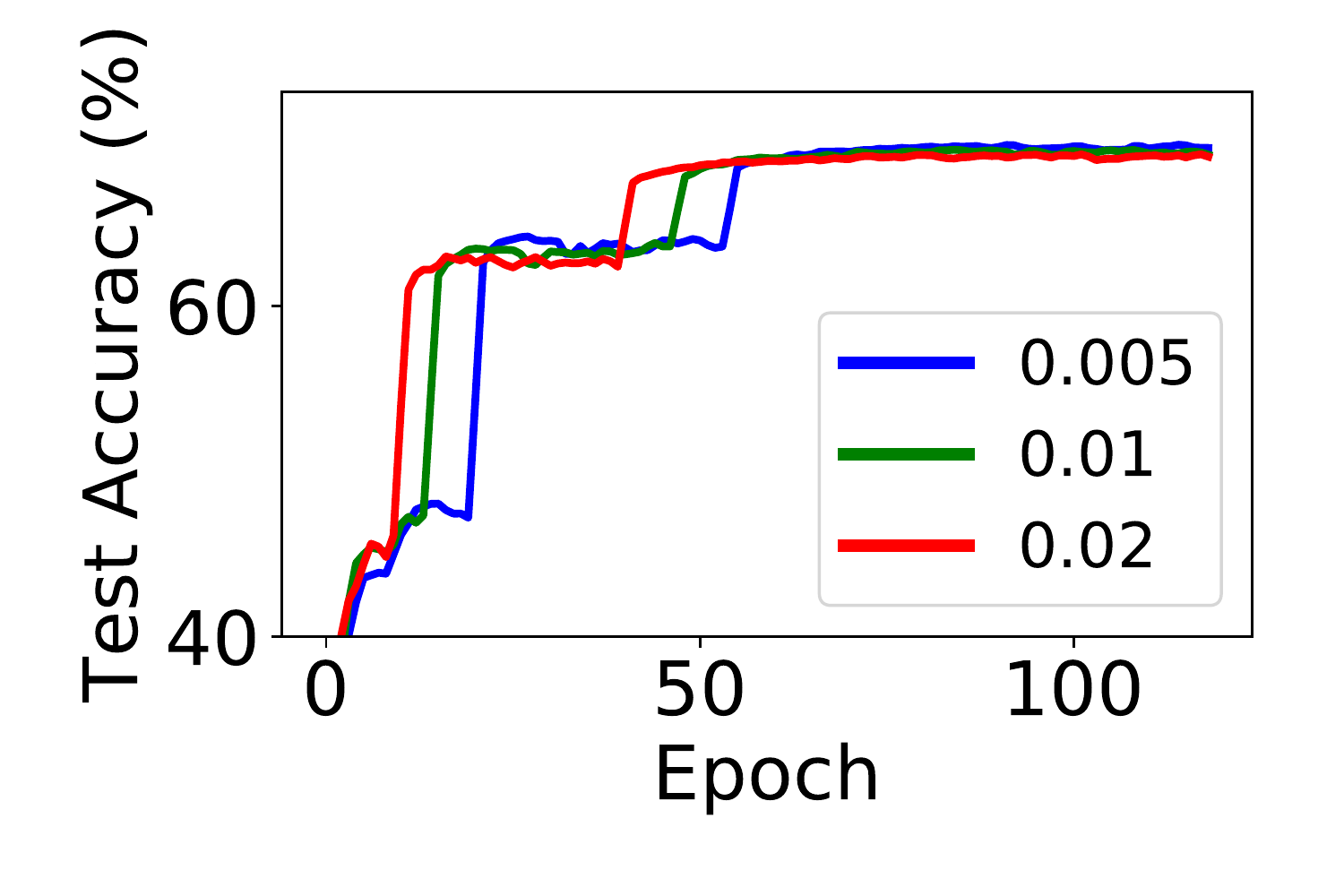}
  \end{subfigure}%
  \begin{subfigure}[t]{.33\linewidth}
    \centering
    \includegraphics[width=\linewidth]{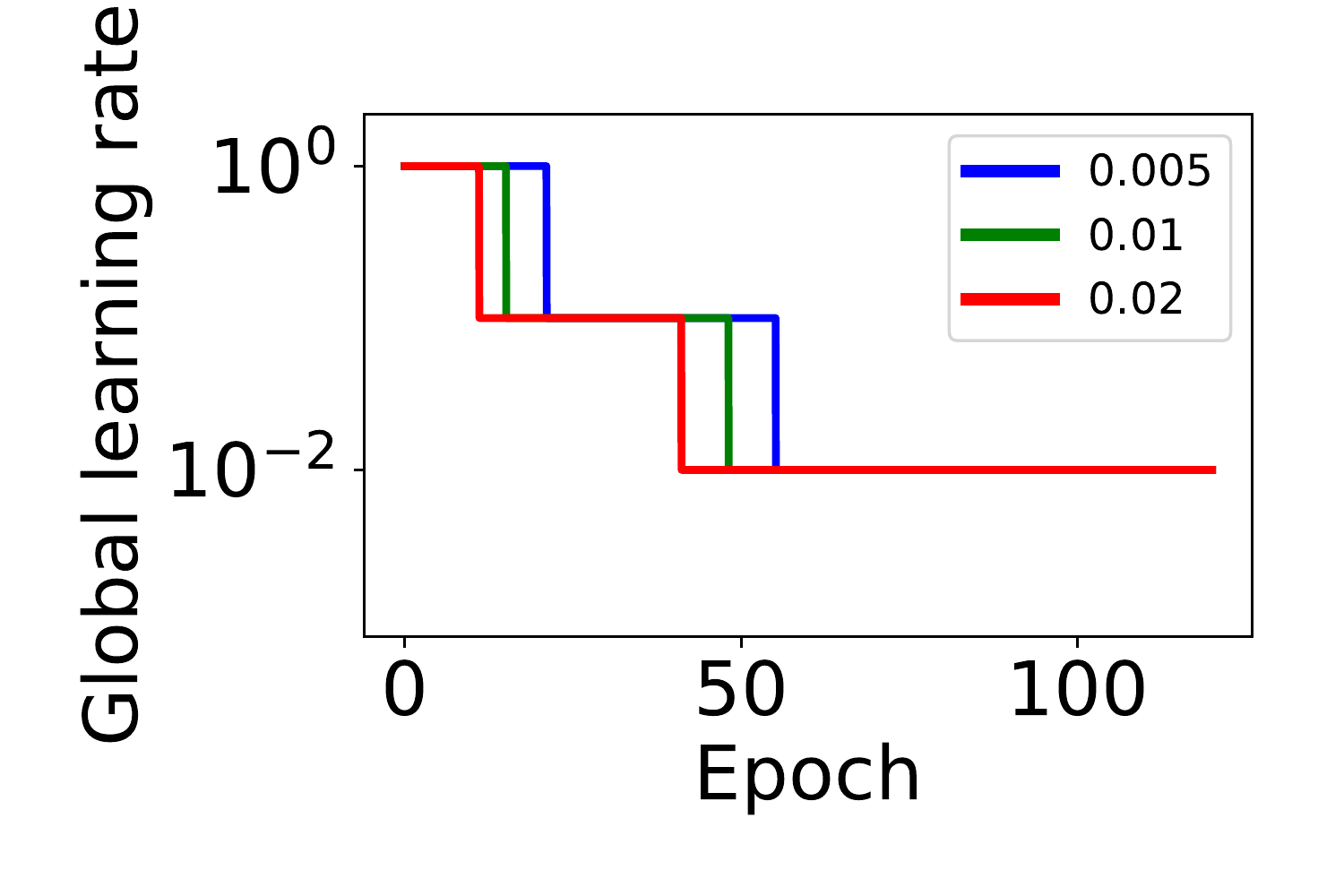}
  \end{subfigure}\\
  \begin{subfigure}[t]{.33\linewidth}
    \centering
    \includegraphics[width=\linewidth]{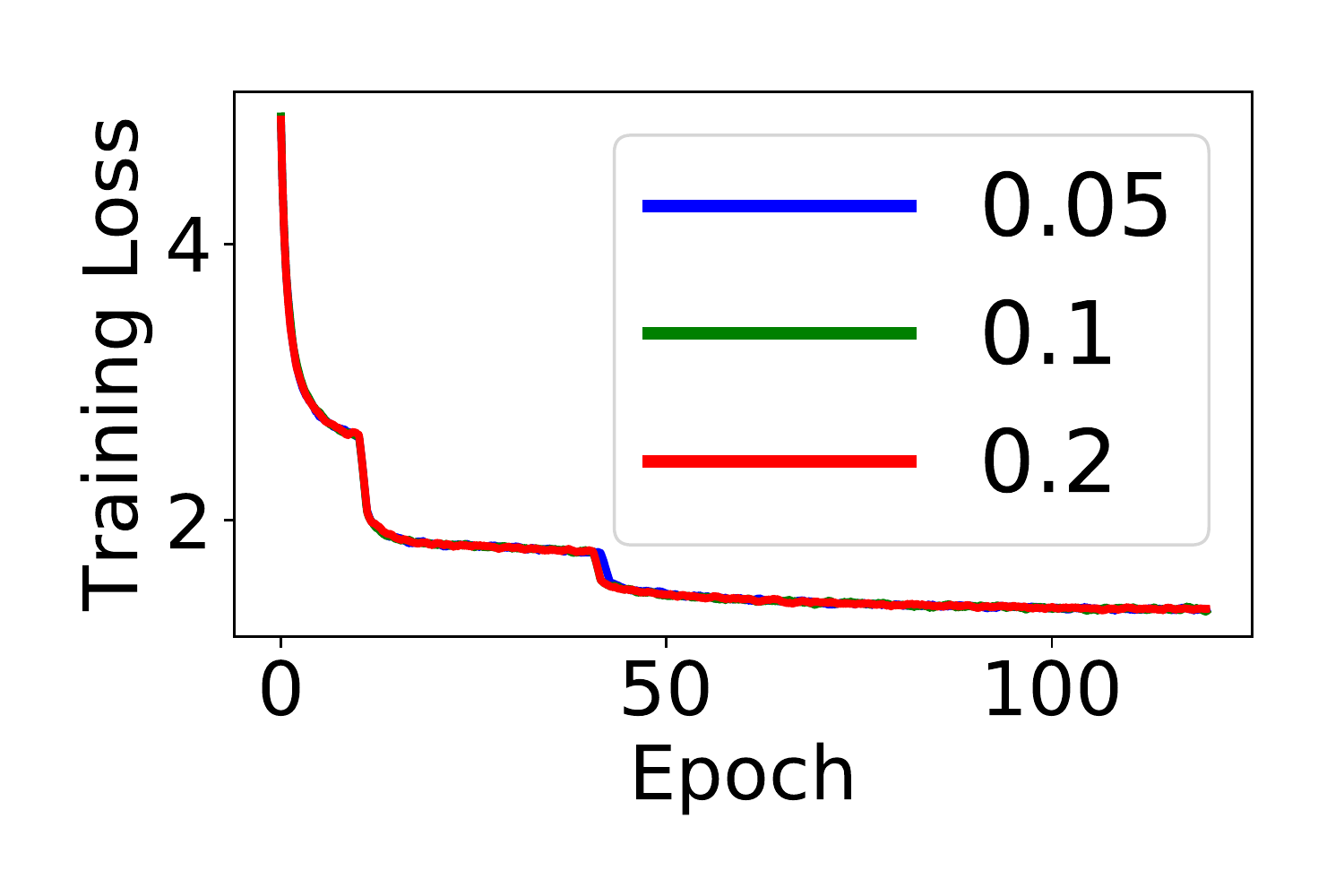}
  \end{subfigure}%
  \begin{subfigure}[t]{.33\linewidth}
    \centering
    \includegraphics[width=\linewidth]{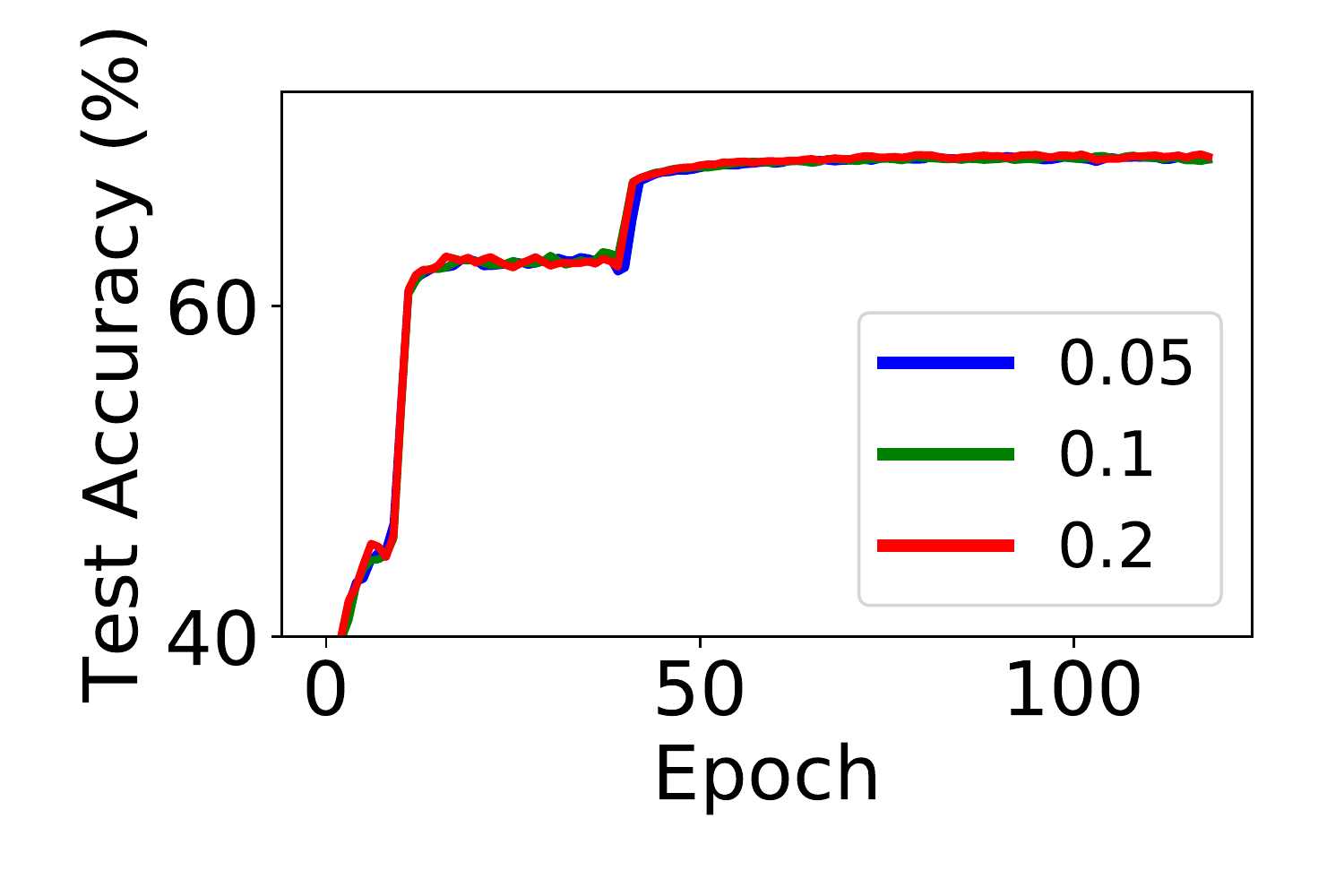}
  \end{subfigure}%
  \begin{subfigure}[t]{.33\linewidth}
    \centering
    \includegraphics[width=\linewidth]{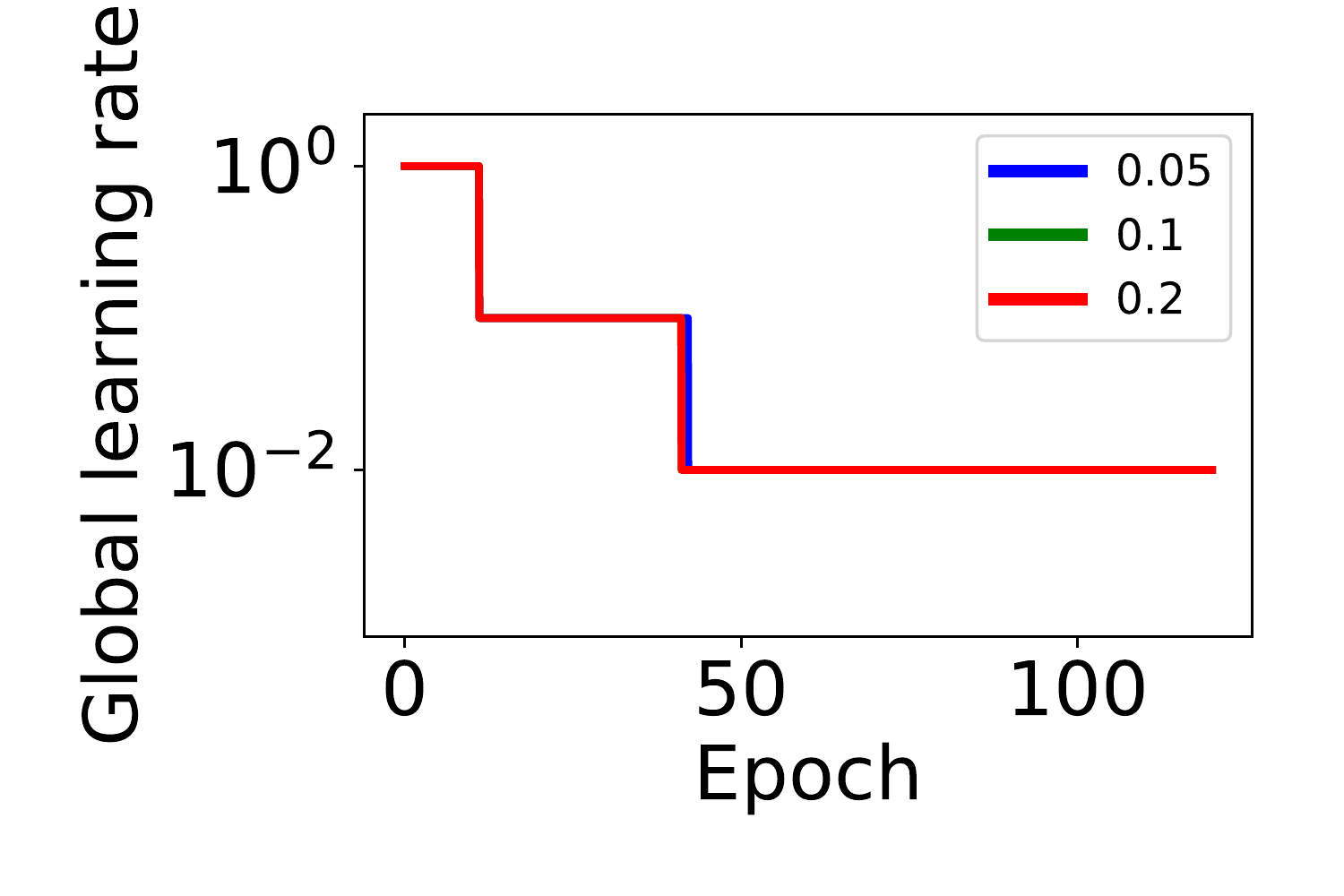}
  \end{subfigure}  \\
  \begin{subfigure}[t]{.33\linewidth}
    \centering
    \includegraphics[width=\linewidth]{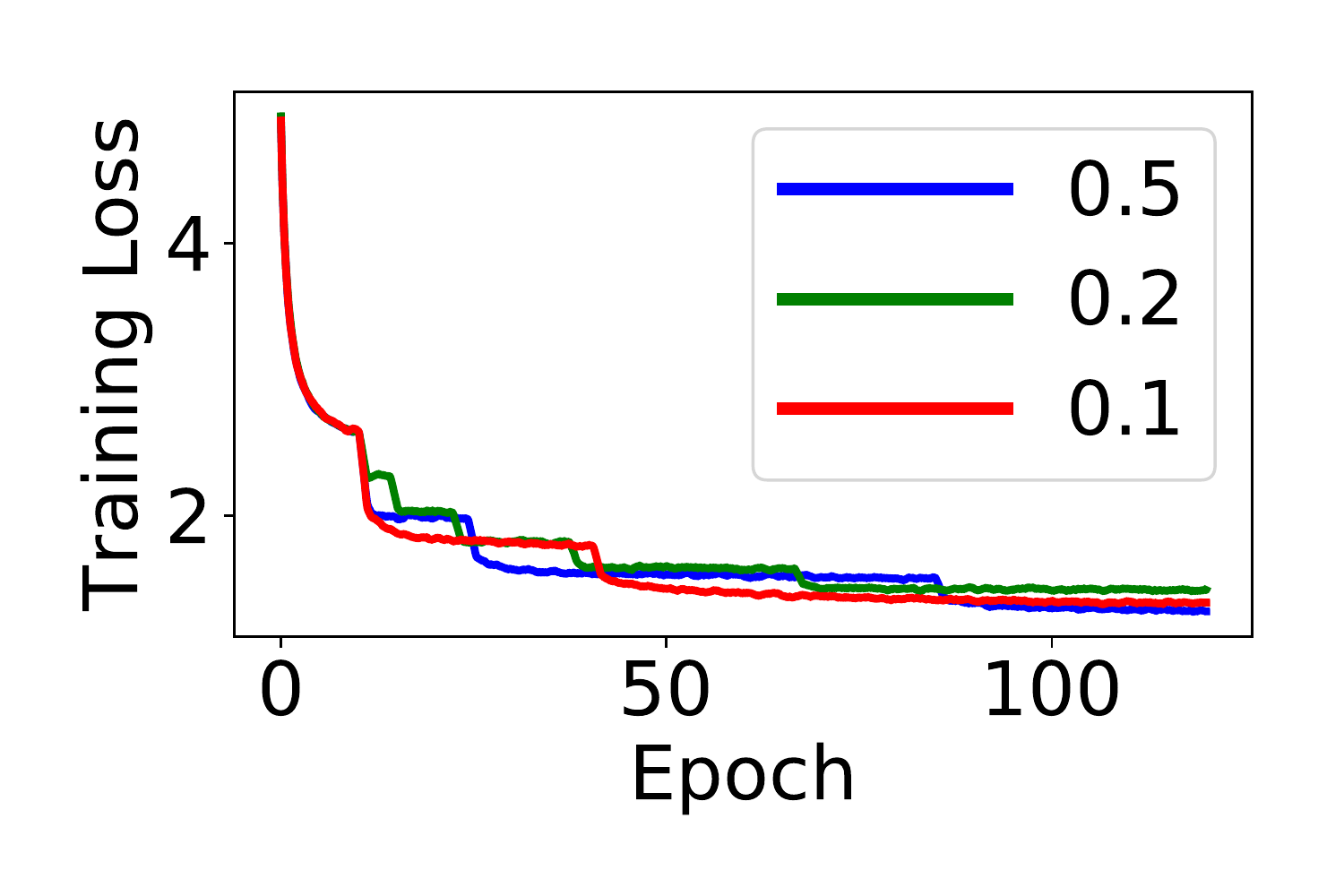}
  \end{subfigure}%
  \begin{subfigure}[t]{.33\linewidth}
    \centering
    \includegraphics[width=\linewidth]{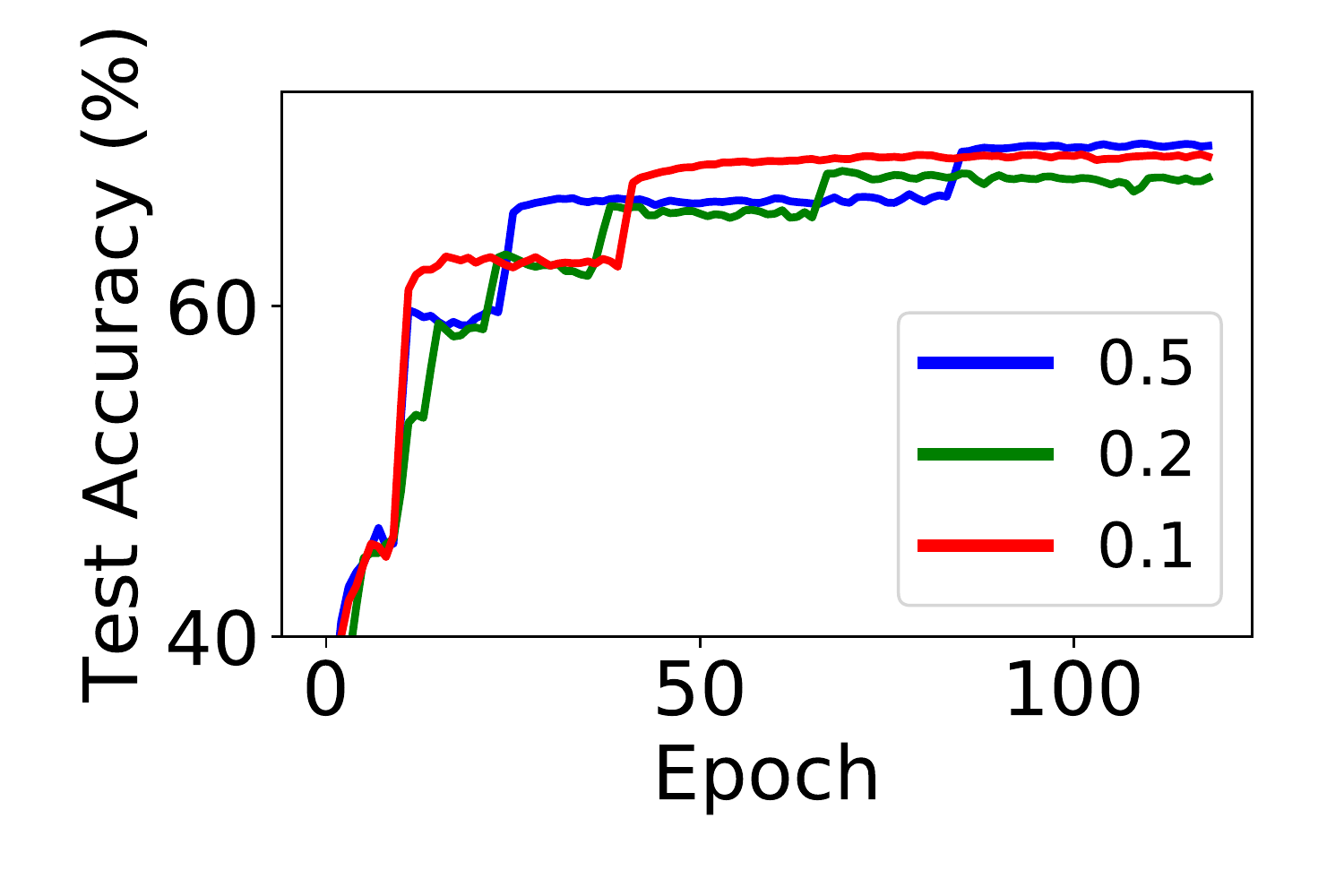}
  \end{subfigure}%
  \begin{subfigure}[t]{.33\linewidth}
    \centering
    \includegraphics[width=\linewidth]{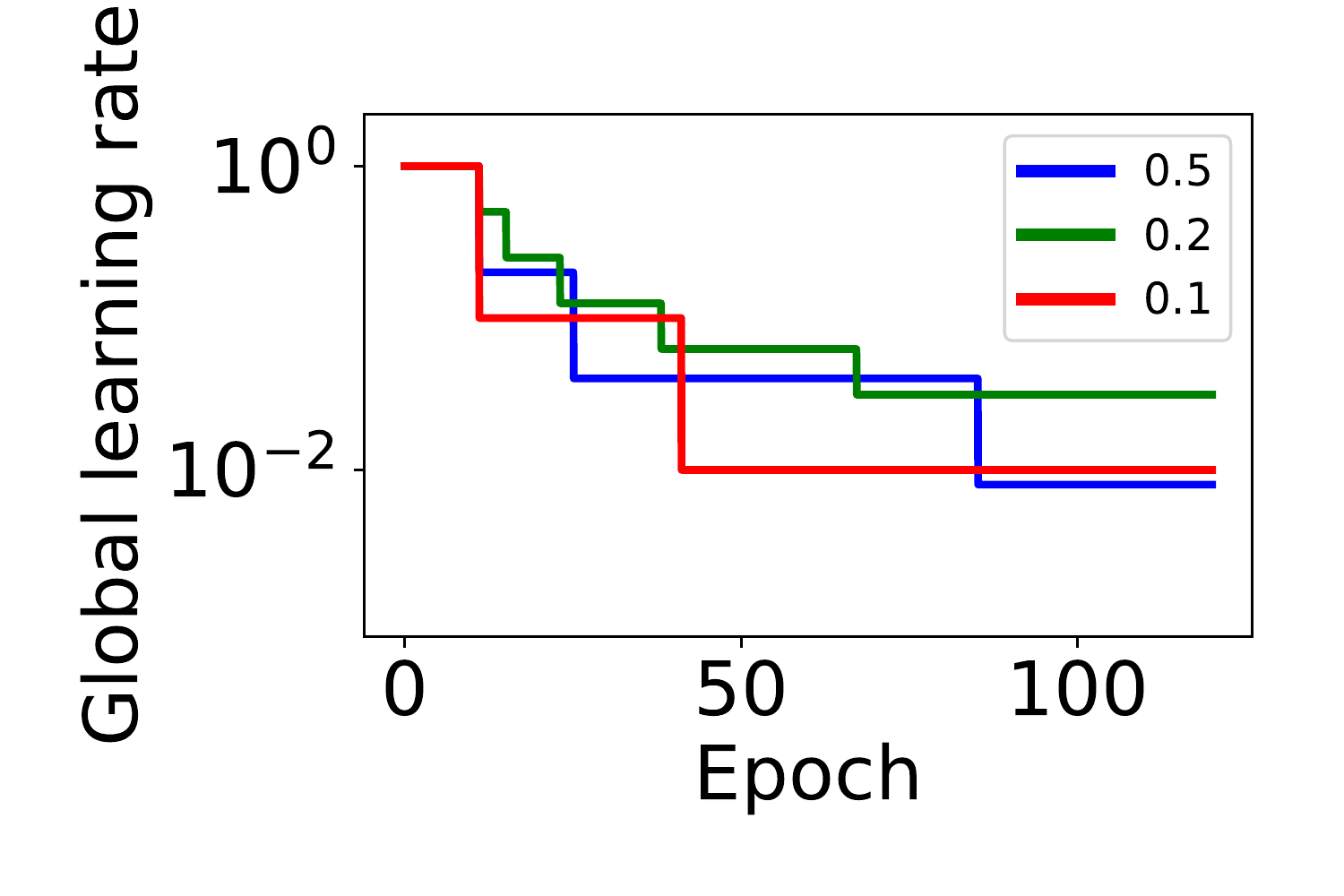}
  \end{subfigure}  
  \caption{Training loss, test accuracy, and learning rate schedule for SASA using different values of $\gamma$, $\delta$ and $\zeta$ around the default $0.2$, $0.02$ and $0.1$. The model is ResNet18 trained on ImageNet, as in \ref{sec:experiments}. Top row: performance for fixed $\gamma=0.2, \zeta=0.1$, and $\delta \in \{0.005, 0.01, 0.02\}$. Middle row: performance for fixed $\delta=0.02, \zeta=0.1$, and $\gamma \in \{0.05, 0.1, 0.2\}$. Bottom row: performance for fixed $\gamma=0.2, \delta=0.02$, and $\zeta \in \{0.5, 0.2, 0.1\}$. Qualitatively, increasing $\delta$ and increasing $\gamma$ both cause the algorithm to drop sooner. The value of $\zeta$ does not influence the final performance, as long as the learning rate finally decays to the same level.}
  \label{fig:imagenetsens}
 \end{figure}

\paragraph{Sensitivity analysis.} We perturb the relative equivalence threshold $\delta$, the confidence level $\gamma$ and the decay rate $\zeta$ around their default values $(0.2, 0.02, 0.1)$ and repeat the CIFAR-10 experiment from the previous section, using the same values for the other hyperparameters. In Figure~\ref{fig:imagenetsens}, the top row shows the performance for fixed $(\gamma, \zeta)=(0.2, 0.1)$ and changing $\delta$. The middle row shows the performance for fixed $(\delta, \zeta)=(0.02, 0.1)$ and changing $\gamma$. The bottom row shows the performance for fixed $(\delta, \gamma)=(0.02, 0.2)$ and changing $\zeta$. Increases in both $\delta$ and $\gamma$ tend to cause the algorithm to drop sooner; this behavior is intuitive from the testing procedure \eqref{eqn:our-test}. For values of the parameters close to the defaults, SASA still obtains good performance. The value of $\zeta$ does not influence the final performance, as long as the learning rate finally decays to the same level.

\begin{figure}[tb]
  \centering
  \begin{subfigure}[t]{.29\linewidth}
    \centering
    \includegraphics[width=\linewidth]{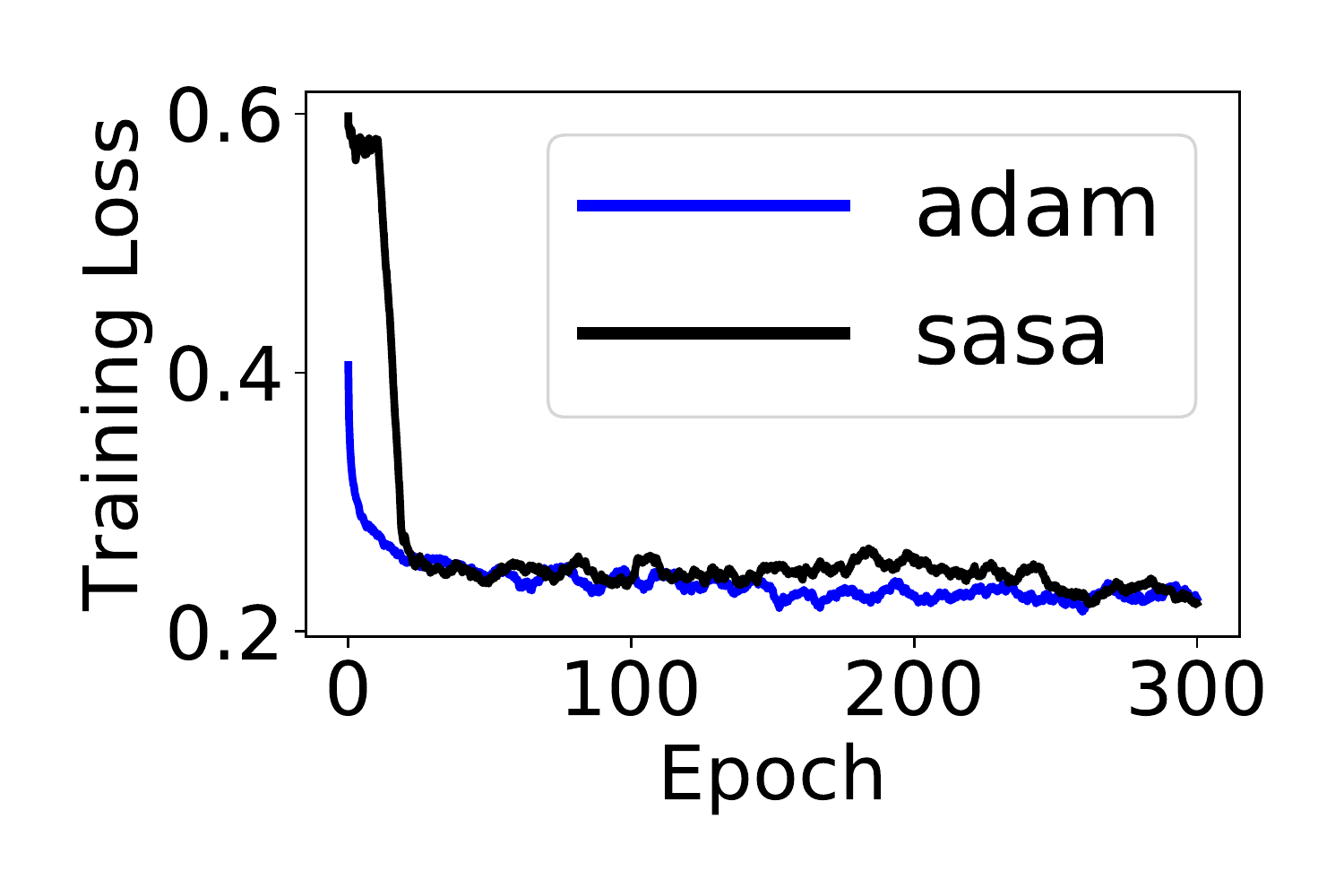}
  \end{subfigure}%
  \begin{subfigure}[t]{.29\linewidth}
    \centering
    \includegraphics[width=\linewidth]{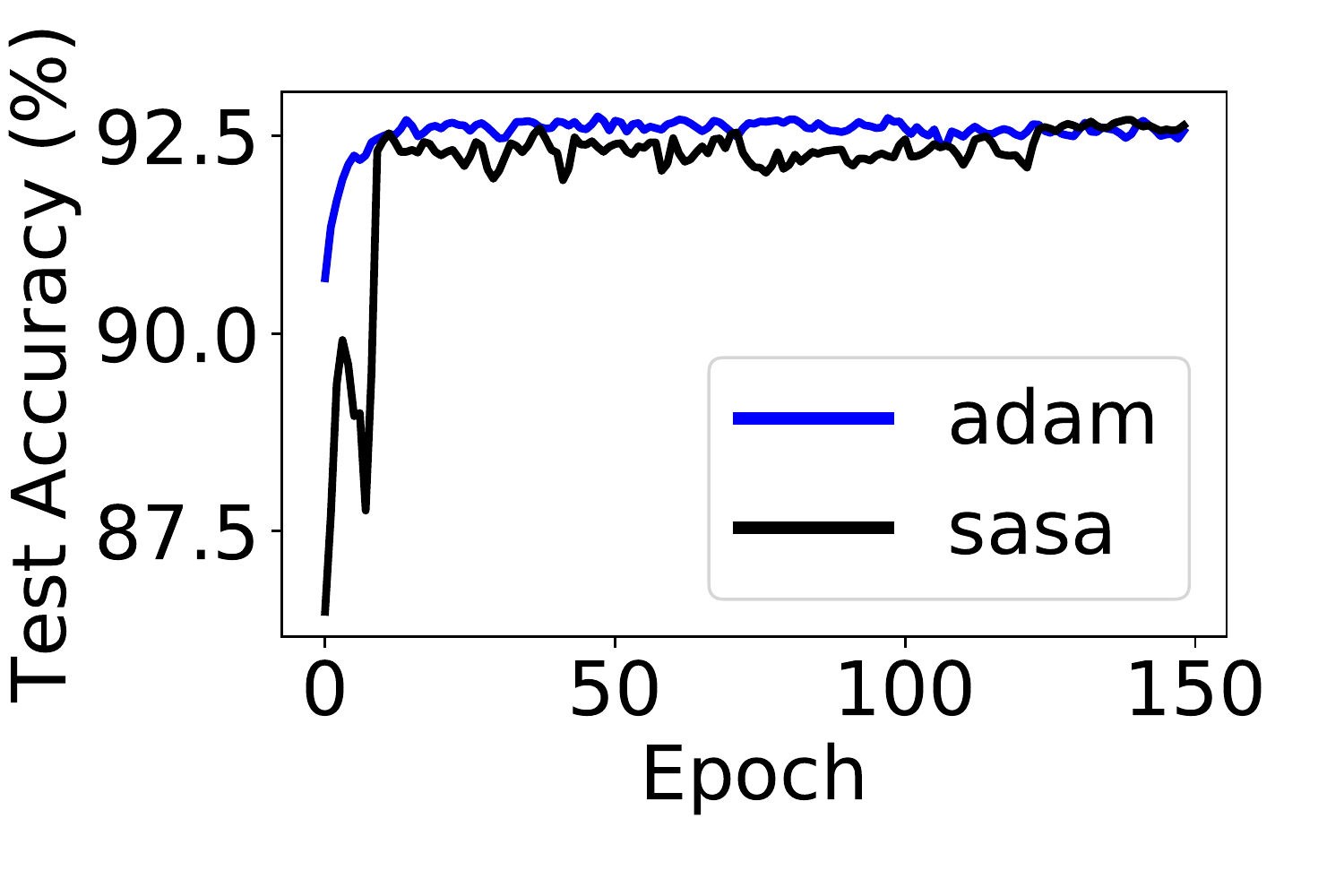}
  \end{subfigure}%
  \begin{subfigure}[t]{.29\linewidth}
    \centering
    \includegraphics[width=\linewidth]{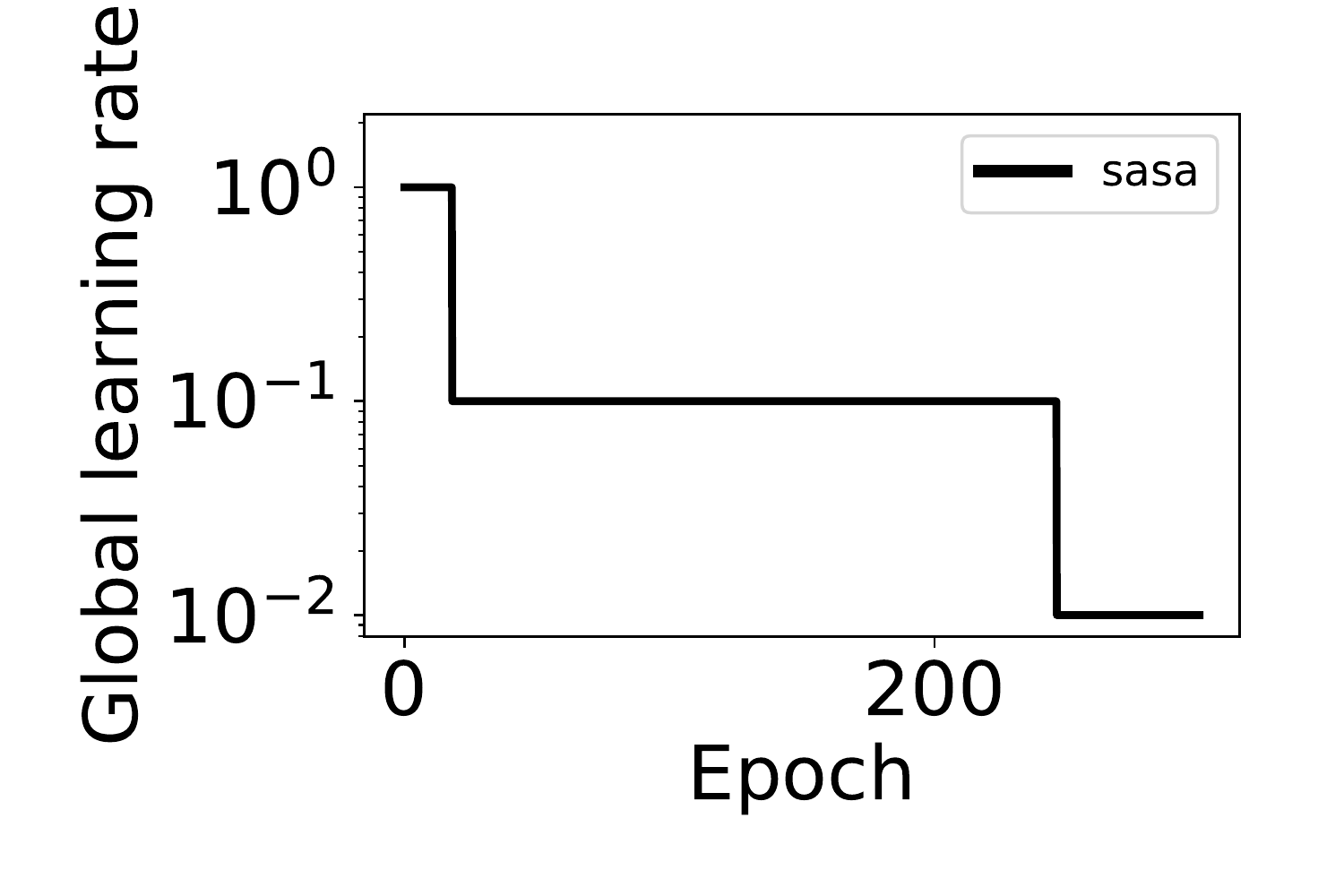}
  \end{subfigure}\\
  \begin{subfigure}[t]{.245\linewidth}
    \centering
    \includegraphics[width=\linewidth]{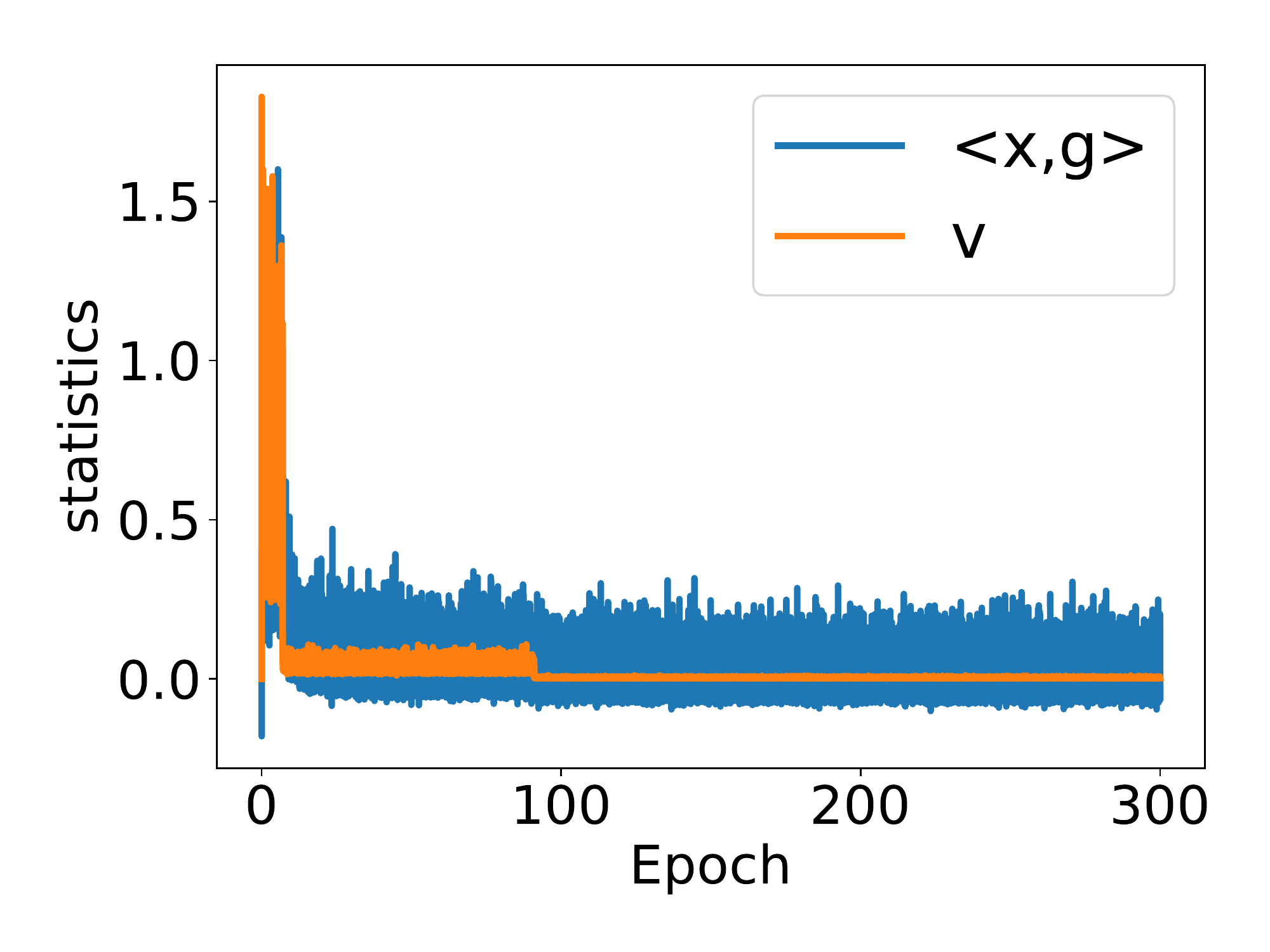}
  \end{subfigure}%
  \begin{subfigure}[t]{.245\linewidth}
    \centering
    \includegraphics[width=\linewidth]{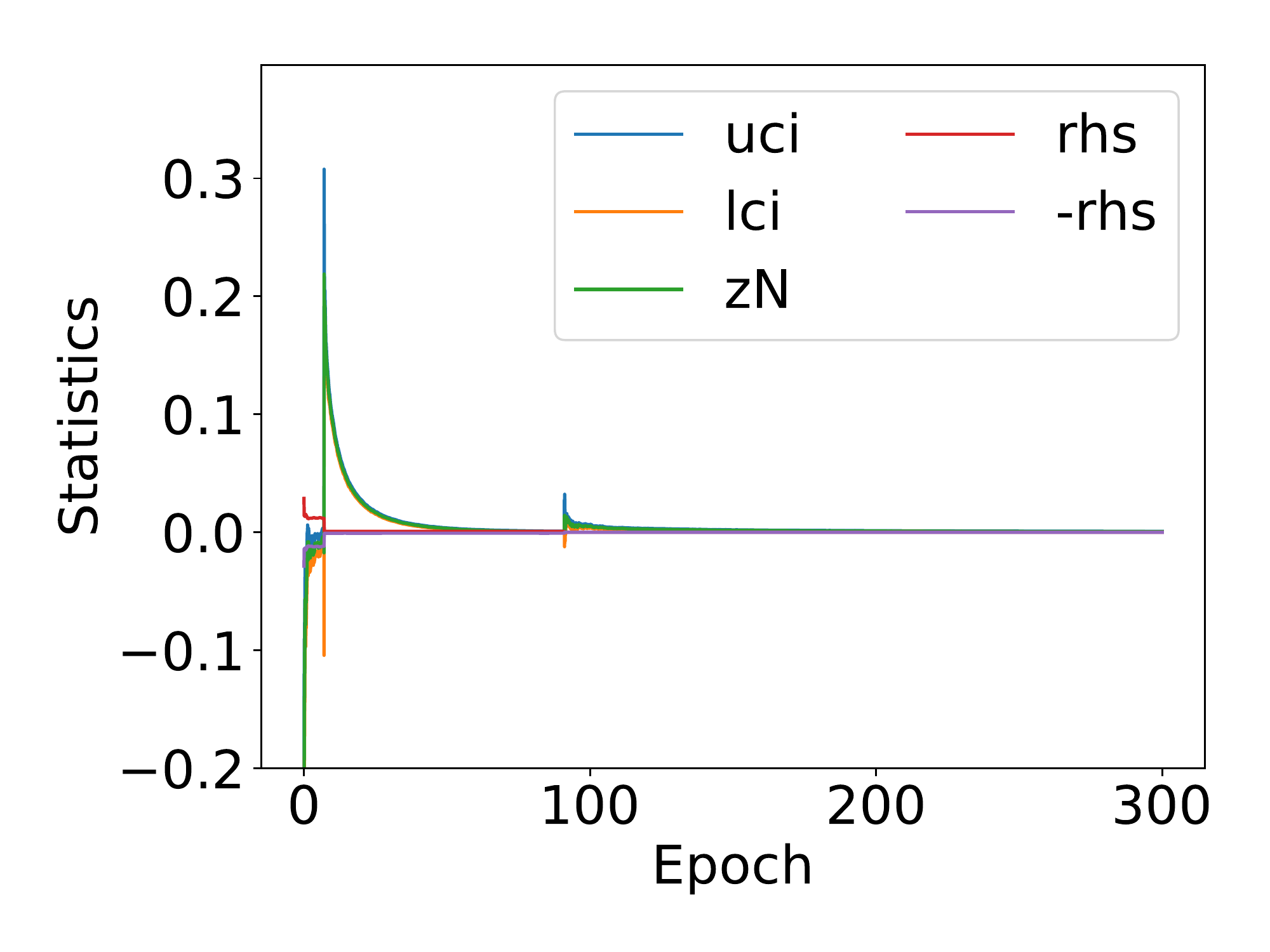}
  \end{subfigure}%
  \begin{subfigure}[t]{.245\linewidth}
    \centering
    \includegraphics[width=\linewidth]{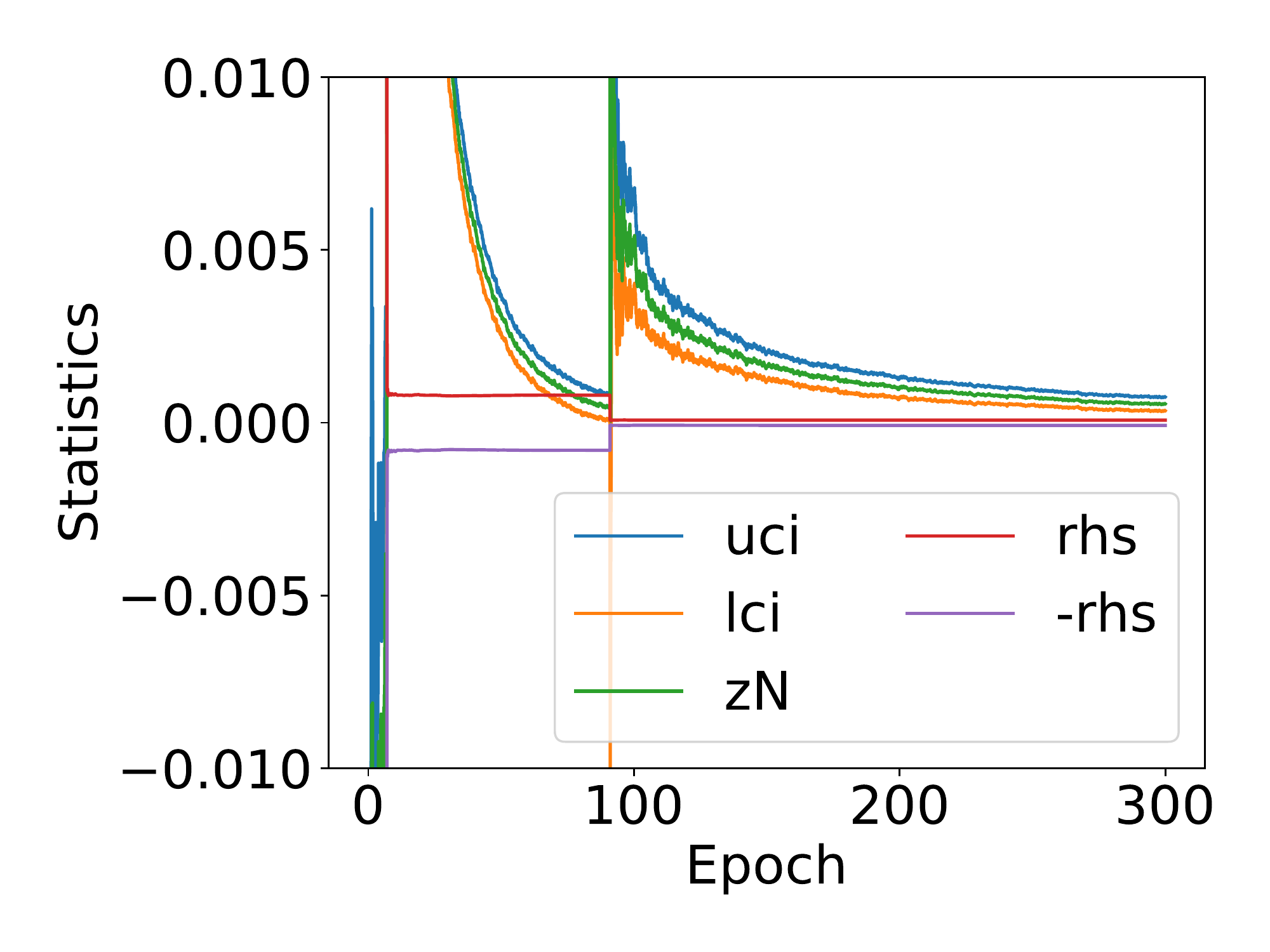}
  \end{subfigure}%
  \begin{subfigure}[t]{.245\linewidth}
    \centering
    \includegraphics[width=\linewidth]{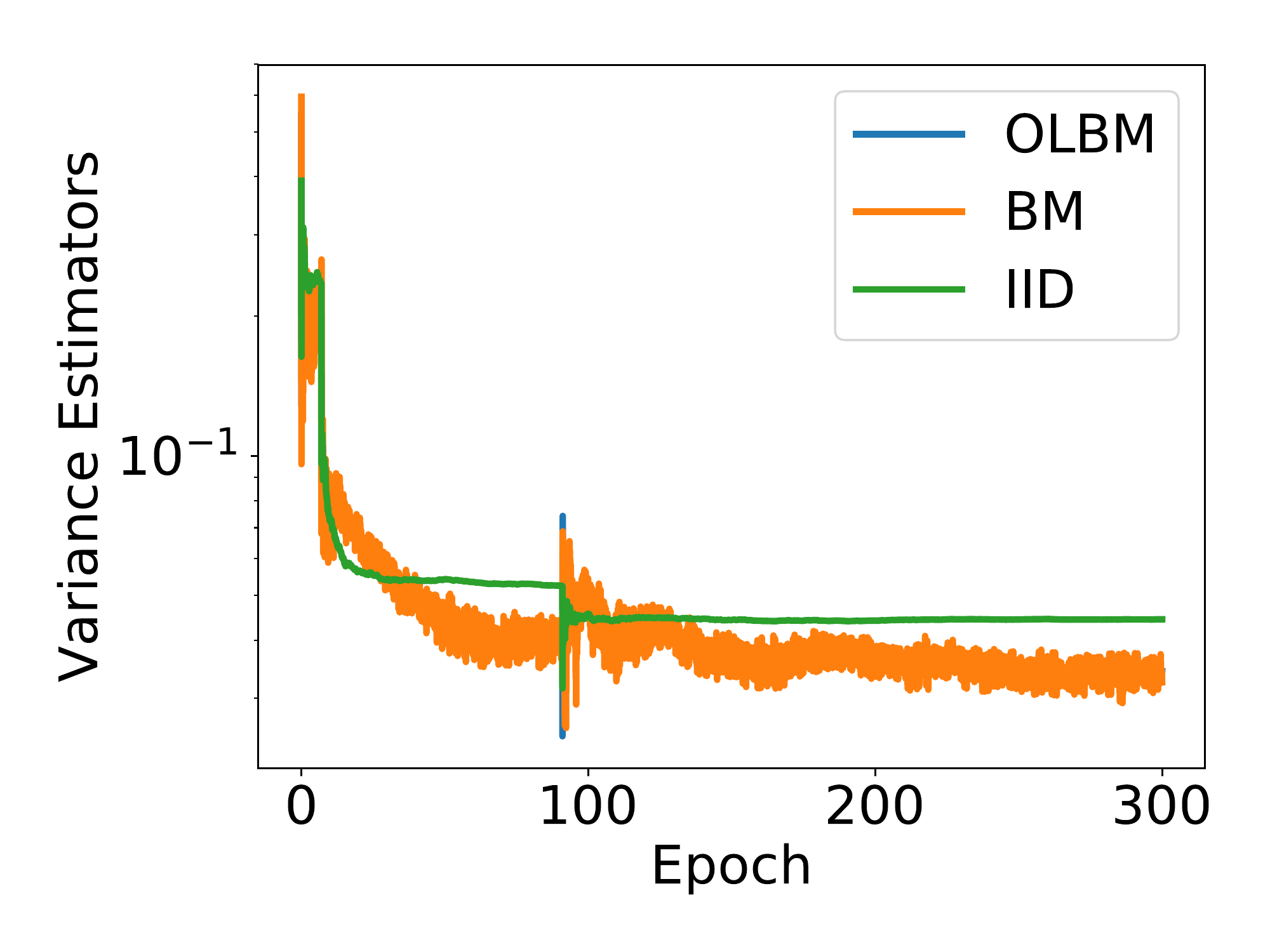}
  \end{subfigure}
  \caption{Top: training loss, test accuracy, and learning rate schedule for SASA and Adam for logistic regression on MNIST. Bottom: Evolution of the different statistics for SASA, as in Figures \ref{fig:cifar_stats} and \ref{fig:imagenet_stats}. 
  SASA uses its default parameters $(\delta, \gamma, \zeta)=(0.02,0.2,0.1)$. Adam uses its default $(\beta_1, \beta_2) = (0.9, 0.999)$ but its initial learning rate $\alpha_0=0.00033$ is obtained from a grid search.}
  \label{fig:lr}
\end{figure}

\begin{figure}[tb]
    \centering
  \begin{subfigure}[t]{.33\linewidth}
    \centering
    \includegraphics[width=\linewidth]{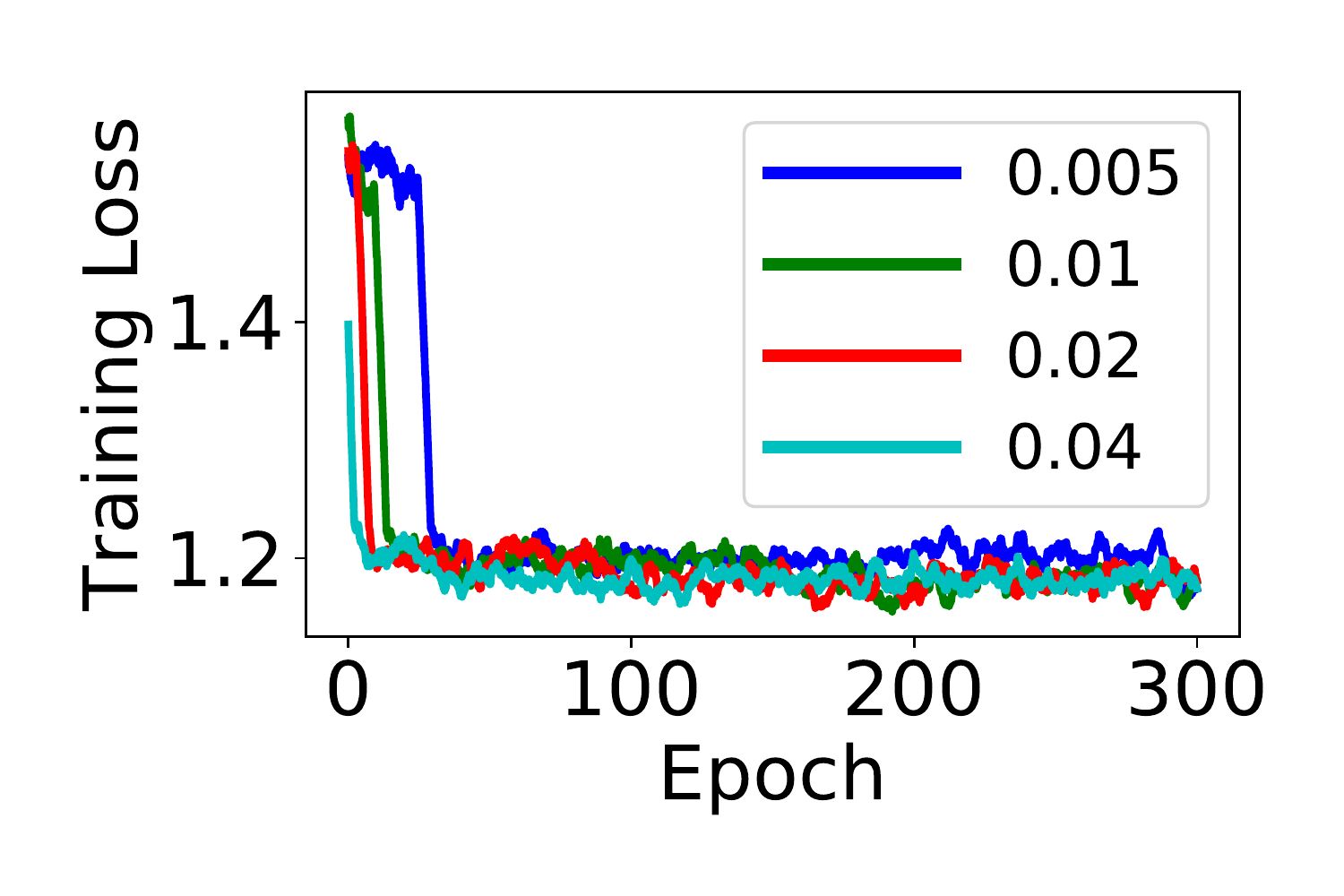}
  \end{subfigure}%
  \begin{subfigure}[t]{.33\linewidth}
    \centering
    \includegraphics[width=\linewidth]{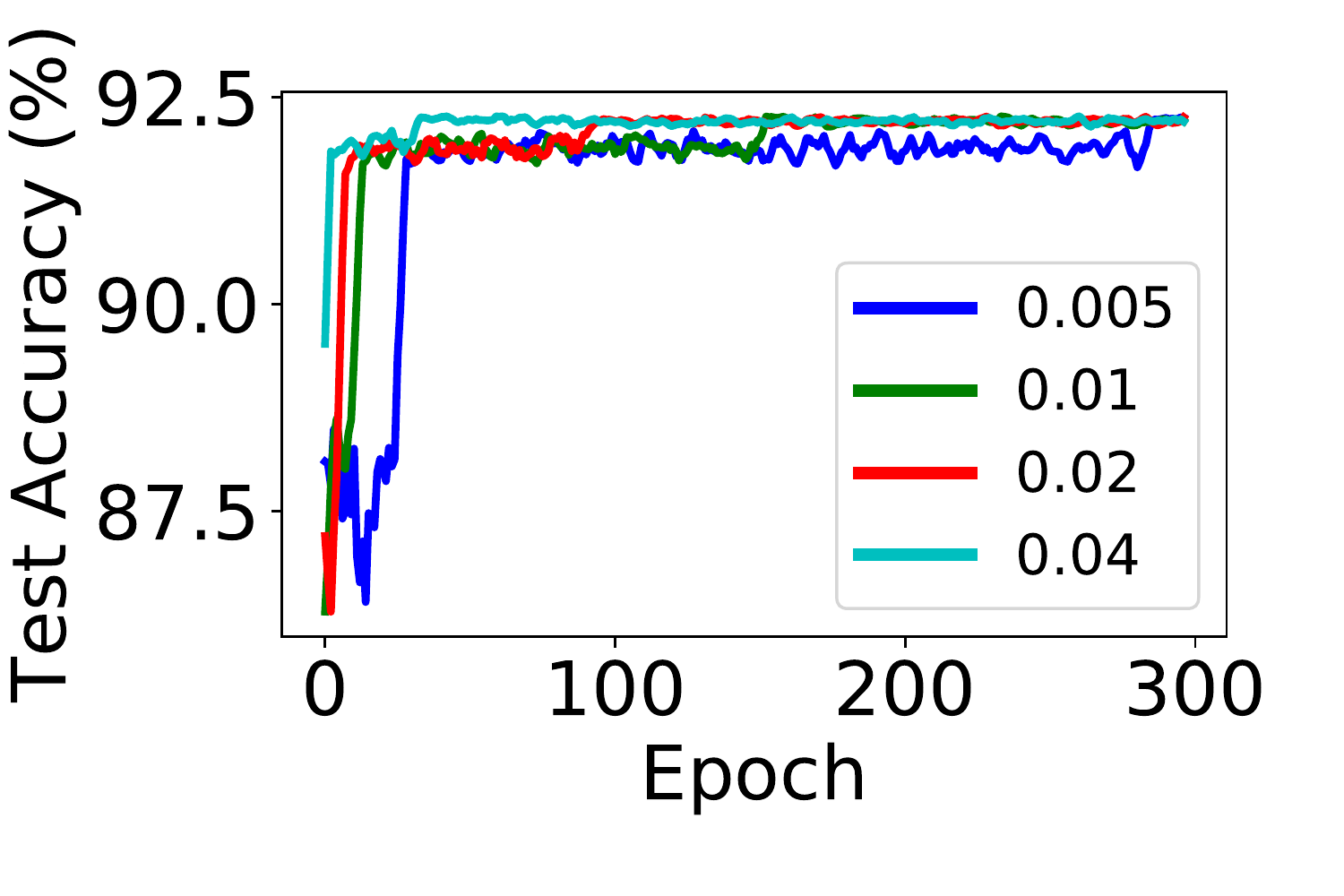}
  \end{subfigure}%
  \begin{subfigure}[t]{.33\linewidth}
    \centering
    \includegraphics[width=\linewidth]{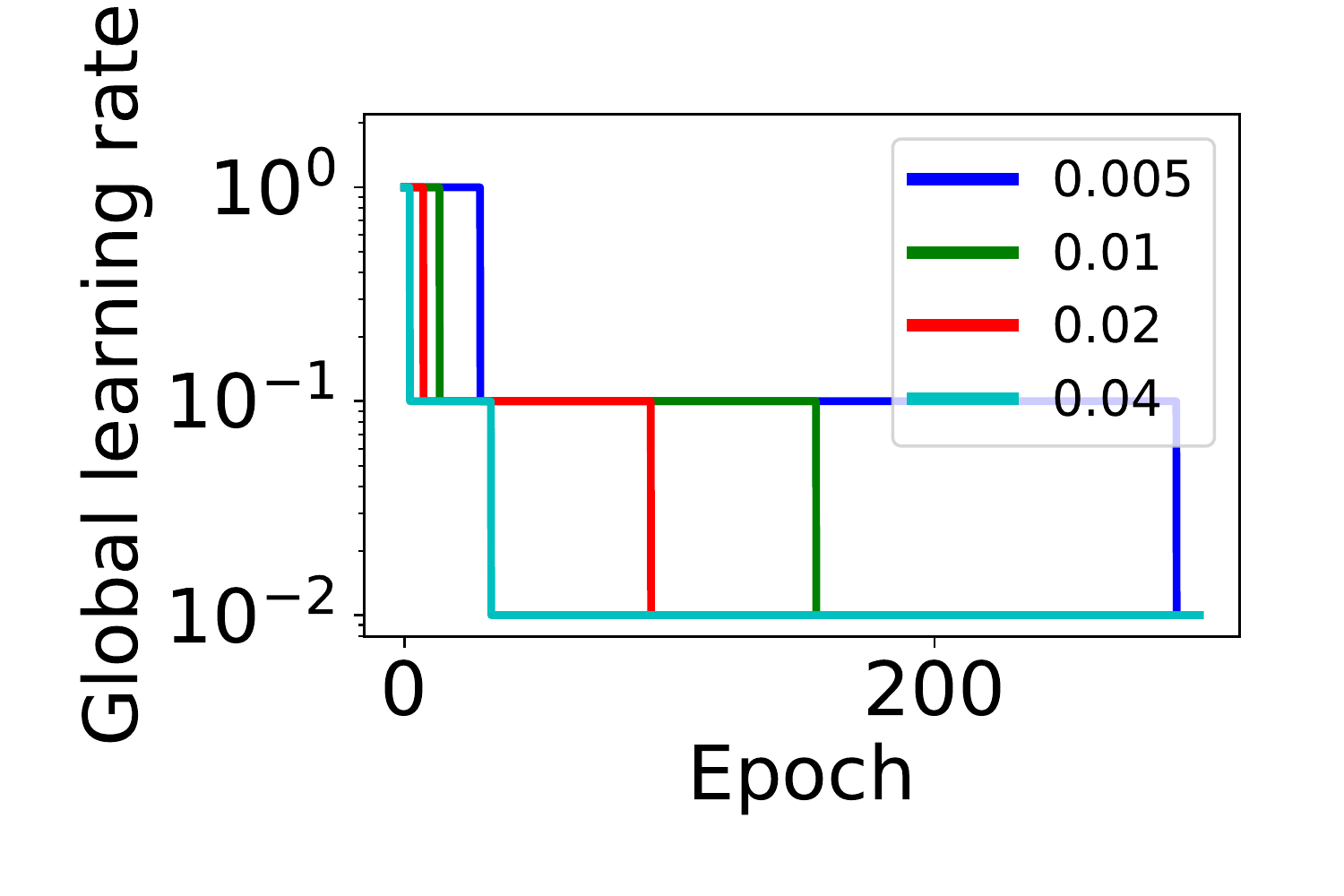}
  \end{subfigure}\\
  \begin{subfigure}[t]{.33\linewidth}
    \centering
    \includegraphics[width=\linewidth]{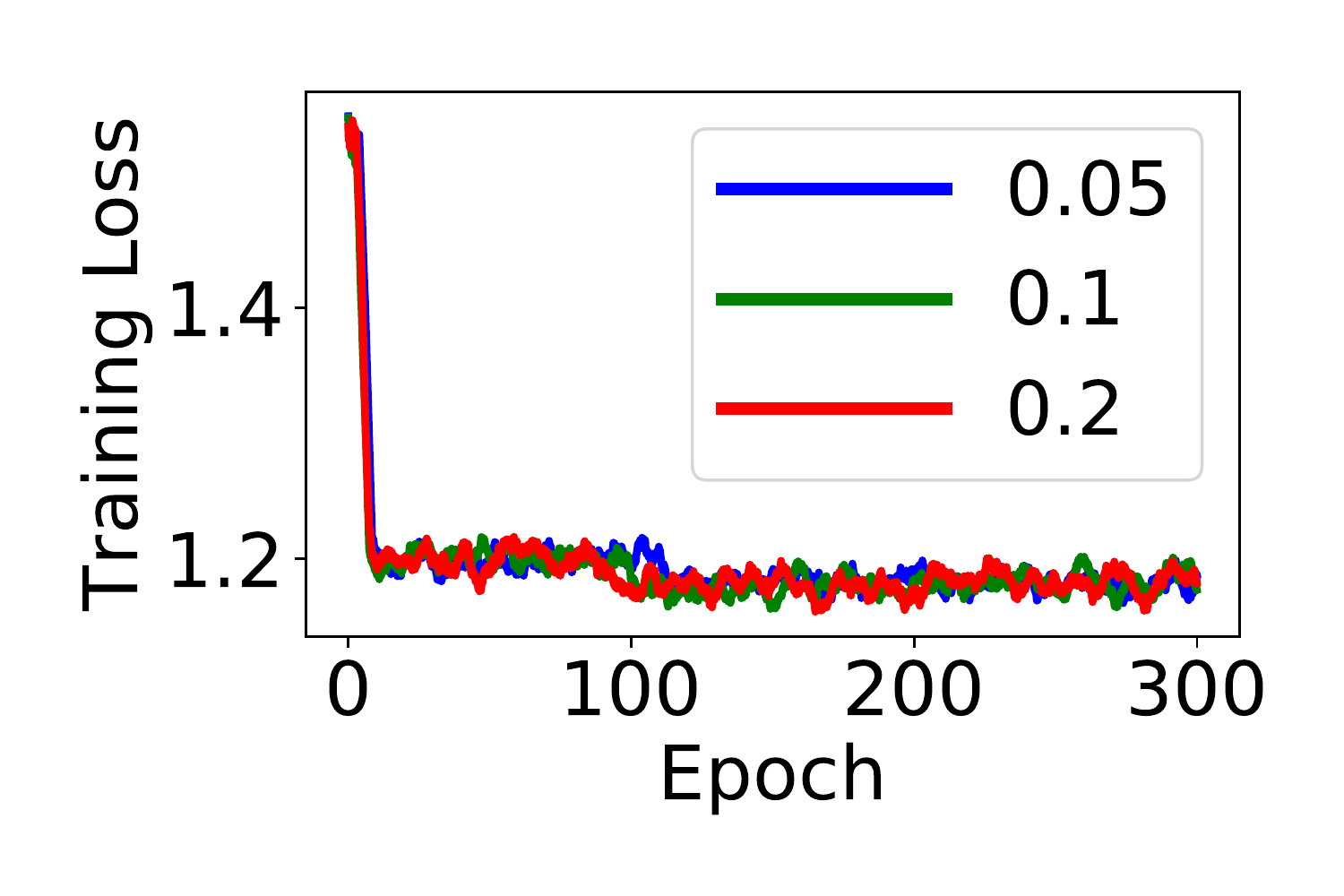}
  \end{subfigure}%
  \begin{subfigure}[t]{.33\linewidth}
    \centering
    \includegraphics[width=\linewidth]{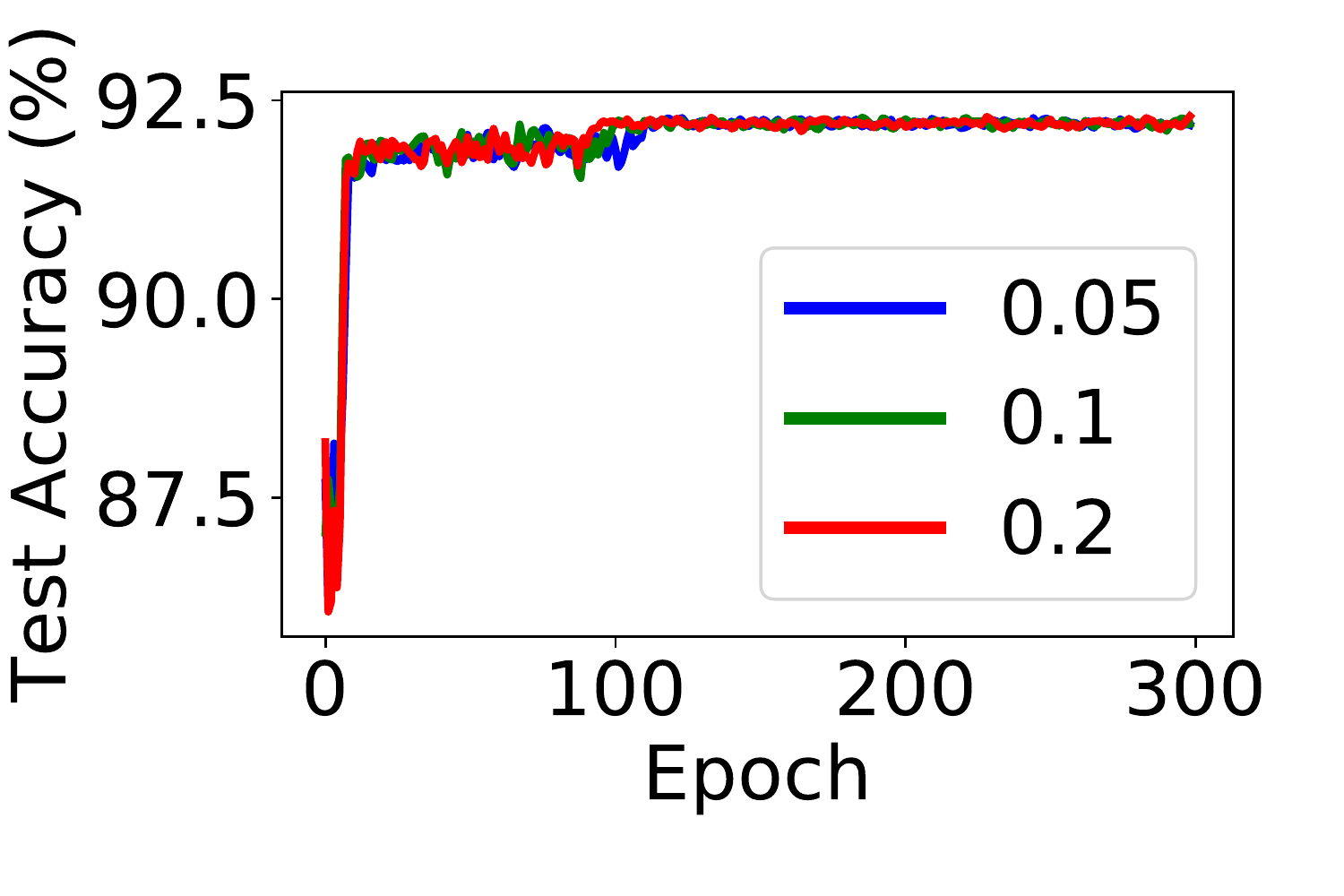}
  \end{subfigure}%
  \begin{subfigure}[t]{.33\linewidth}
    \centering
    \includegraphics[width=\linewidth]{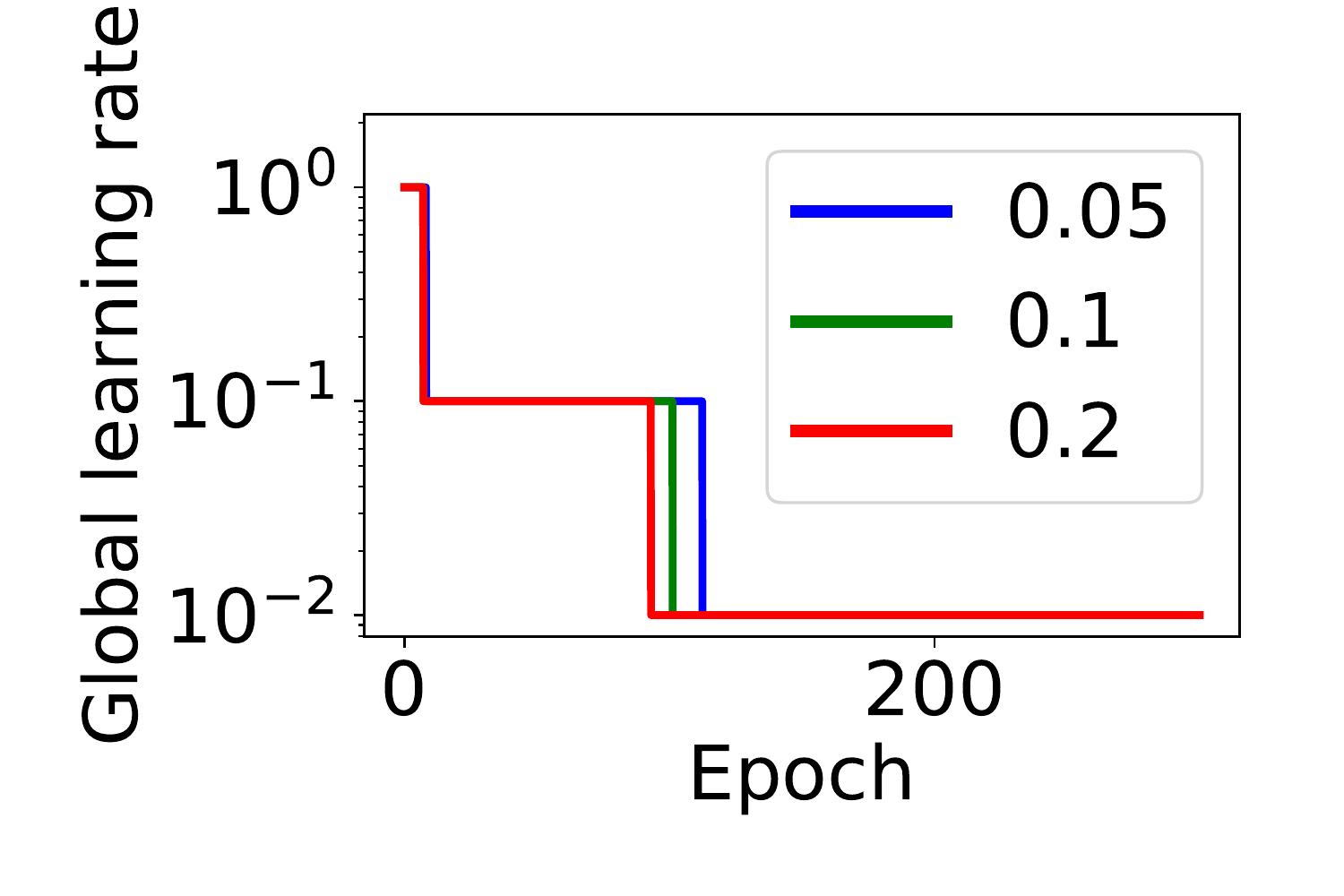}
  \end{subfigure}  \\
  \begin{subfigure}[t]{.33\linewidth}
    \centering
    \includegraphics[width=\linewidth]{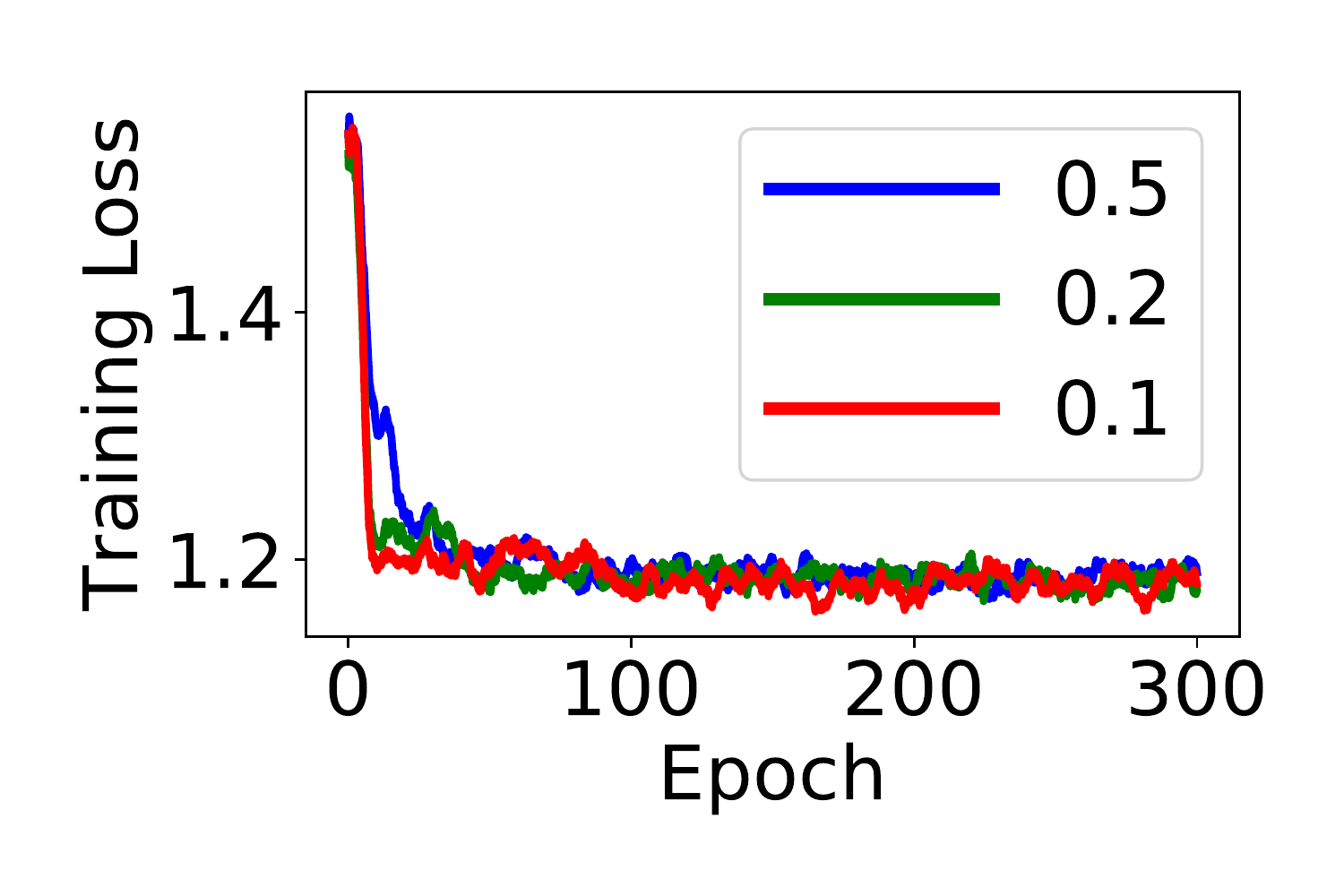}
  \end{subfigure}%
  \begin{subfigure}[t]{.33\linewidth}
    \centering
    \includegraphics[width=\linewidth]{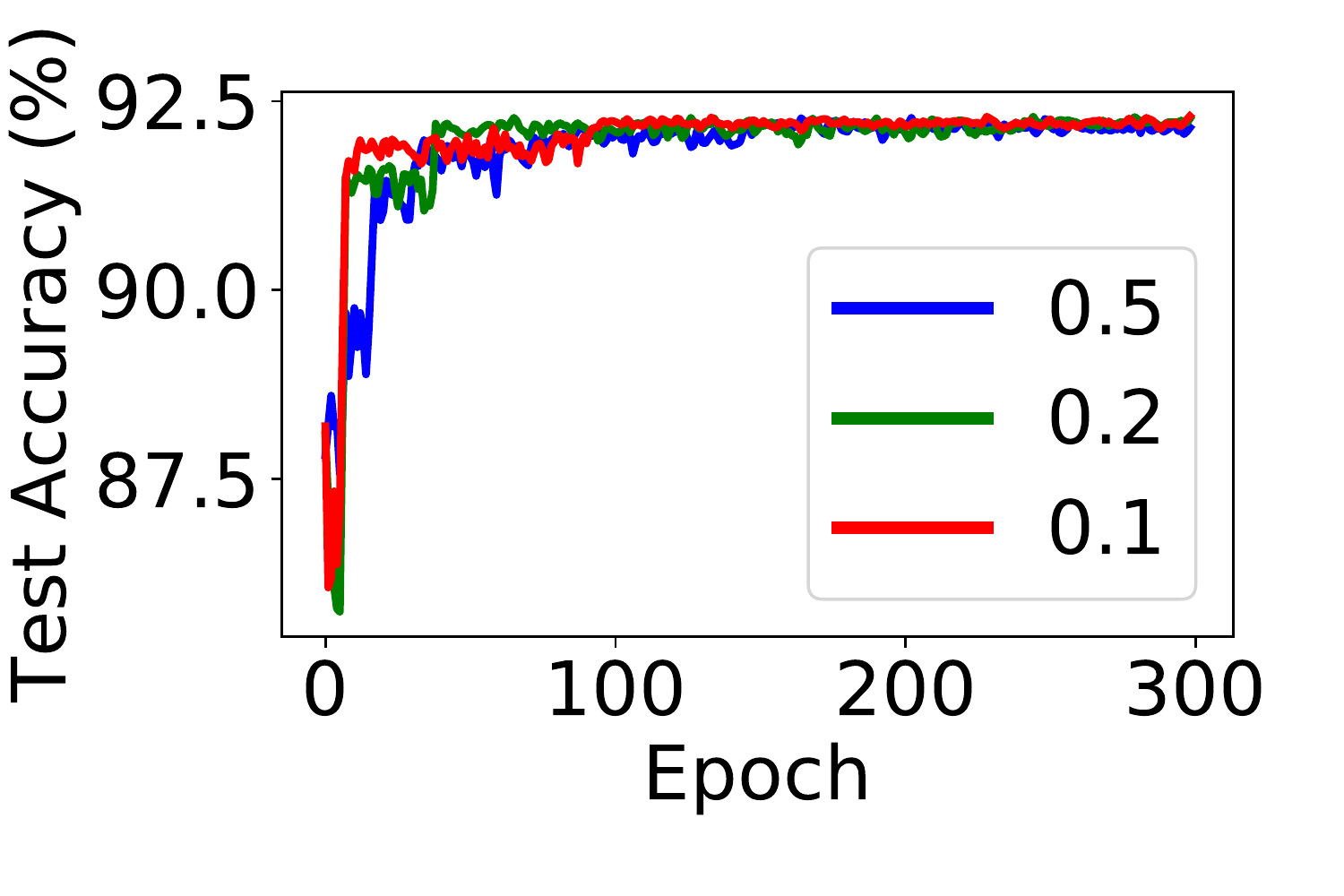}
  \end{subfigure}%
  \begin{subfigure}[t]{.33\linewidth}
    \centering
    \includegraphics[width=\linewidth]{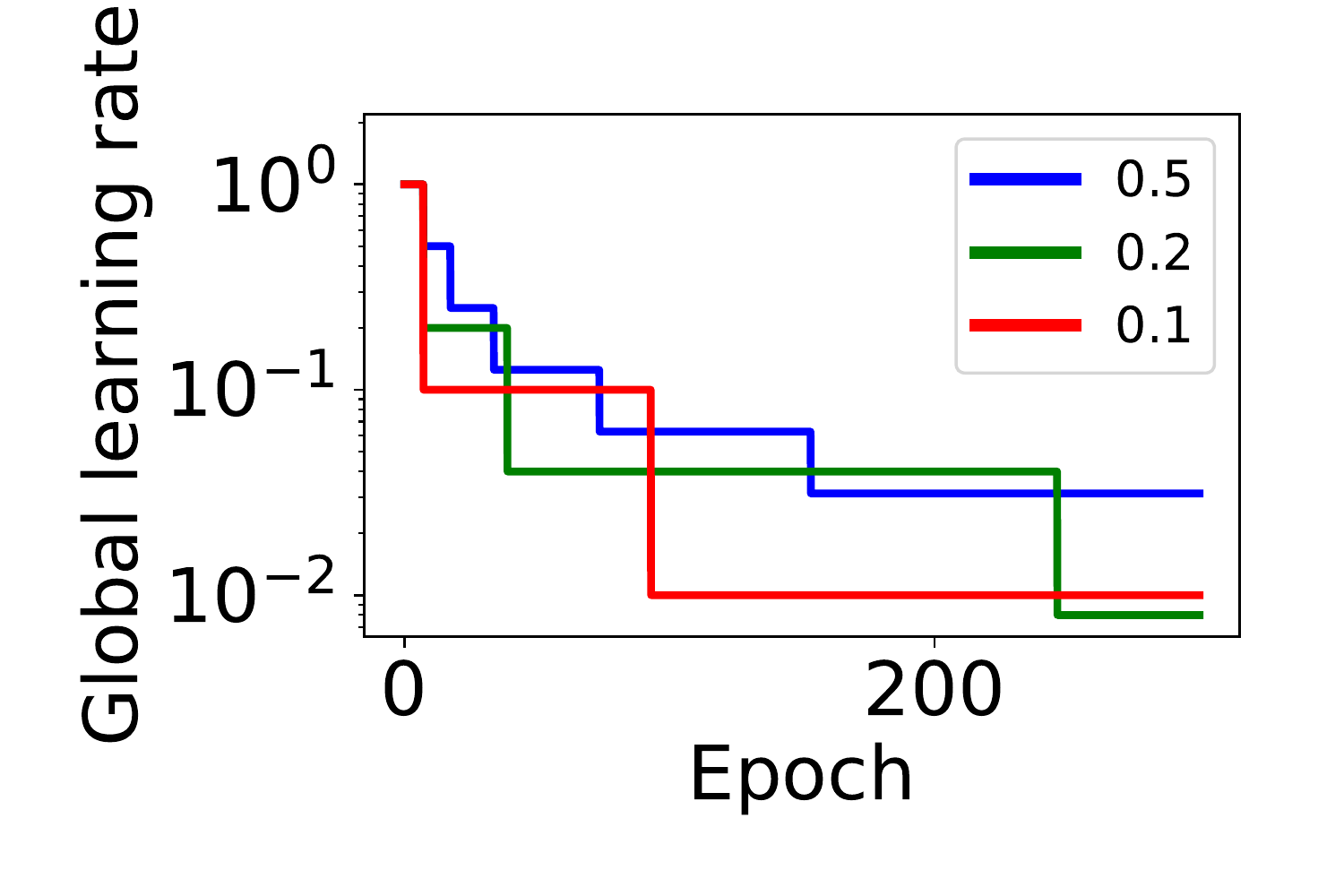}
  \end{subfigure}  
  \caption{Training loss, test accuracy, and learning rate schedule for SASA using different values of $\gamma$, $\delta$ and $\zeta$ around the default $0.2$, $0.02$ and $0.1$. The model is the logistic regression trained on MNIST. Top row: performance for fixed $\gamma=0.2, \zeta=0.1$, and $\delta \in \{0.005, 0.01, 0.02, 0.04\}$. Middle row: performance for fixed $\delta=0.02, \zeta=0.1$, and $\gamma \in \{0.05, 0.1, 0.2\}$. Bottom row: performance for fixed $\gamma=0.2, \delta=0.02$, and $\zeta \in \{0.5, 0.2, 0.1\}$. Qualitatively, increasing $\delta$ and increasing $\gamma$ both cause the algorithm to drop sooner. The value of $\zeta$ does not influence the final performance, as typically the learning rate automatically decays to the same level.}
  \label{fig:mnistsens}
 \end{figure}

\subsection{RNN experiments in Section \ref{sec:experiments}}
For the RNN experiment, we trained the PyTorch word-level language model example
\citeyearpar{pytorchmodel} with 600 hidden units, 600-dimensional embeddings, dropout 0.65, and tied weights. All optimizers also used gradient clipping with 2.0 as the threshold and weight decay 0.0005. We set $\alpha_0$ and $\zeta$ for SGM and SASA to be 2.0 and 0.25, respectively. Because Adam was tuned in this example using the validation set, we also used $\zeta=0.25$ for Adam. The optimal $\alpha_0$ for Adam was 0.5, chosen from the grid $\{0.1, 0.5, 1.0, 2.0, 3.0\}$.

\subsection{Additional experiment: training logistic regression on the MNIST dataset}
We train a logistic regression model on the MNIST dataset with
weight decay $0.0005$. 

\paragraph{Default value performance.} Figure~\ref{fig:lr} shows SASA's performance with default parameters. For this convex optimization problem, SASA and Adam achieve similar performance. SASA uses its default parameters $(\delta, \gamma, \zeta)=(0.02,0.2,0.1)$ and initial $\alpha_0=1.0$. Adam uses its default $(\beta_1, \beta_2) = (0.9, 0.999)$ and its initial learning rate $lr=0.00033$ is obtained from a grid search over $\{0.01, 0.0033, 0.001, 0.00033, 0.0001\}$.

\paragraph{Sensitivity analysis.} As with the experiments on CIFAR-10 and ImageNet, we perturb the relative equivalence threshold $\delta$, the confidence level $\gamma$, and the decay rate $\zeta$ around their default values $(0.2, 0.02, 0.1)$. In Figure~\ref{fig:mnistsens}, the top row shows the performance for fixed $(\gamma, \zeta)=(0.2, 0.1)$ and changing $\delta$. The middle row shows the performance for fixed $(\delta, \zeta)=(0.02, 0.1)$ and changing $\gamma$. The bottom row shows the performance for fixed $(\delta, \gamma)=(0.02, 0.2)$ and changing $\zeta$. The results are the qualitatively the same as in Figures \ref{fig:cifarsens} and \ref{fig:imagenetsens}.

\subsection{Additional experiment: training MaskRCNN on the COCO dataset}
We train a Mask-RCNN model~\cite{he2017mask} with a Feature Pyramid Network (FPN)~\cite{lin2017feature} as a backbone for both object detection and instance segmentation on the the COCO dataset~\cite{lin2014microsoft}. The FPN backbone is based on the ResNet50, and the implementation is based on the MaskRCNN-benchmark repo~\cite{massa2018mrcnn}. In the recommend training setting, the model is trained for 90000 iterations with the SGM optimizer. The learning rate is scheduled to decay by 10 ($\zeta=0.1$) at iteration 60000 and 80000. Readers can refer to \url{https://github.com/facebookresearch/maskrcnn-benchmark/blob/master/configs/e2e_mask_rcnn_R_50_FPN_1x.yaml} for a detailed experiment setup. This hyperparameter setting is carefully tuned to reach the reported performance: object detection mean average precision (bbox-AP) 37.8\% and instance segmentation mean average precision (segm-AP) 34.2\%; see~\url{https://github.com/facebookresearch/maskrcnn-benchmark/blob/master/MODEL_ZOO.md}.

\begin{figure}[tb]
  \centering
  \begin{subfigure}[t]{.245\linewidth}
    \centering
    \includegraphics[width=\linewidth]{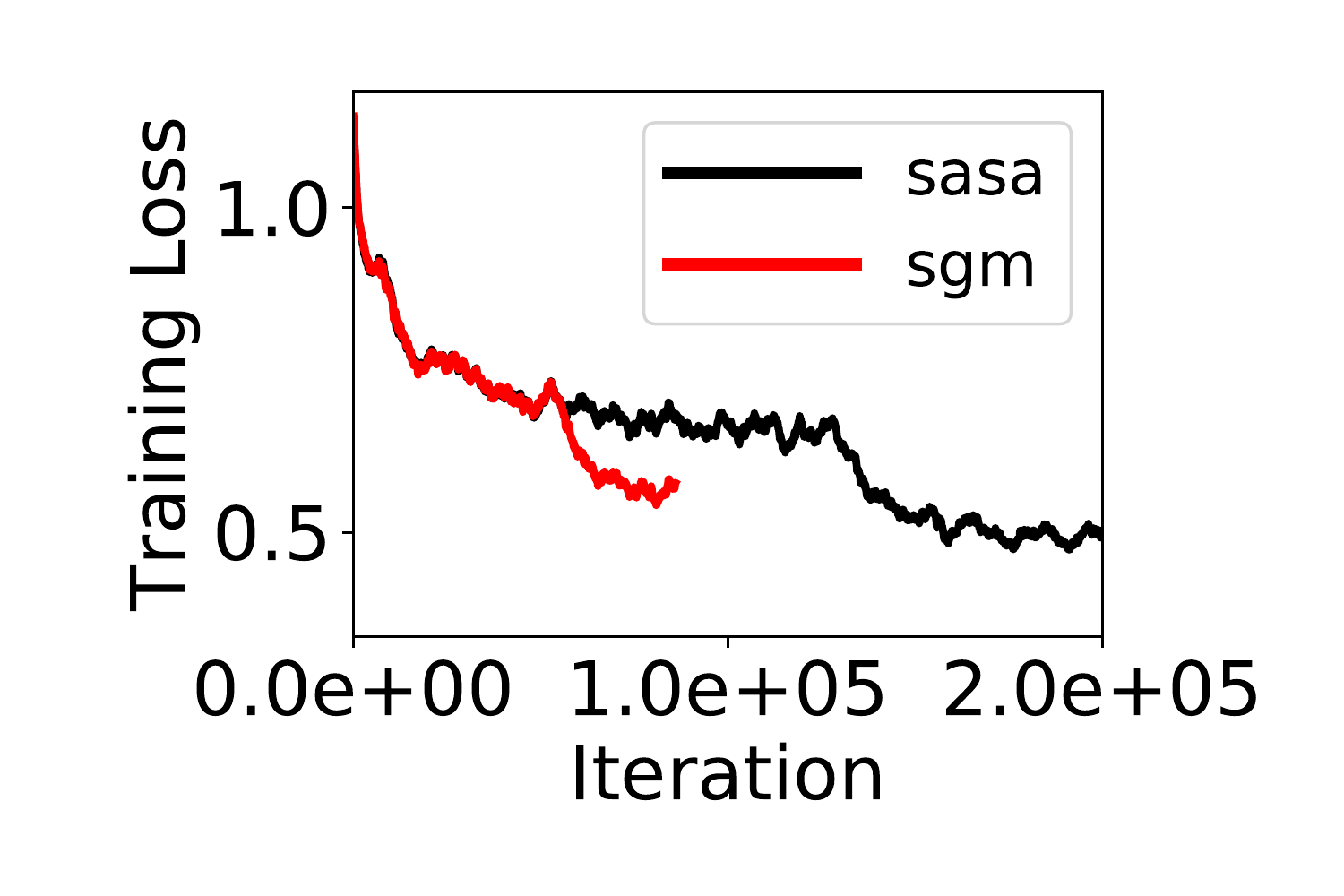}
  \end{subfigure}%
  \begin{subfigure}[t]{.245\linewidth}
    \centering
    \includegraphics[width=\linewidth]{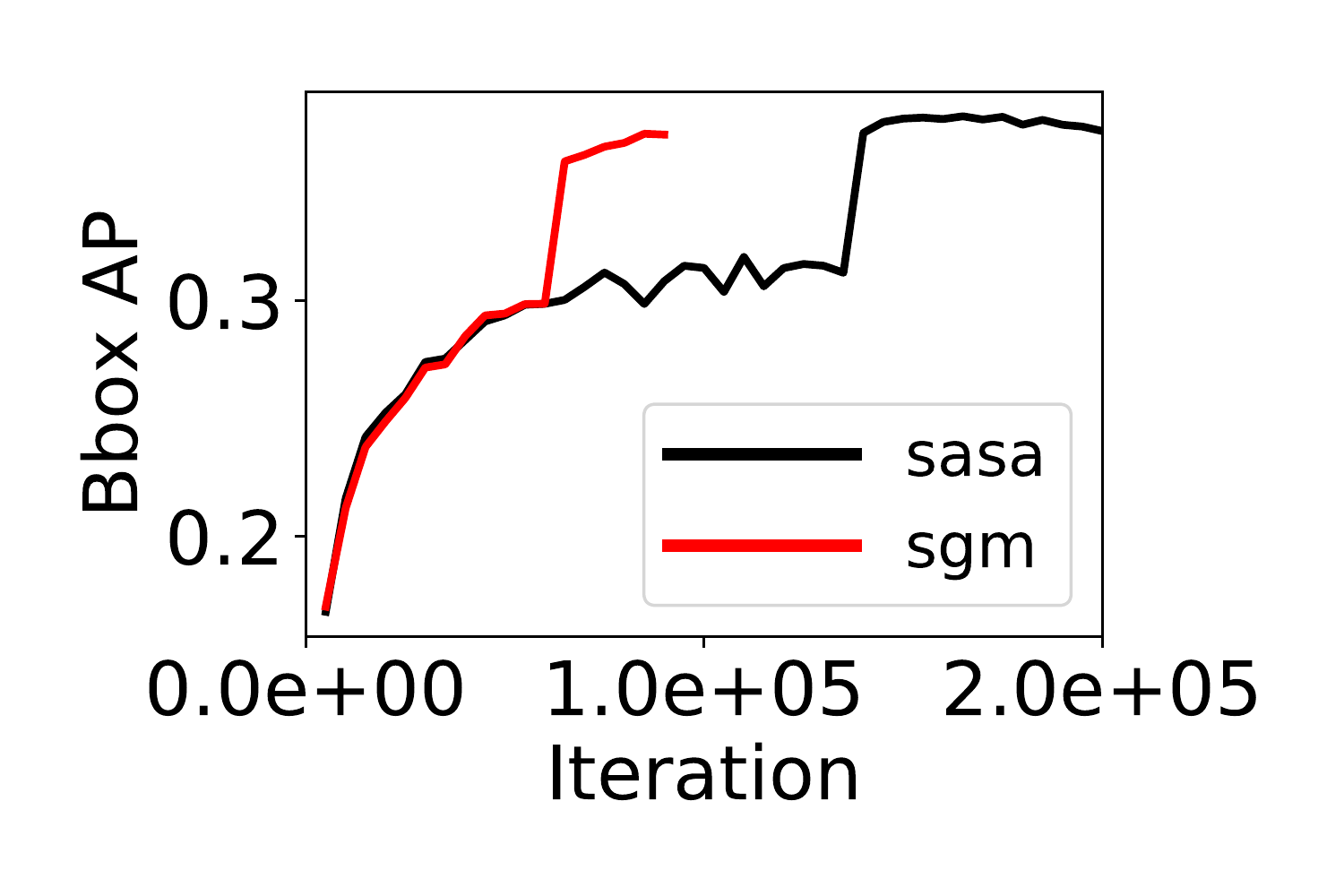}
  \end{subfigure}%
  \begin{subfigure}[t]{.245\linewidth}
    \centering
    \includegraphics[width=\linewidth]{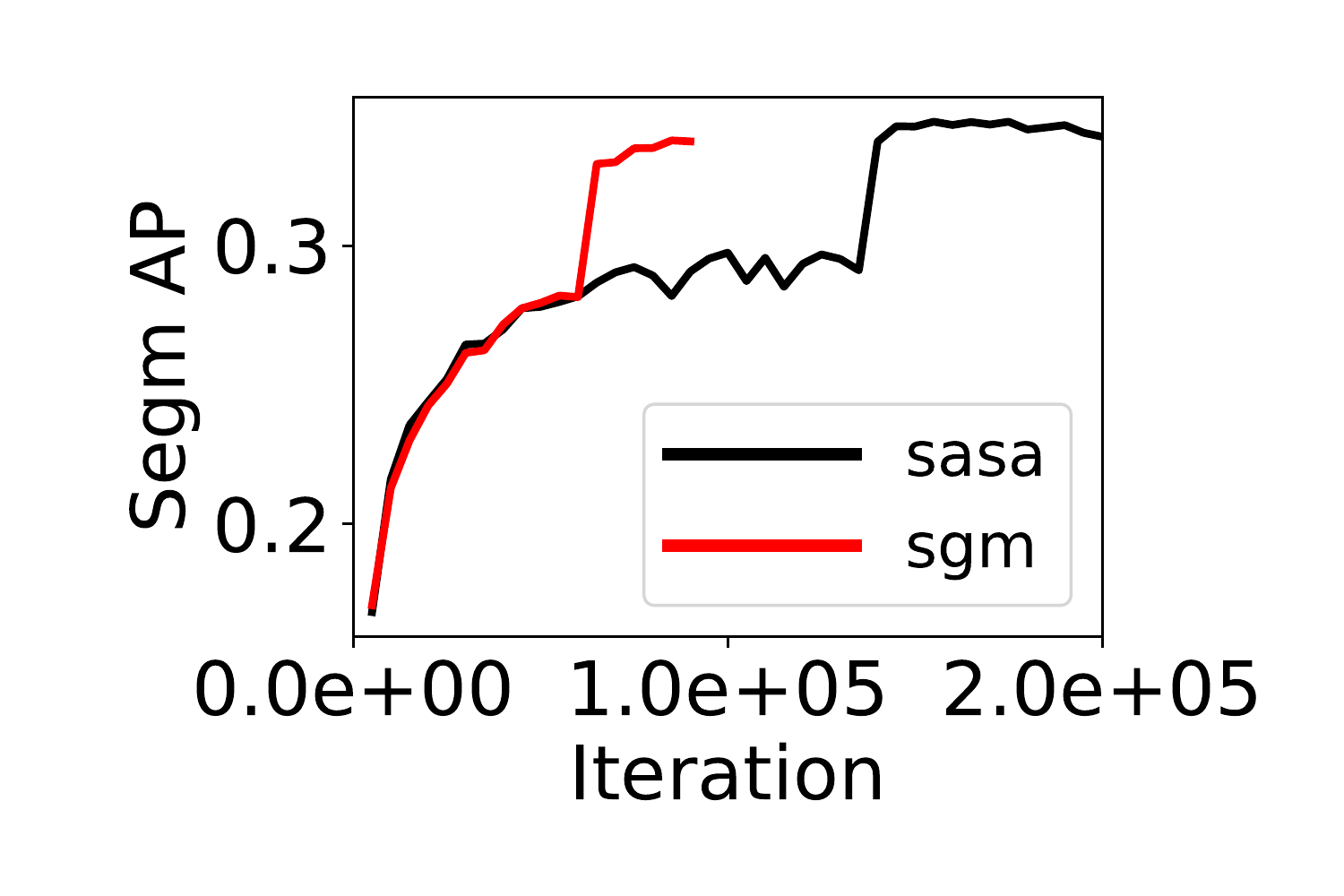}
  \end{subfigure}%
  \begin{subfigure}[t]{.245\linewidth}
    \centering
    \includegraphics[width=\linewidth]{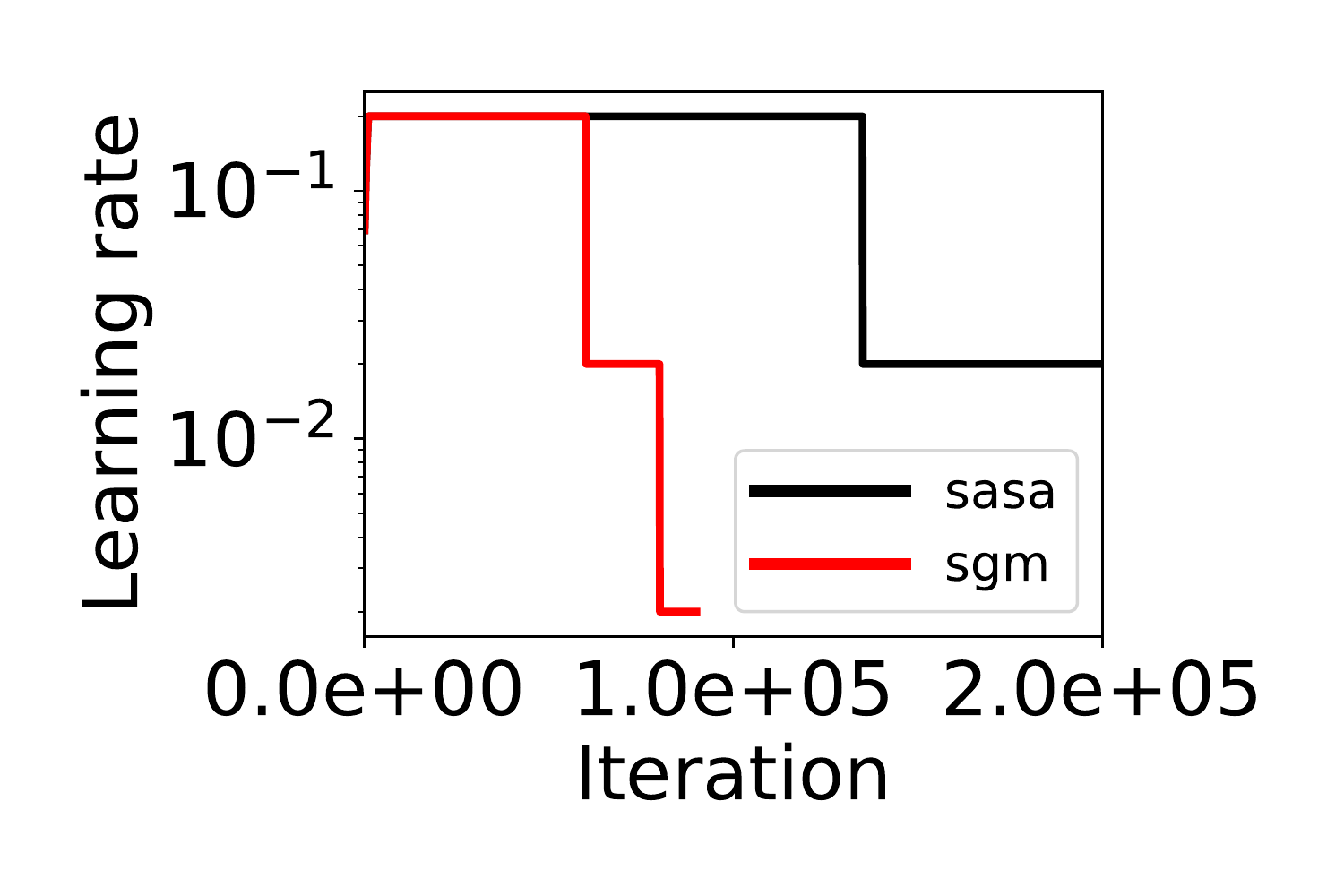}
  \end{subfigure}\\
  \begin{subfigure}[t]{.245\linewidth}
    \centering
    \includegraphics[width=\linewidth]{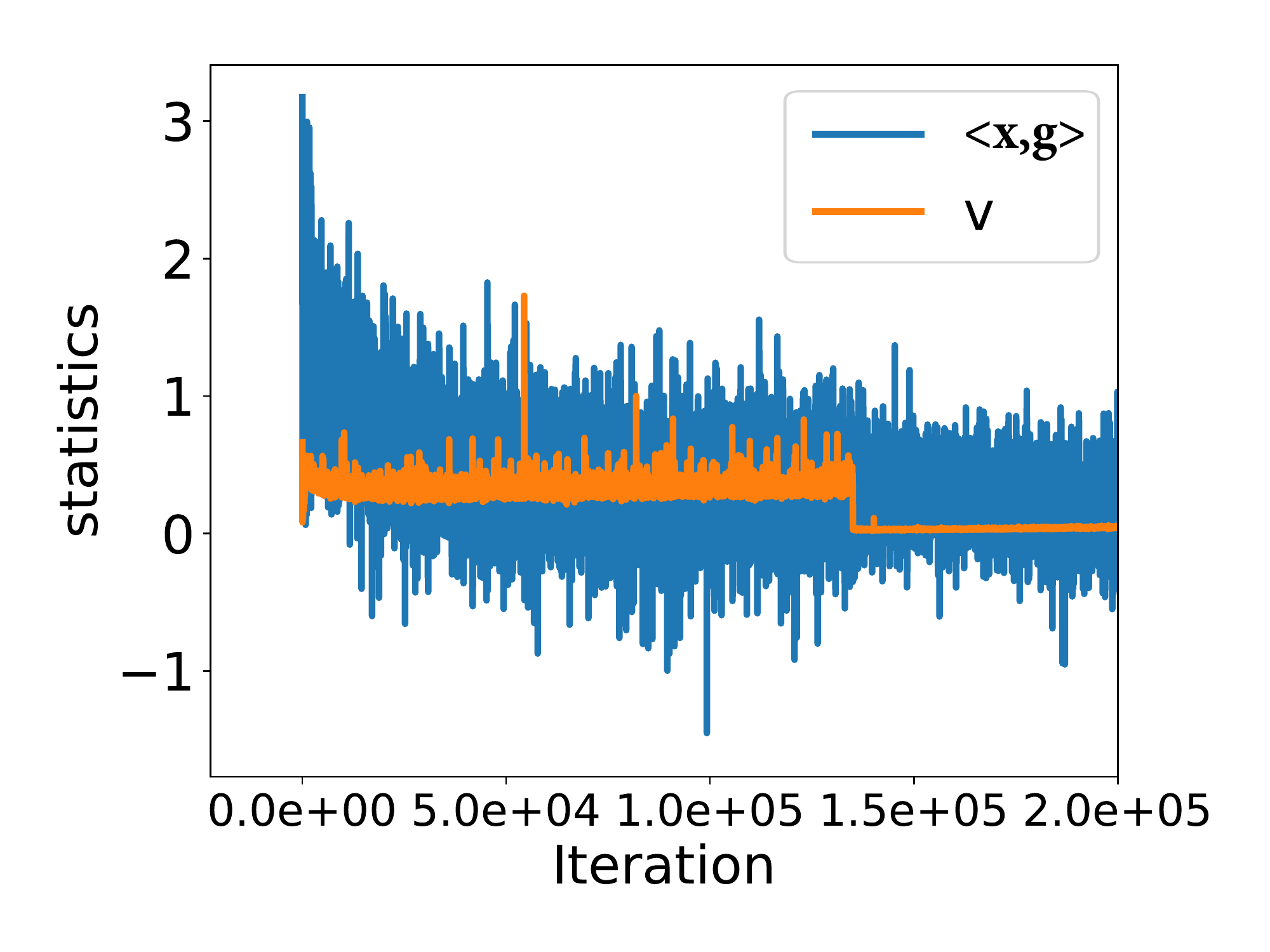}
  \end{subfigure}%
  \begin{subfigure}[t]{.245\linewidth}
    \centering
    \includegraphics[width=\linewidth]{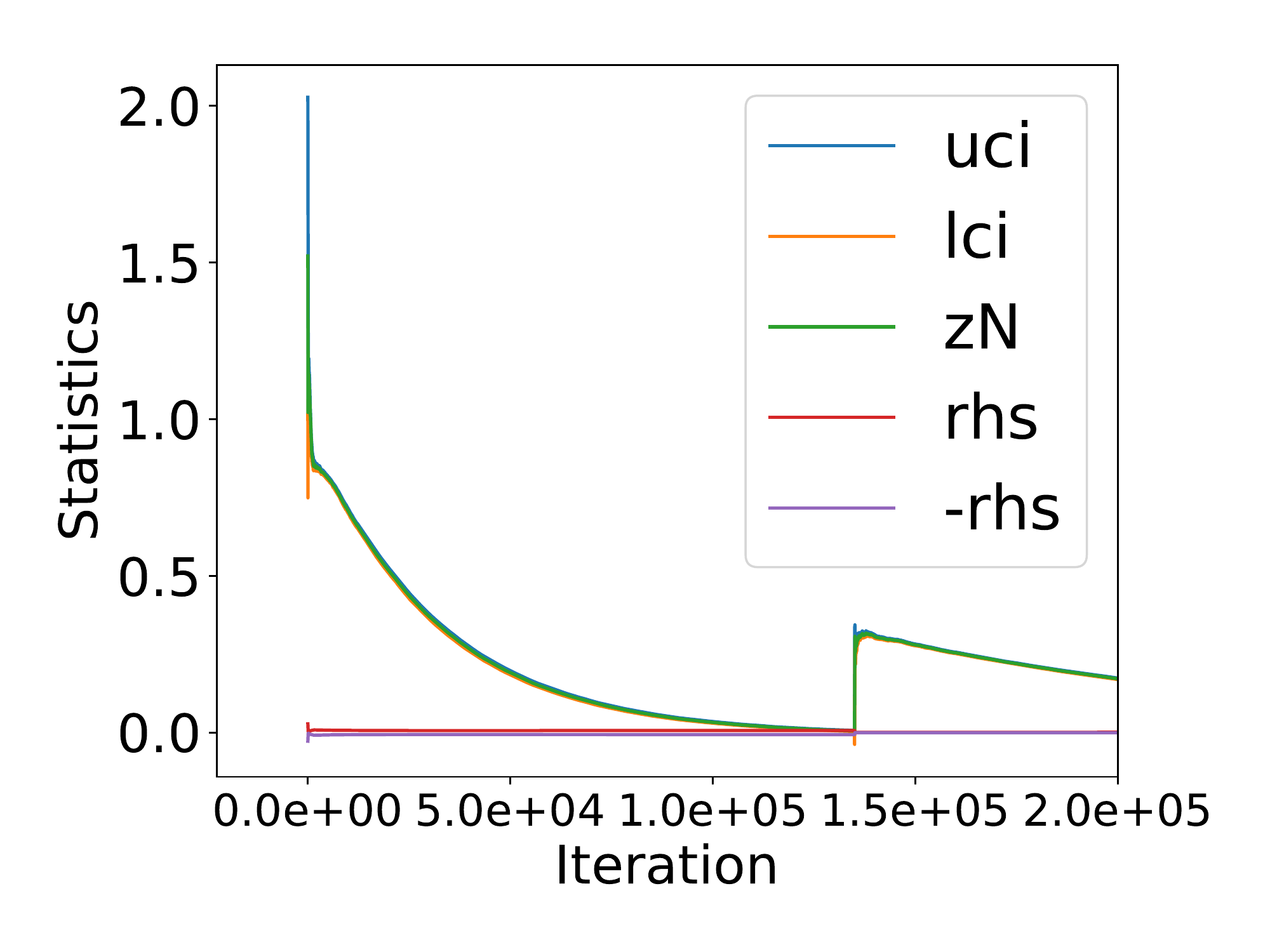}
  \end{subfigure}%
  \begin{subfigure}[t]{.245\linewidth}
    \centering
    \includegraphics[width=\linewidth]{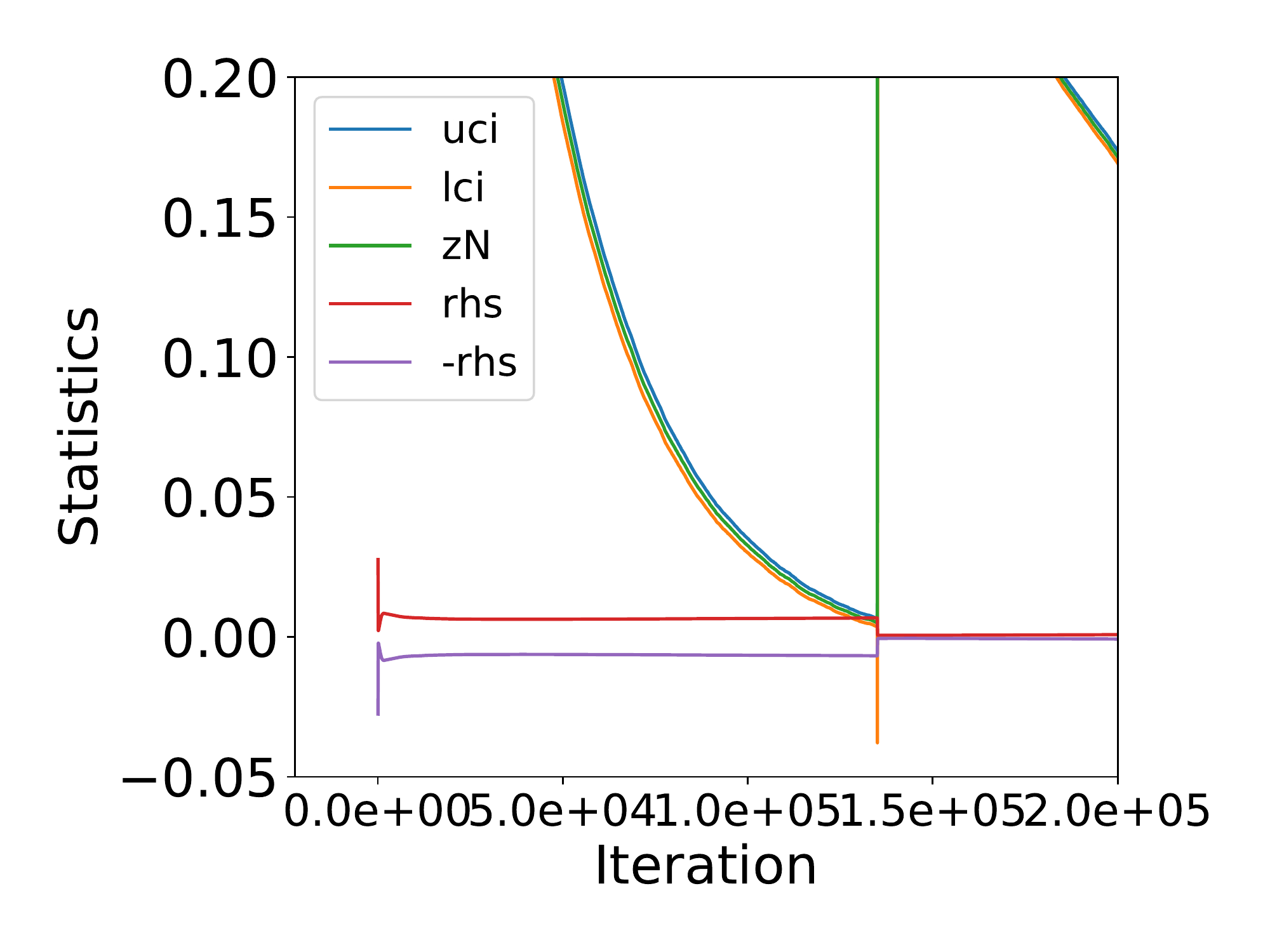}
  \end{subfigure}%
  \begin{subfigure}[t]{.245\linewidth}
    \centering
    \includegraphics[width=\linewidth]{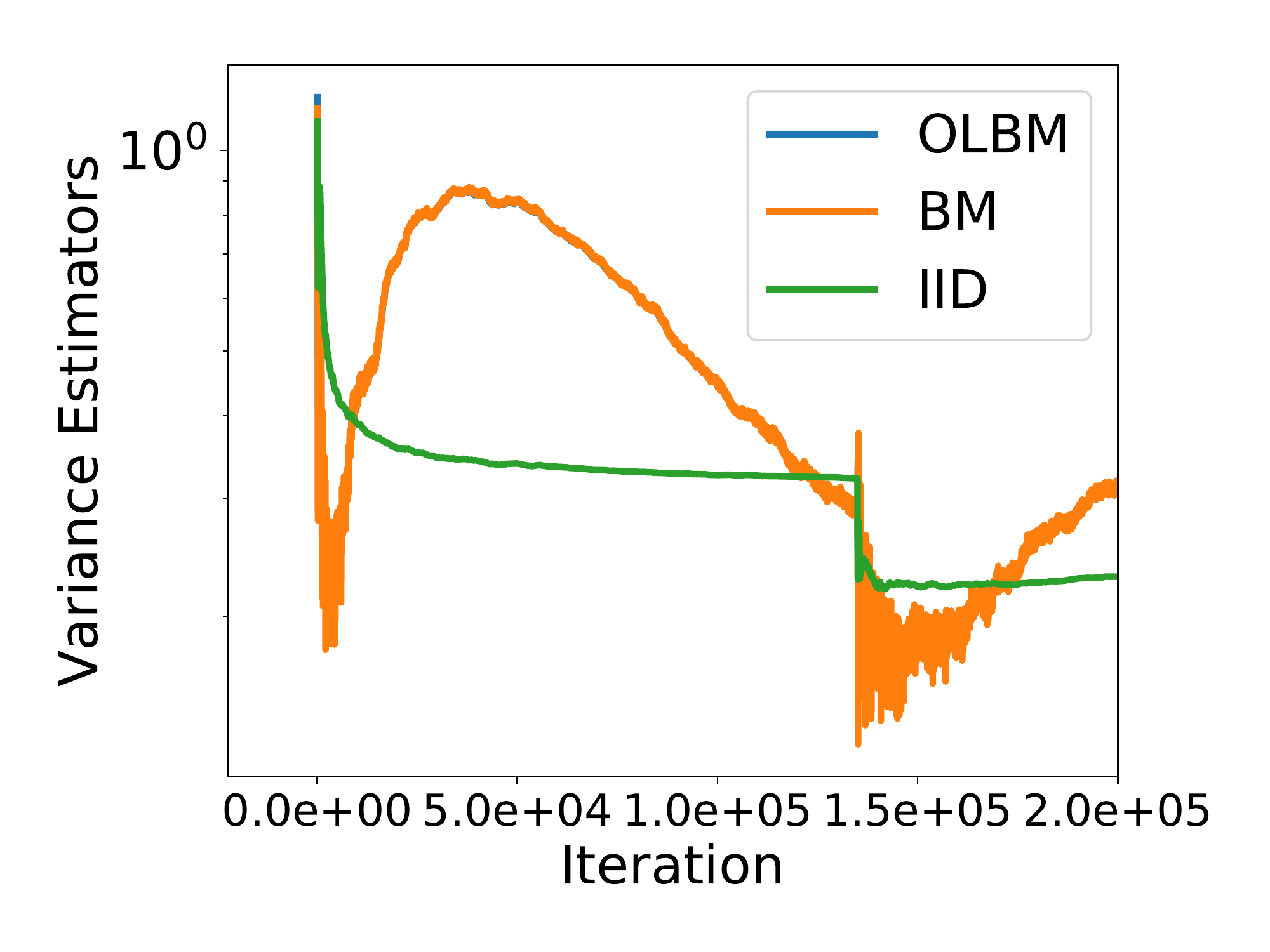}
  \end{subfigure}
  \caption{Top: training loss, test accuracy, and learning rate schedule for SASA and SGM for MaskRCNN training on COCO. Bottom: Evolution of the different statistics for SASA, as in Figures \ref{fig:cifar_stats} and \ref{fig:imagenet_stats}. 
  SASA uses its default parameters $(\delta, \gamma, \zeta)=(0.02,0.2,0.1)$. The SGM is scheduled to decay the learning rate by 10 ($\zeta=0.1$) twice, once at iteration 60000 and once at iteration 90000. SASA takes more iterations (double) to reach a slightly better performance without any parameter tuning.}
  \label{fig:coco}
\end{figure}

\paragraph{Default value performance.} Figure~\ref{fig:coco} shows SASA's performance with default parameters. For this challenging task, SASA achieves a slightly better performance than the hand-tuned SGM optimizer \emph{without any parameter tuning.} However, SASA with default parameters takes longer to achieve comparable performance, because SASA decides to decay the learning rate later than the hand-tuned SGM. Notice that SASA only decreases the learning rate once and already surpasses the performance of the hand-tuned SGM. We believe that if the learning rate is decreased again, the performance can be further improved. However, when the training reaches the maximum iteration 200000, the training loss is still constantly decreasing, so the dynamics have not reached a stationary distribution. This prevents SASA from decreasing its learning rate. Meanwhile, the model starts to overfit at this stage, which suggests that we should either decrease the learning rate or stop the training. As mentioned in Section 5, a combination of stationary detection (SASA) and overfitting detection is a promising direction toward a fully automated optimizer.

\begin{figure}[t]
  \centering
  \begin{subfigure}[t]{.23\linewidth}
    \centering
    \includegraphics[width=\linewidth]{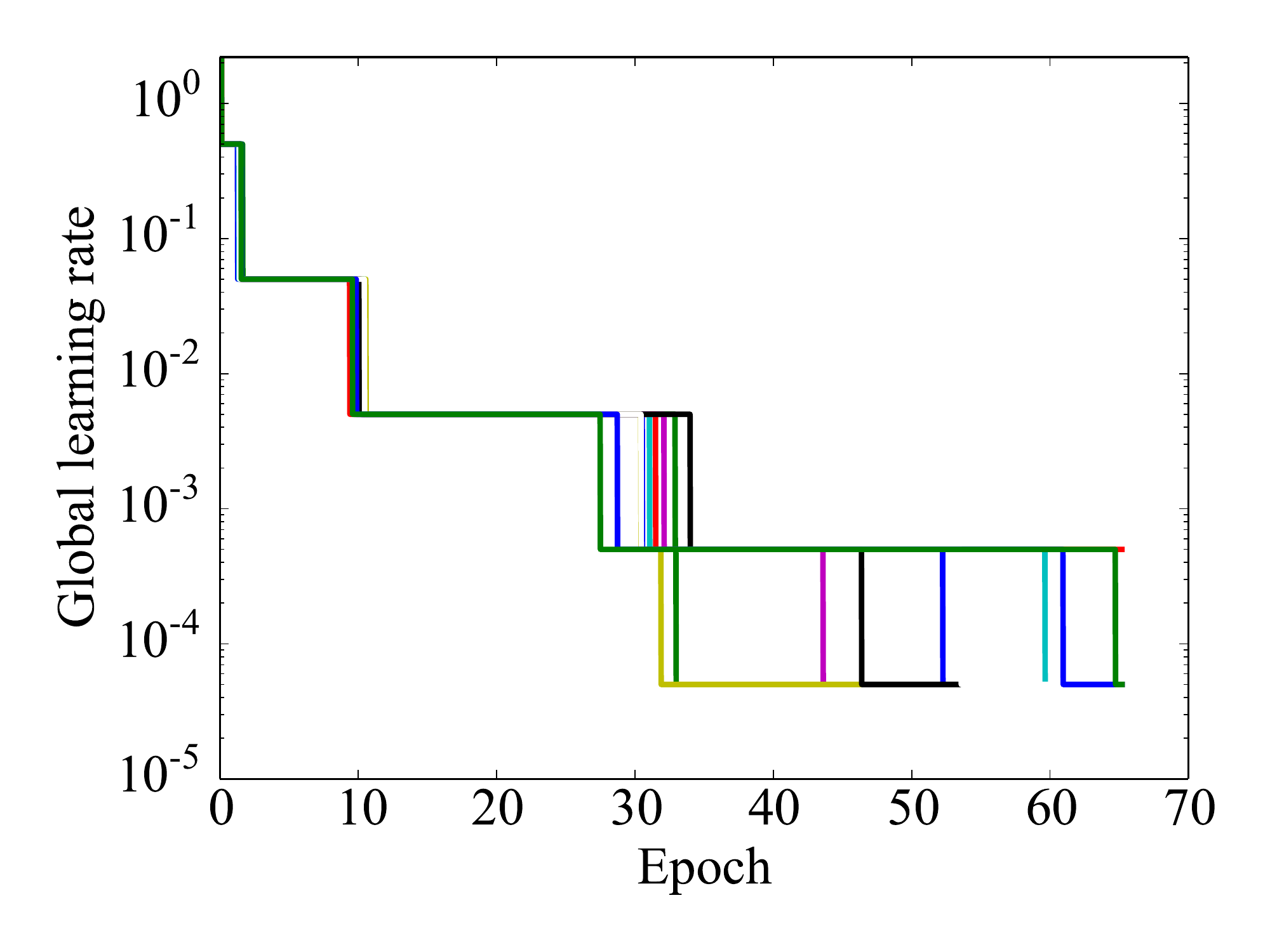}
    \subcaption{}
  \end{subfigure}%
  \begin{subfigure}[t]{.23\linewidth}
    \centering
    \includegraphics[width=\linewidth]{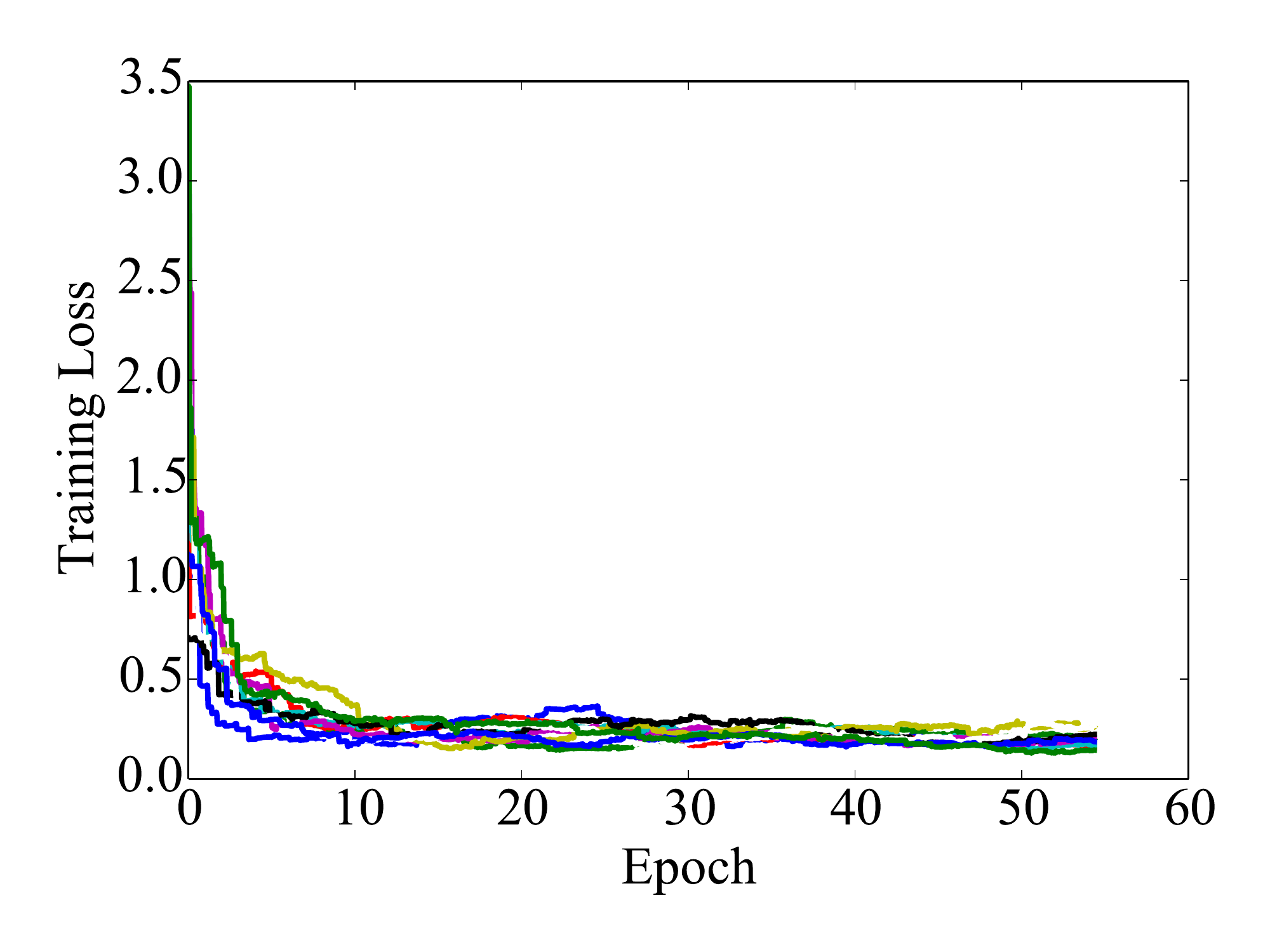}
    \subcaption{}
  \end{subfigure}%
  \begin{subfigure}[t]{.23\linewidth}
    \centering
    \includegraphics[width=\linewidth]{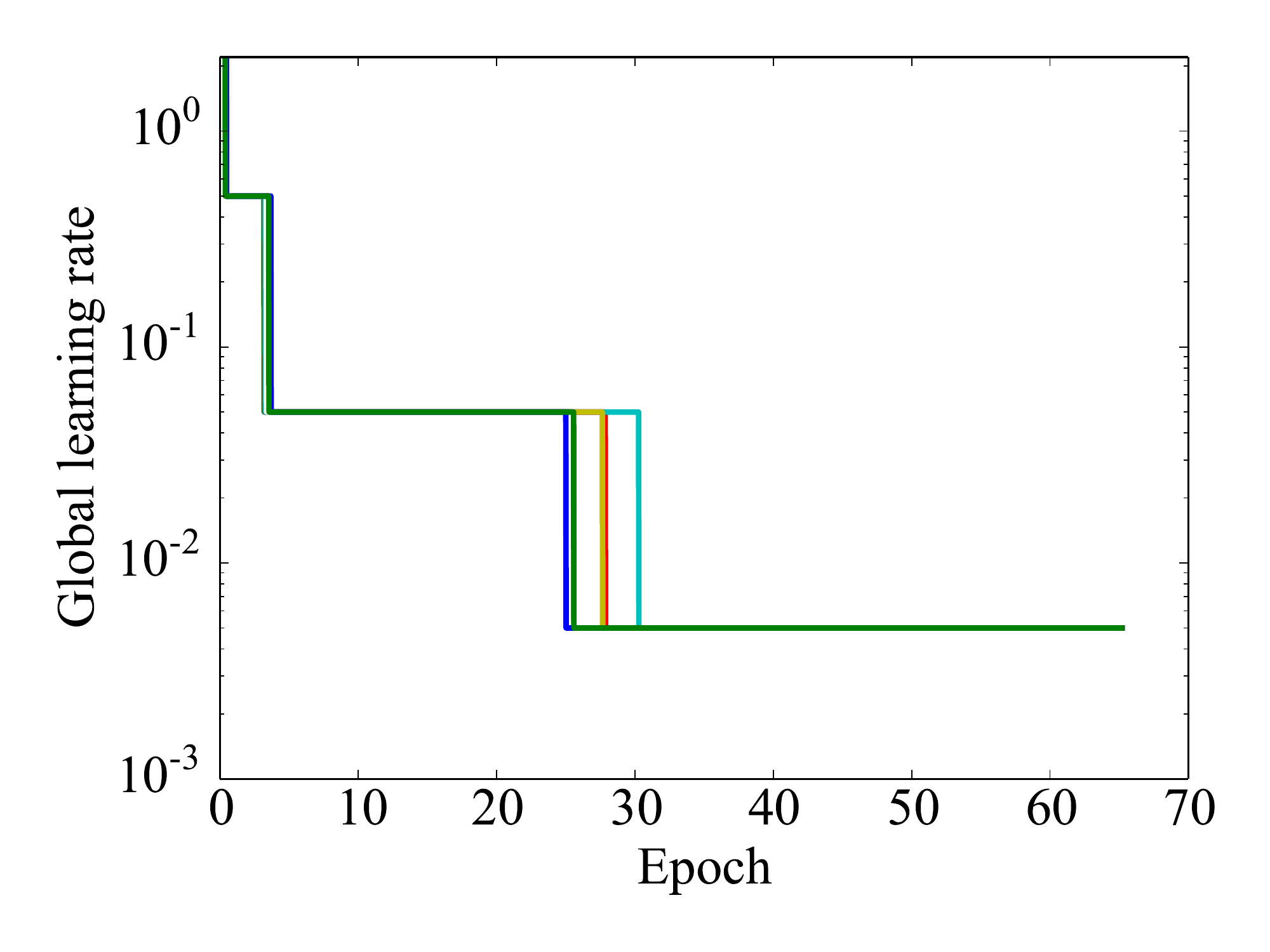}
    \subcaption{}
  \end{subfigure}%
  \begin{subfigure}[t]{.23\linewidth}
    \centering
    \includegraphics[width=\linewidth]{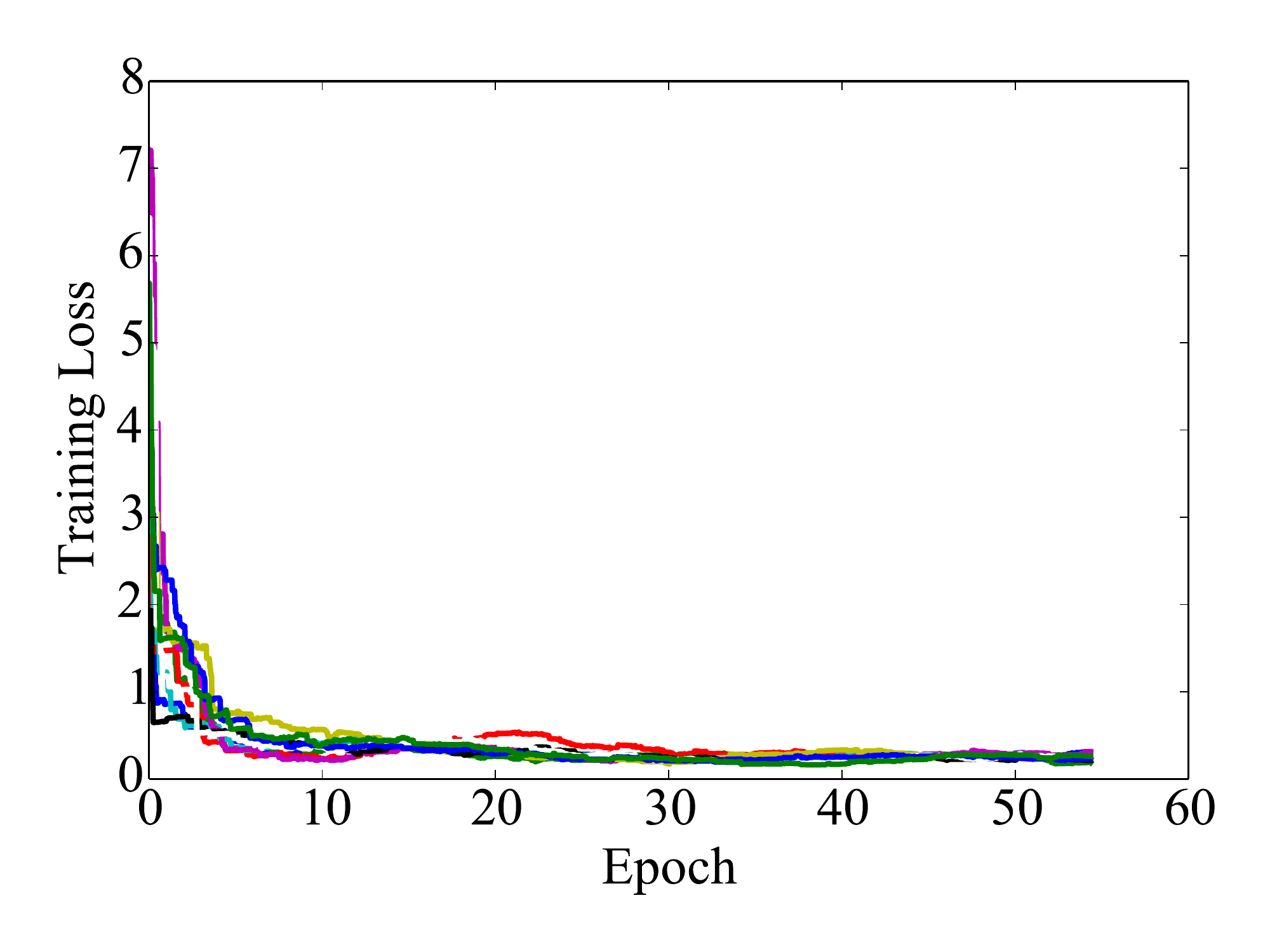}
    \subcaption{}
  \end{subfigure}\\
  \caption{Variance in learning rate schedule and training loss for the two tests \eqref{eqn:yaida-test} (Panels (a)-(b)) and \eqref{eqn:our-test} (Panels (c)-(d)) for a logistic regression model on MNIST, using batch size one and test frequency $M=100$ iterations. Ten independent runs are shown for each method. With the same value of $\delta$, the variance in the learning rate schedule for Yaida's method \eqref{eqn:yaida-test} is much higher.}
  \label{fig:mnist_variance}
  \end{figure}

\begin{figure}[tb]
    \centering
  \begin{subfigure}[t]{.5\linewidth}
    \centering
    \includegraphics[width=\linewidth]{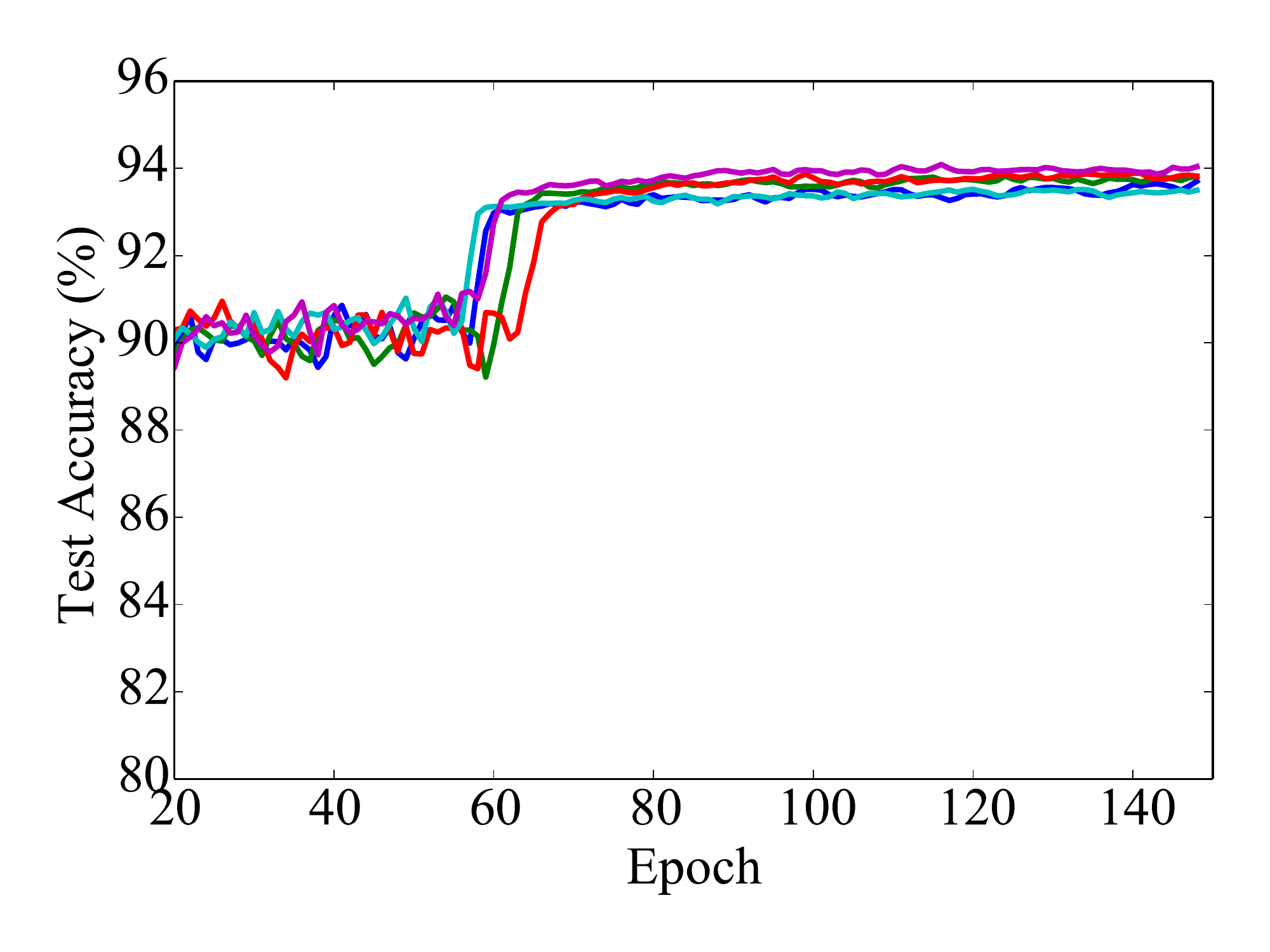}
  \end{subfigure}%
  \begin{subfigure}[t]{.5\linewidth}
    \centering
    \includegraphics[width=\linewidth]{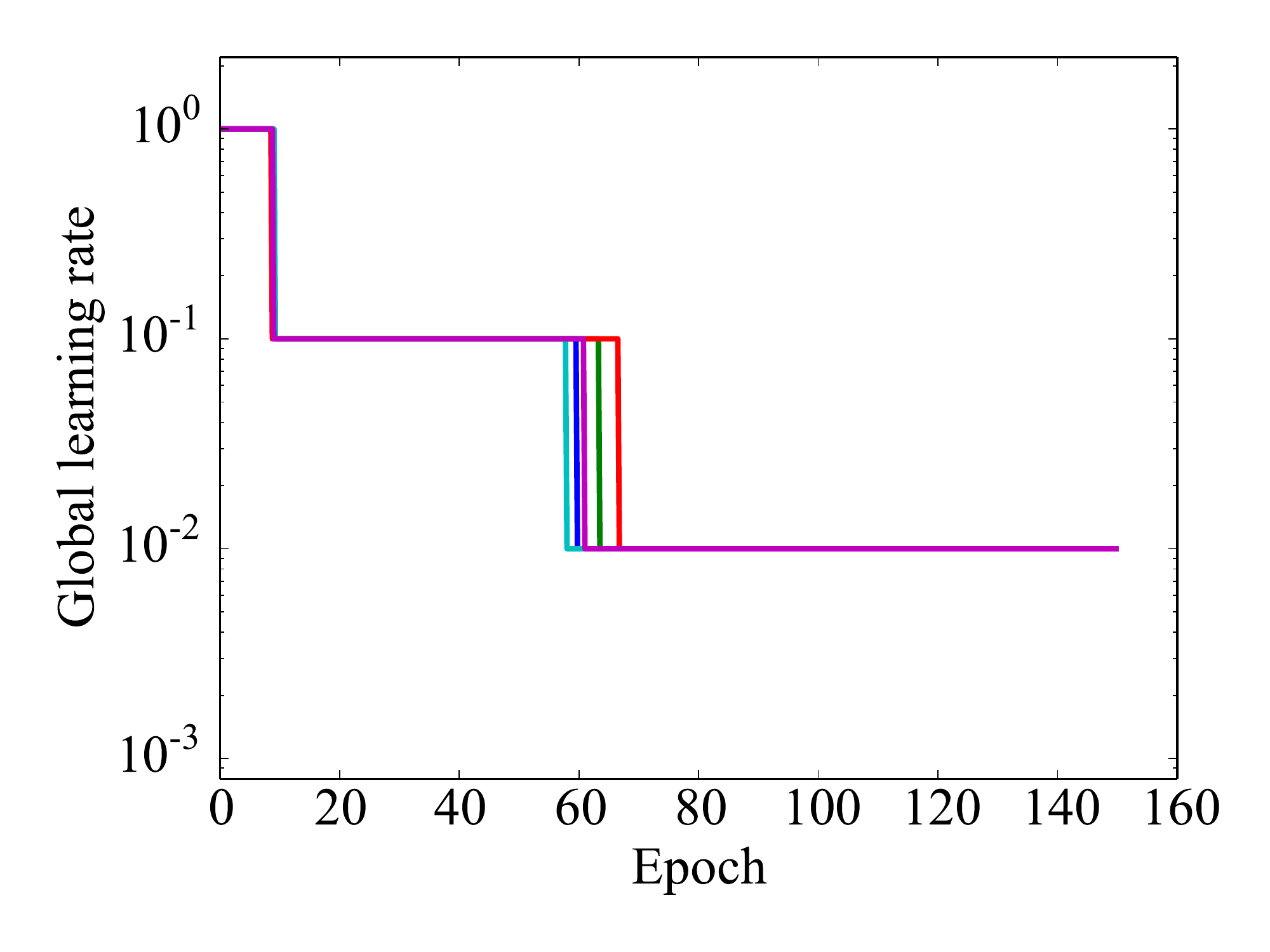}
  \end{subfigure}\\
  \begin{subfigure}[t]{.5\linewidth}
    \centering
    \includegraphics[width=\linewidth]{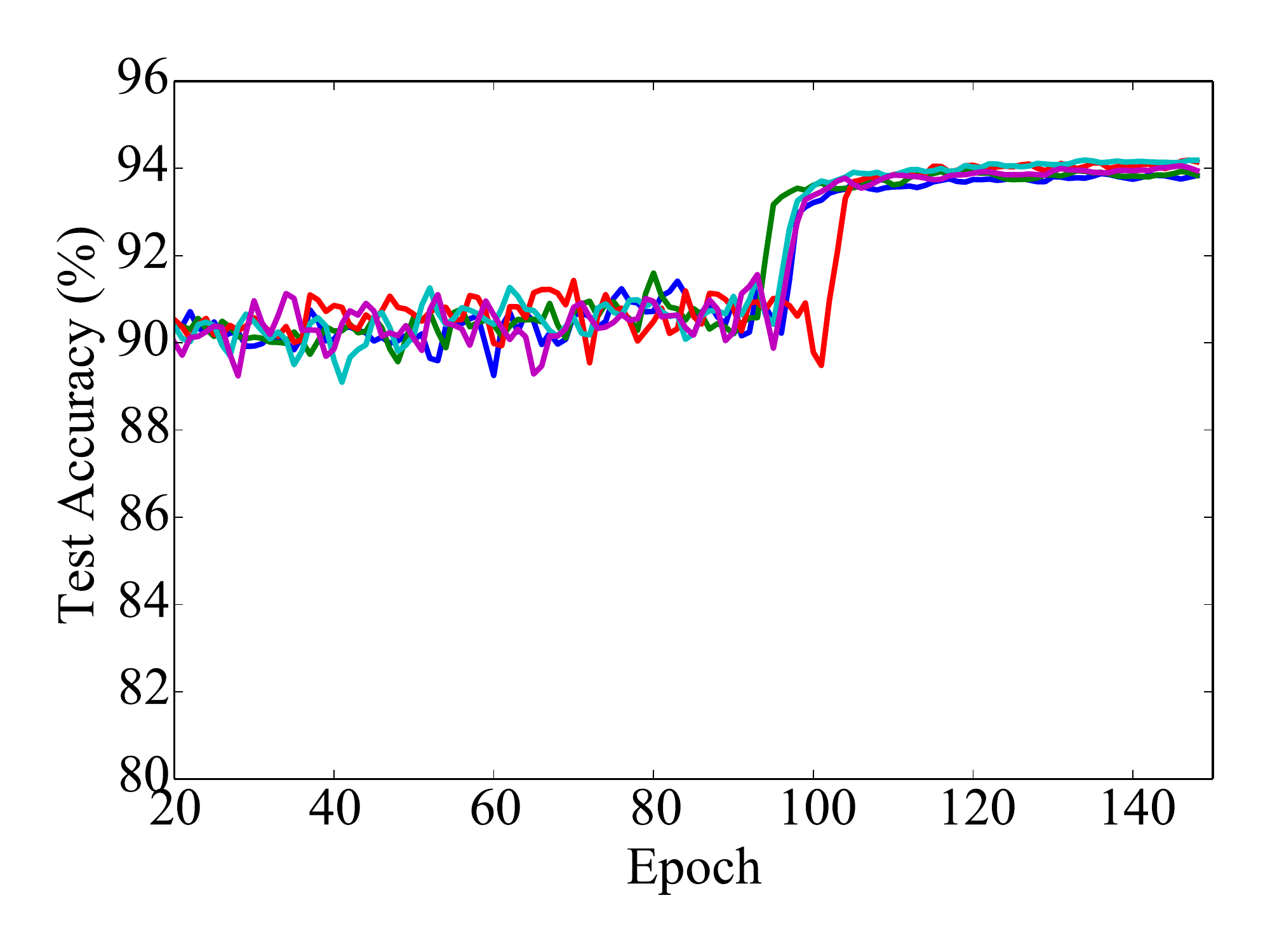}
  \end{subfigure}%
  \begin{subfigure}[t]{.5\linewidth}
    \centering
    \includegraphics[width=\linewidth]{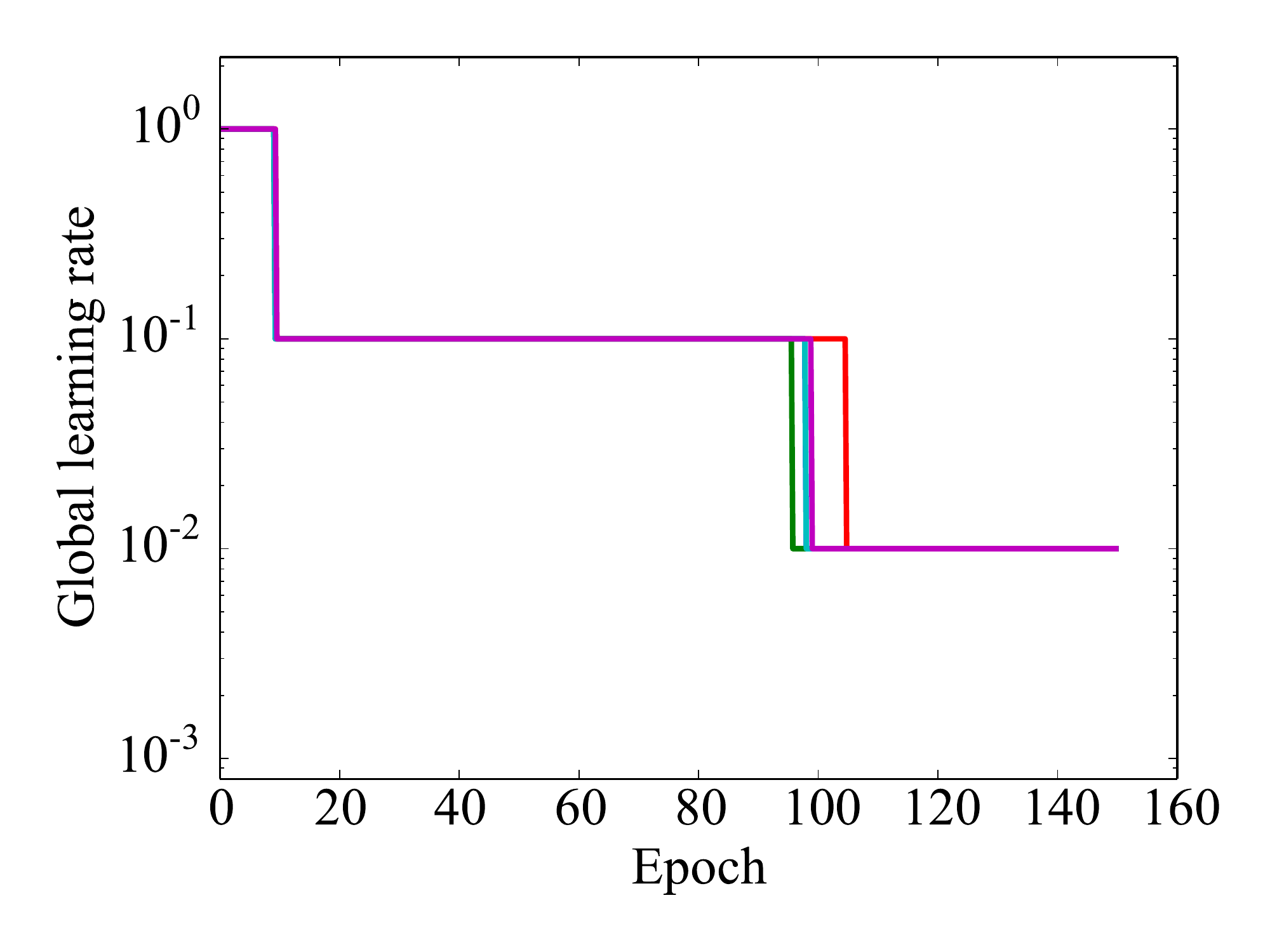}
  \end{subfigure}
  \caption{Test accuracy and learning rate schedule when using Yaida's ratio test \eqref{eqn:yaida-test} (top row) and our statistical test \eqref{eqn:our-test} (bottom row) with $M=10$, $\delta=0.02$, and with SASA using $\gamma=0.2$. The standard deviation of both the learning rate drops and the best test set performance is higher for Yaida's test: 3.2 epochs vs 2.9 epochs, and 0.17\% vs 0.12\%. The mean performance of the statistical test is also marginally higher, 94.08\% vs 93.84\% test accuracy.}
  \label{fig:changingM}
\end{figure}

\section{Comparison with Yaida's test}
\label{apdx:vs-yaida}
The variance experiment in Figure \ref{fig:variancefig} can be interpreted as showing that for a fixed testing frequency $M$, the statistical procedure \eqref{eqn:our-test} is more robust to changes in the noise level of the samples than the heuristic test \eqref{eqn:yaida-test}. We essentially repeat this experiment in Figure \ref{fig:mnist_variance}, which shows the performance of the two testing methods \eqref{eqn:yaida-test} and \eqref{eqn:our-test} on a logistic regression model trained on MNIST. We used the same procedure as in all other logistic regression experiments, except with batch size one. We test the statistics every $M=100$ iterations and plot the results of ten independent runs for each method, using a fixed $M$ as in Figure \ref{fig:variancefig}. While the final training and test loss for the two methods are similar (92.7\% $\pm 0.16$ for SASA, 92.7\% $\pm 0.2$ for \eqref{eqn:yaida-test}), the variance in the learning rate schedules for Yaida's method is dramatically higher. On strongly convex problems, this may not cause poor performance, but as shown in Figure \ref{fig:variancefig}, it can cause dramatic results in more general settings. This experiment gives a further indication that when using a fixed test frequency $M$, explicitly accounting for the variance in $\bar{z}_N$, as in SASA, is critical for robust performance. Finally, Figure \ref{fig:changingM} shows a complementary result on CIFAR-10: even when the batch size is large (128), the statistical approach is less sensitive to using a small testing frequency $M$. While this effect is on a much smaller scale than the others, it indicates that Yaida's heuristic (performing the test \eqref{eqn:yaida-test} once per epoch) is more sensitive than SASA to the choice of the testing frequency.

Figure \ref{fig:variancefig}, Figure \ref{fig:mnist_variance}, and Figure \ref{fig:changingM} indicate that the statistical test is more robust to changes in noise and testing frequency than Yaida's deterministic ratio test. However, Figure \ref{fig:changingM} indicates that this method can obtain similar (albeit less robust) performance on large deep learning datasets, and our practical results can be taken more generally as large-scale evidence that methods for detecting stationarity have good practical performance when used as adaptive optimizers. Still, our formulation recovers Yaida's when $\gamma=1$, heuristics like "test once per epoch" are not always available---such as in an online training setting---so robustness to the test frequency $M$ is desirable, and we have demonstrated that SASA is less sensitive to noise in several regimes, such as small batch size and high test frequency. For these reasons we believe SASA will be more robust in practice, and we hope it leads to more research on using statistical tests in optimization.

\section{Generalized Pflug condition and comparison with Yaida's condition}
\label{sec:stat-quadratic}
In this section, we provide a generalization of Pflug's stationary condition to the case of SGM for quadratic functions. We also compare the two stationary conditions (Pflug's and Yaida's) and show that Yaida's stationary condition works much better for practical machine learning problems.

\subsection{Derivation of the generalized Pflug stationary condition}
As in \citep{Pflug90nonasymptotic,MandtHoffmanBlei2017}, the derivation is based on two assumptions:
\begin{enumerate}
    \item The quadratic objective assumption:
    \begin{equation}\label{eqn:quadratic}
        F(x) = (1/2)x^T A x, 
    \end{equation}
    where $A$ is positive definite.
    \item The i.i.d. additive noise assumption:
    \begin{equation}\label{eqn:additive-noise}
        \gk=\nabla F(\xk) + \xik,
    \end{equation}
    where $\xik$ is independent of $\xk$, and for all $k\geq 0$ satisfies
    \begin{equation}\label{eqn:noise-mean-cov}
        \E\bigl[\xik\bigr]=0, \qquad \E\bigl[\xik(\xik)^T\bigr] = \Sigma_\xi.
    \end{equation}
\end{enumerate}
\citet{MandtHoffmanBlei2017} observe that this noise assumption
can hold approximately when $\alpha$ is small and the dynamics of SGM are approaching
stationarity around a local minimum.

For the dynamics of SGM with constant $\alpha$ and $\beta$, i.e., \eqref{eqn:sgm-fixed}, the sequence $\{(x^k, d^k, g^k)\}$ is assumed to converge to a stationary distribution $\pi(x,d,g)$, as we defined in Section~\ref{sec:stationary-condition}. We denote $x$'s covariance matrix under the stationary distribution as
\begin{equation}\label{eqn:Sigma-x}
    \Sigma_x = \lim_{k\to\infty} \E\bigl[\xk(\xk)^T\bigr].
\end{equation}
The following theorem characterizes the dependence of $\Sigma_x$ on~$A$, $\alpha$ and~$\beta$. It also derives an asymptotic expression of $\E_{\pi}[\langle g, d \rangle]$ in terms of~$A$, $\alpha$ and~$\beta$.

\begin{theorem}\label{thm:stationarity}
Suppose $F(x) = (1/2)x^T A x$, where $A$ is 
positive definite with maximum eigenvalue $L$,
and $\gk$ satisfies~\eqref{eqn:additive-noise} and~\eqref{eqn:noise-mean-cov}.
If we choose $\alpha\in(0,1/L)$ and $\beta\in[0,1)$ in~\eqref{eqn:sgm-fixed}, 
then $\Sigma_x$ defined in~\eqref{eqn:Sigma-x} exists. Moreover, we have
\begin{equation}\label{eqn:Sigma-x-1st-order}
    A\Sigma_x + \Sigma_x A = \alpha \Sigma_\xi + O(\alpha^2)
\end{equation}
and
\begin{equation}\label{eqn:gkp-dk-correlation2}
    \E_{\pi}[\langle g, d \rangle]
    = -\frac{\alpha(1-\beta)}{2(1+\beta)} \tr(A\Sigma_\xi) + O(\alpha^2).
\end{equation}
\end{theorem}

Theorem~\ref{thm:stationarity} states that when~$\alpha$ is small,
we can approximate $\Sigma_x$ by solving the linear equation
$A\Sigma_x + \Sigma_x A = \Sigma_\xi$.
Moreover, the variance $\tr(\Sigma_x)$ decreases to zero as $\alpha\to 0$.
It is well known that larger~$\beta$ often leads to faster transient 
convergence when SGM is far away from a local minimum. 
According to~\eqref{eqn:Sigma-x-1st-order}, it does not
affect the covariance in steady state, especially for small~$\alpha$.
%

Equation~\eqref{eqn:gkp-dk-correlation2} implies that for
small~$\alpha$, the vectors $\gk$ and $\dk$ will eventually have
negative correlation.
Interestingly, their correlation is less negative for larger~$\beta$.

Assuming ergodicity, $\E_{\pi}[\langle g, d \rangle]$ can be evaluated by the history average
\begin{equation}\label{eqn:estimate-gkp-dk}
\E_{\pi}\bigl[\langle\gk, \dk\rangle\bigr]
\approx \frac{1}{N}\sum_{i=k+1}^{k+N} \langle g^{i}, d^i \rangle,
\end{equation}
where~$N$ can be chosen to control the quality of estimation.
If an online estimate of $\tr(A\Sigma_\xi)$ is also available,
then we can check if the relation established in~\eqref{eqn:gkp-dk-correlation2}
holds in a statistical sense, which serves as a test of stationarity.

Since we do not assume any knowledge of~$A$ or~$\Sigma_\xi$, it
can be hard to estimate $\tr(A\Sigma_\xi)$ using simple statistics.
To address this challenge, \citet{Pflug83} constructed a novel scheme
that requires three stochastic gradients at each iteration.
Specifically, at each iteration~$k$, we first compute two stochastic
gradients $\gk_1$ and $\gk_2$ of~$F$ at $\xk$, and we let $r^k =
(\gk_1 - \gk_2)/2$ (in a data-parallel training setting, $\gk_1$ and
$\gk_2$ can be computed from separate processing units, and thus can
be obtained without extra delay).  Next, we let $\xtk = \xk + \alpha
r^k$ and compute another stochastic gradient $\tilde{g}^k$ of~$F$ at
$\xtk$.  Then, it can be shown \citep{Pflug83} that
\begin{equation}\label{eqn:A-Sigma-estimate}
    \E[\langle r^k, \tilde{g}^k\rangle] = \frac{\alpha}{2}\tr(A\Sigma_{\xi}).
\end{equation}
We can thus obtain an online estimate of $\tr(A\Sigma_\xi)$ using the
running average of $\langle r^k, \tilde{g}^k\rangle$ in a similar way
to~\eqref{eqn:estimate-gkp-dk}.

As suggested by \citet{Pflug83}, a less wasteful use of the
stochastic gradients is to define $\gk=(\gk_1+\gk_2)/2$ and use it
in~\eqref{eqn:sgm-fixed}. This averaging reduces the covariance of
$\gkp$ and $\dkp$ by a factor of $1/2$, which together
with~\eqref{eqn:gkp-dk-correlation2} implies
\begin{equation} \label{eqn:gkp-dk-negative}
\E_{\pi}\bigl[\langle g, d \rangle\bigr]
\approx -\frac{\alpha(1-\beta)}{4(1+\beta)}\tr(A\Sigma_{\xi}),
\end{equation}
where we still use $\pi$ to denote the new stationary condition.
Combining~\eqref{eqn:A-Sigma-estimate} and~\eqref{eqn:gkp-dk-negative},
we conclude that for small~$\alpha$,
\begin{equation}\label{eqn:expected-test}
\E_{\pi}\bigl[\langle g, d \rangle\bigr]
\approx -\frac{1-\beta}{2(1+\beta)}\E\bigl[\langle r^k,\tilde{g}^k\rangle\bigr]
\end{equation}
holds if the dynamics~\eqref{eqn:sgm-fixed} reach stationarity. Both sides of \eqref{eqn:expected-test} can be estimated by the history average during the training, thanks to ergodicity.

Unfortunately, evaluating this estimator requires 33\% more training iterations than regular SGM due to the stochastic gradients used to compute the point $\tilde{x}^k$.

\subsection{Comparing stationary conditions}
\label{apdx:comparing}
\begin{figure}[t]
  \centering
  \begin{subfigure}[t]{.25\linewidth}
    \centering
    \includegraphics[width=\linewidth]{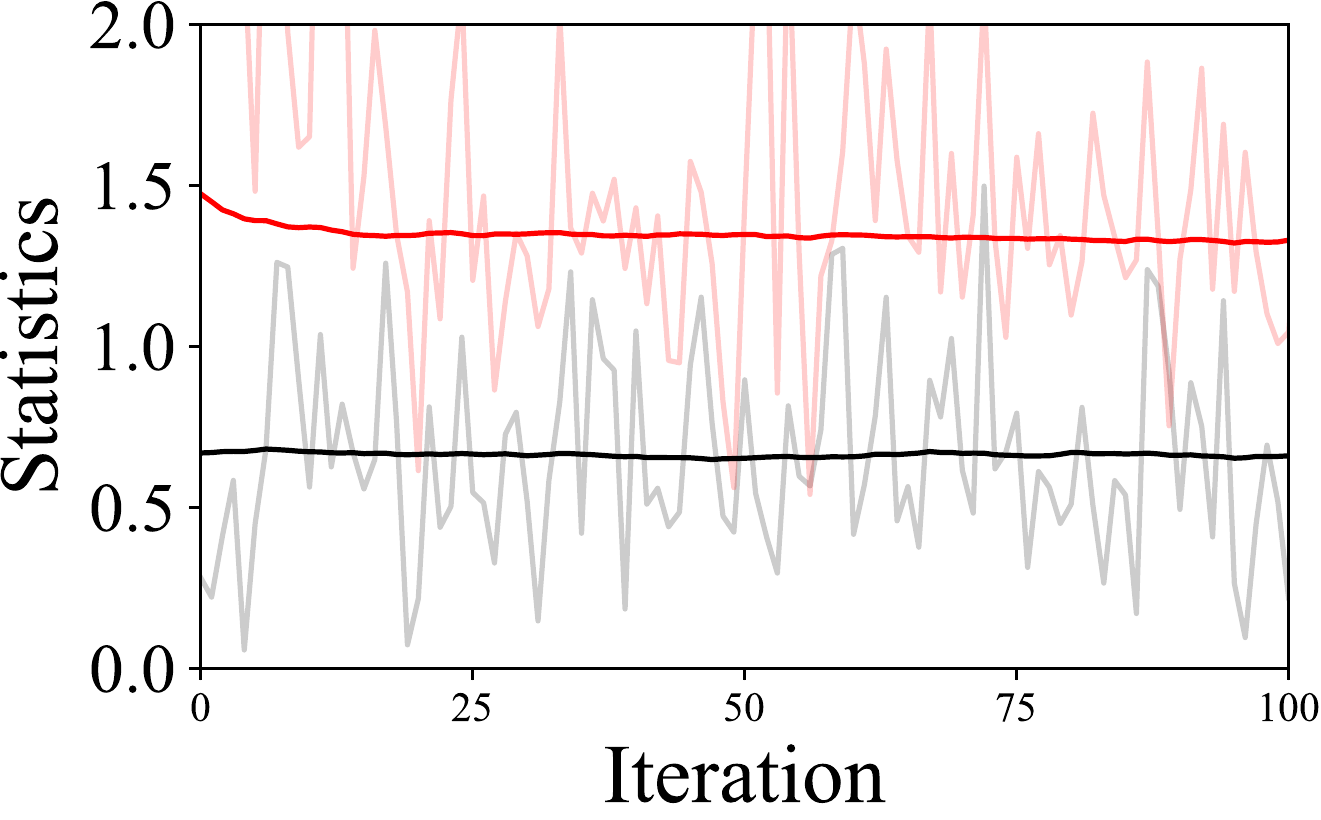}
  \end{subfigure}%
  \begin{subfigure}[t]{.25\linewidth}
    \centering
    \includegraphics[width=\linewidth]{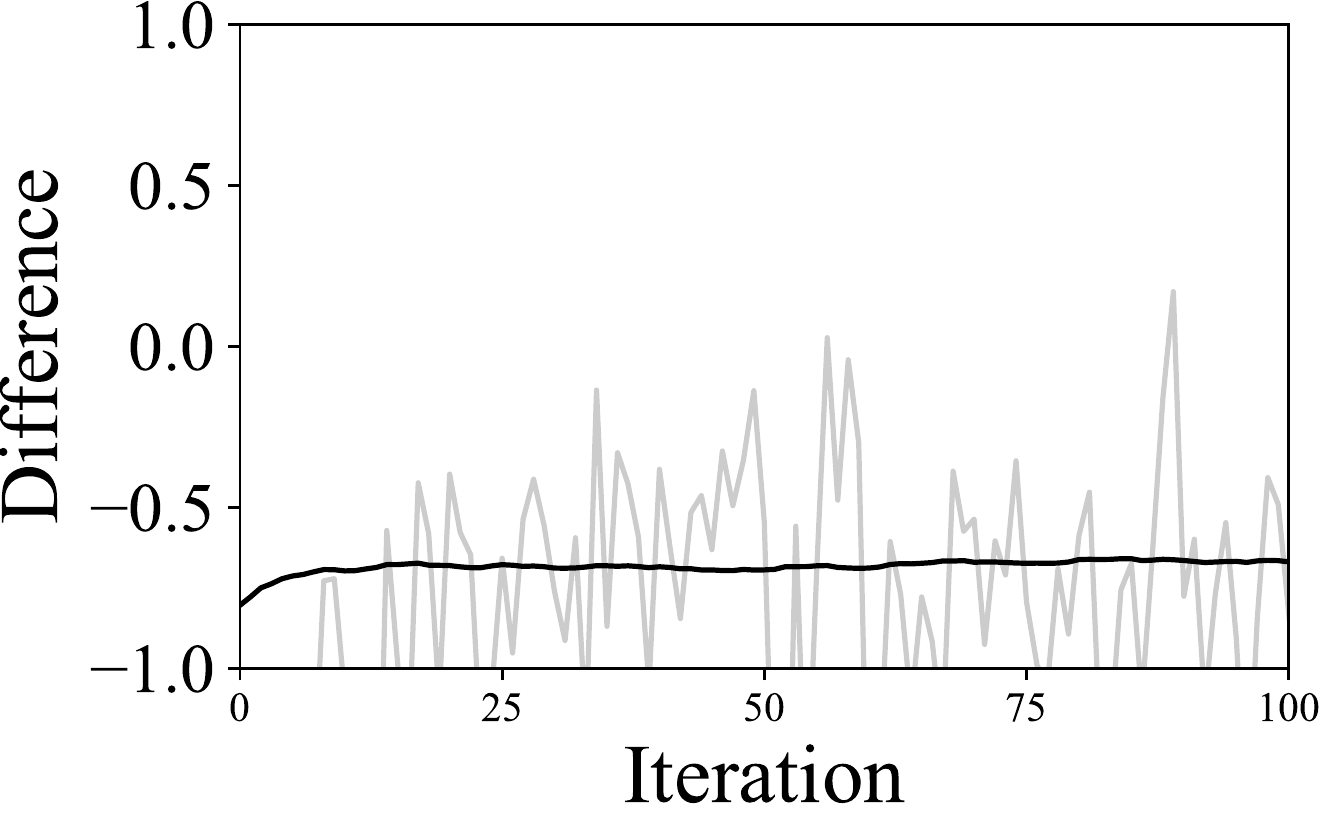}
  \end{subfigure}%
  \begin{subfigure}[t]{.25\linewidth}
    \centering
    \includegraphics[width=\linewidth]{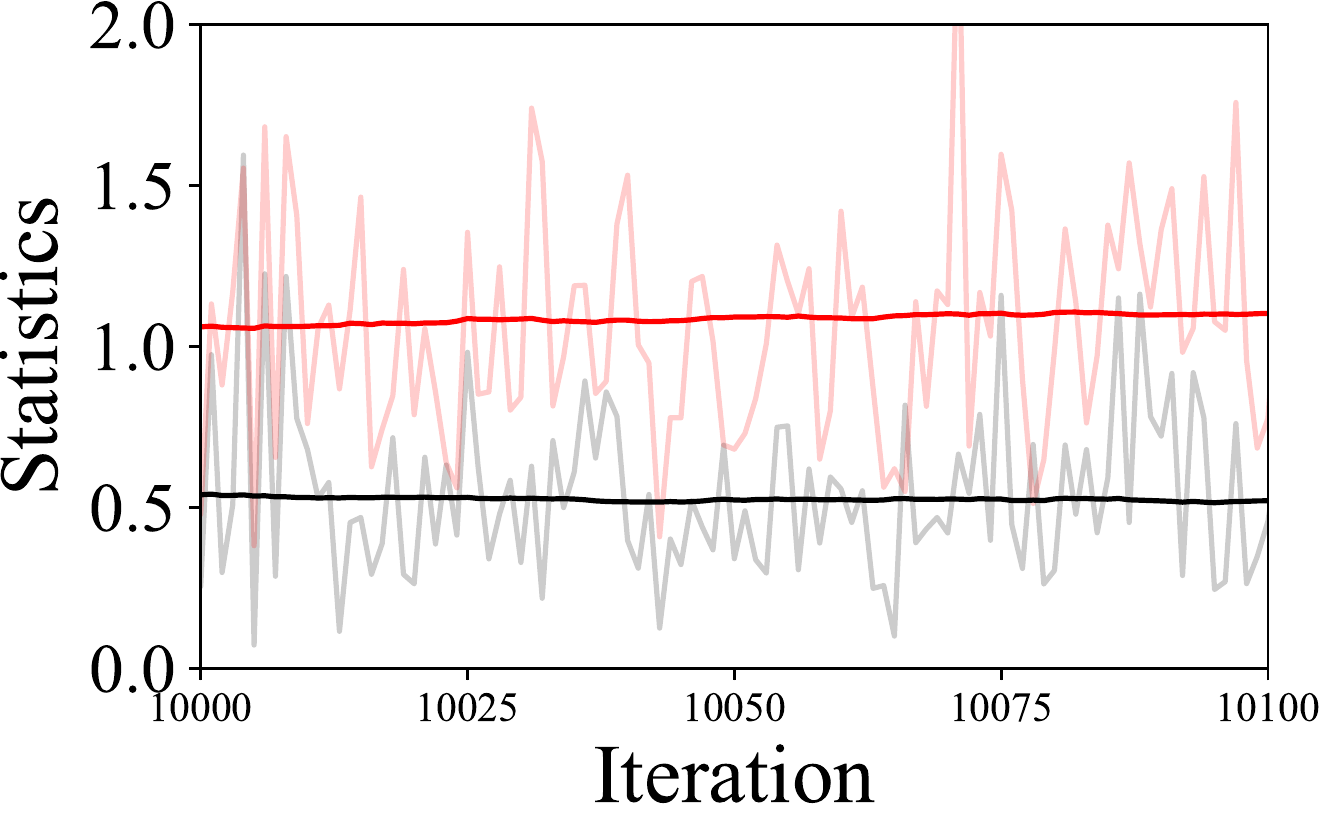}
  \end{subfigure}%
  \begin{subfigure}[t]{.25\linewidth}
    \centering
    \includegraphics[width=\linewidth]{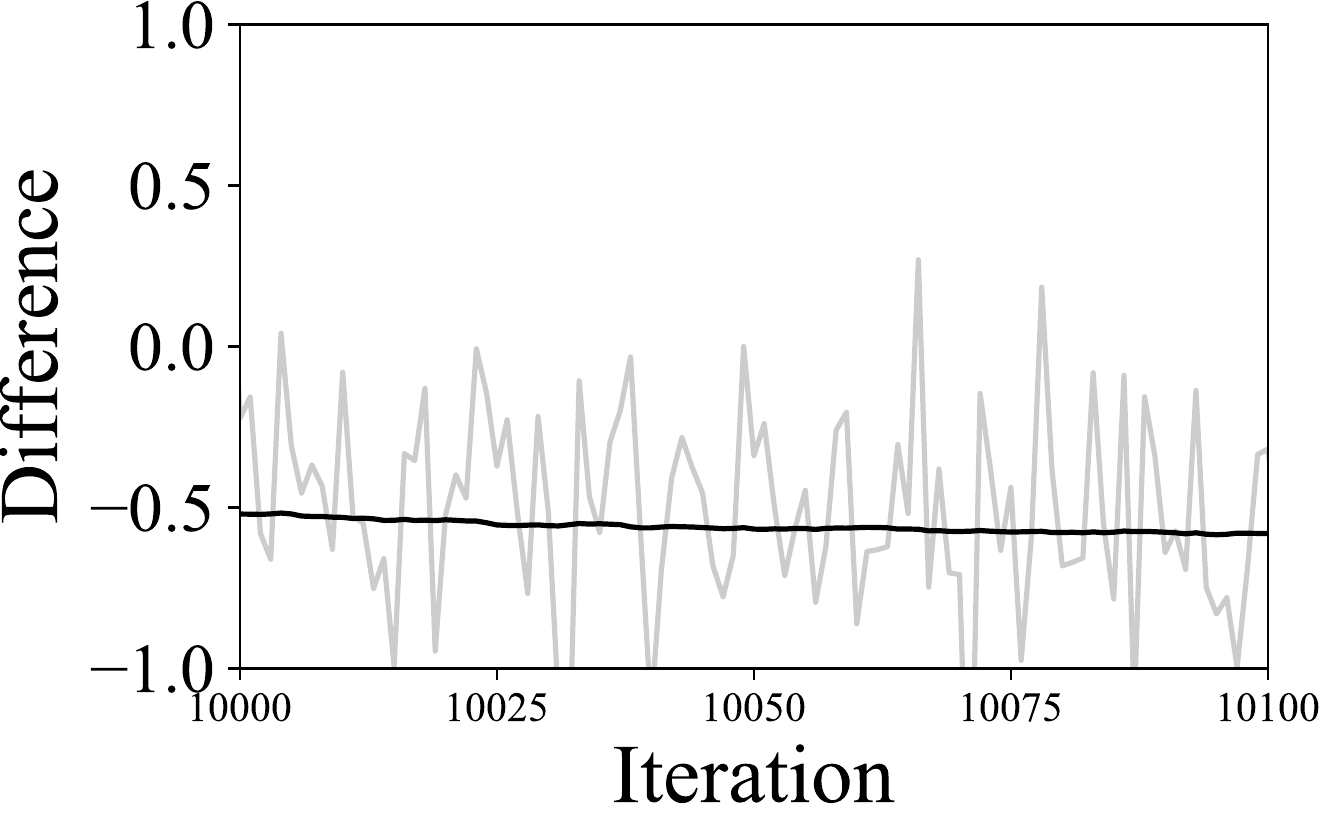}
  \end{subfigure}\\
  \begin{subfigure}[t]{.25\linewidth}
    \centering
    \includegraphics[width=\linewidth]{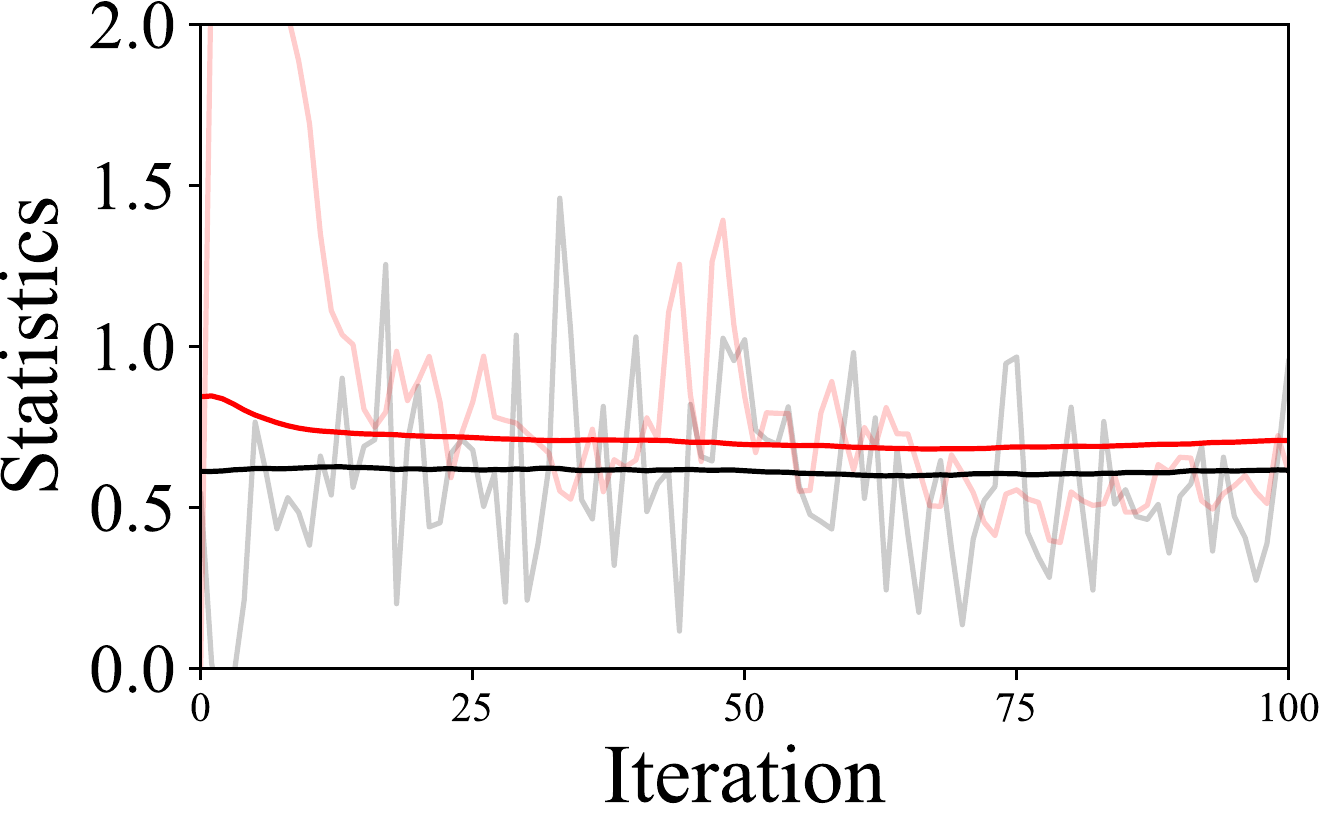}
  \end{subfigure}%
  \begin{subfigure}[t]{.25\linewidth}
    \centering
    \includegraphics[width=\linewidth]{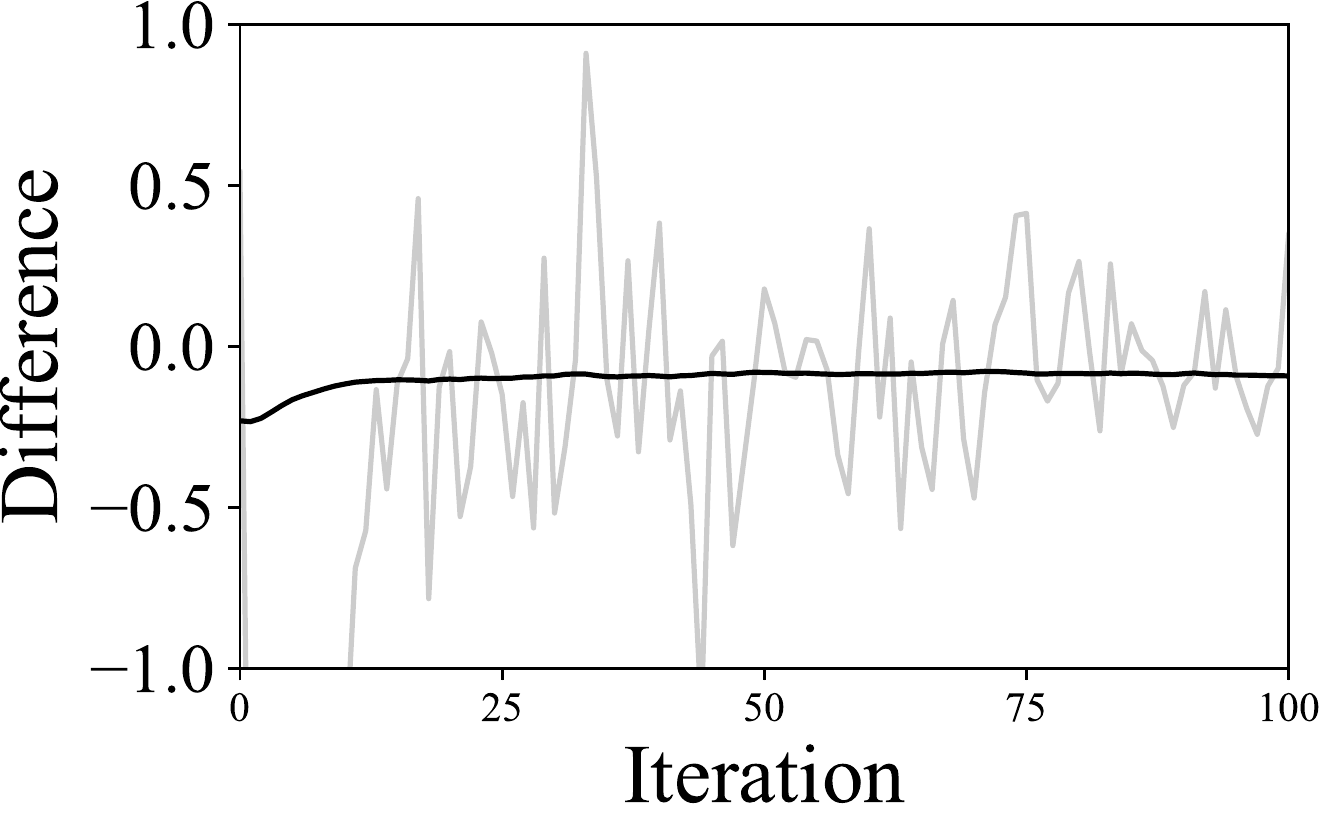}
  \end{subfigure}%
  \begin{subfigure}[t]{.25\linewidth}
    \centering
    \includegraphics[width=\linewidth]{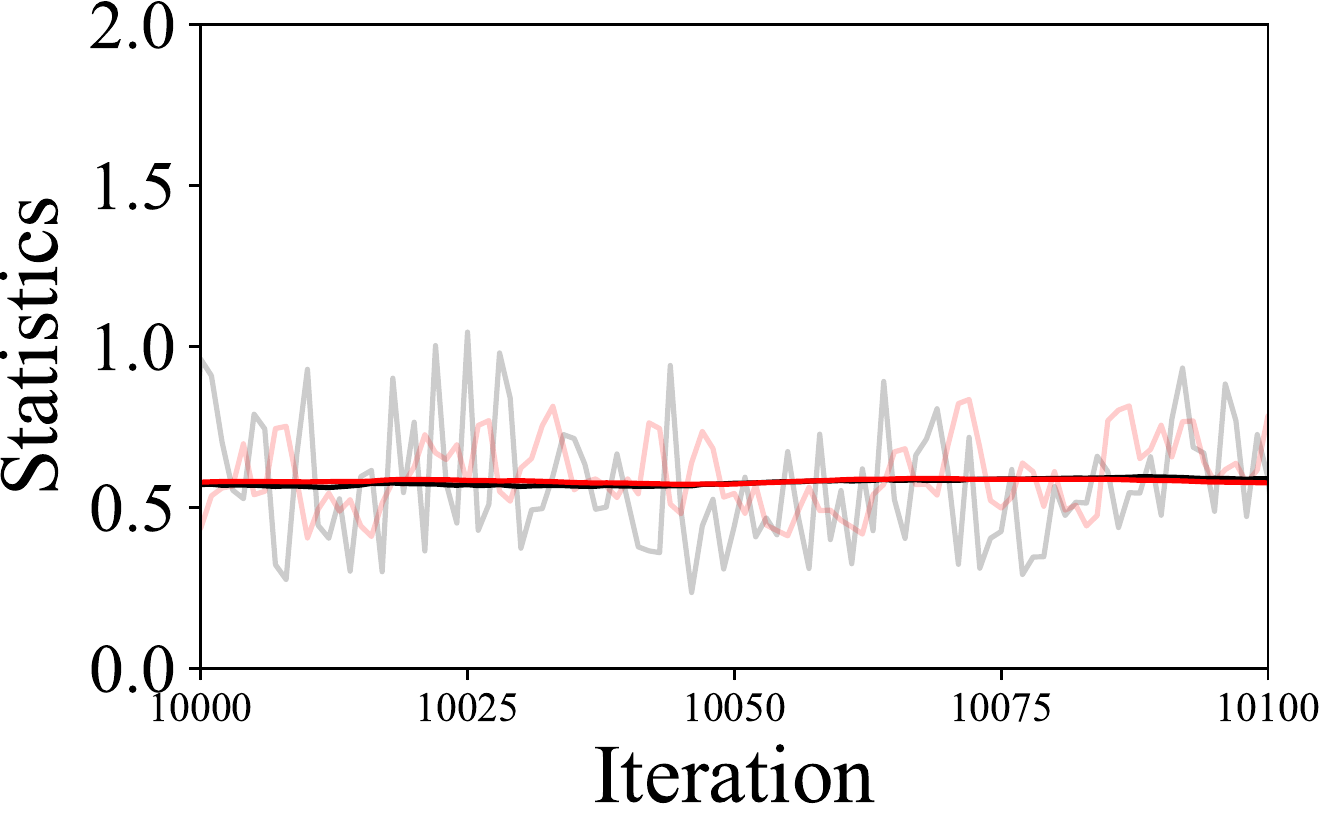}
  \end{subfigure}%
  \begin{subfigure}[t]{.25\linewidth}
    \centering
    \includegraphics[width=\linewidth]{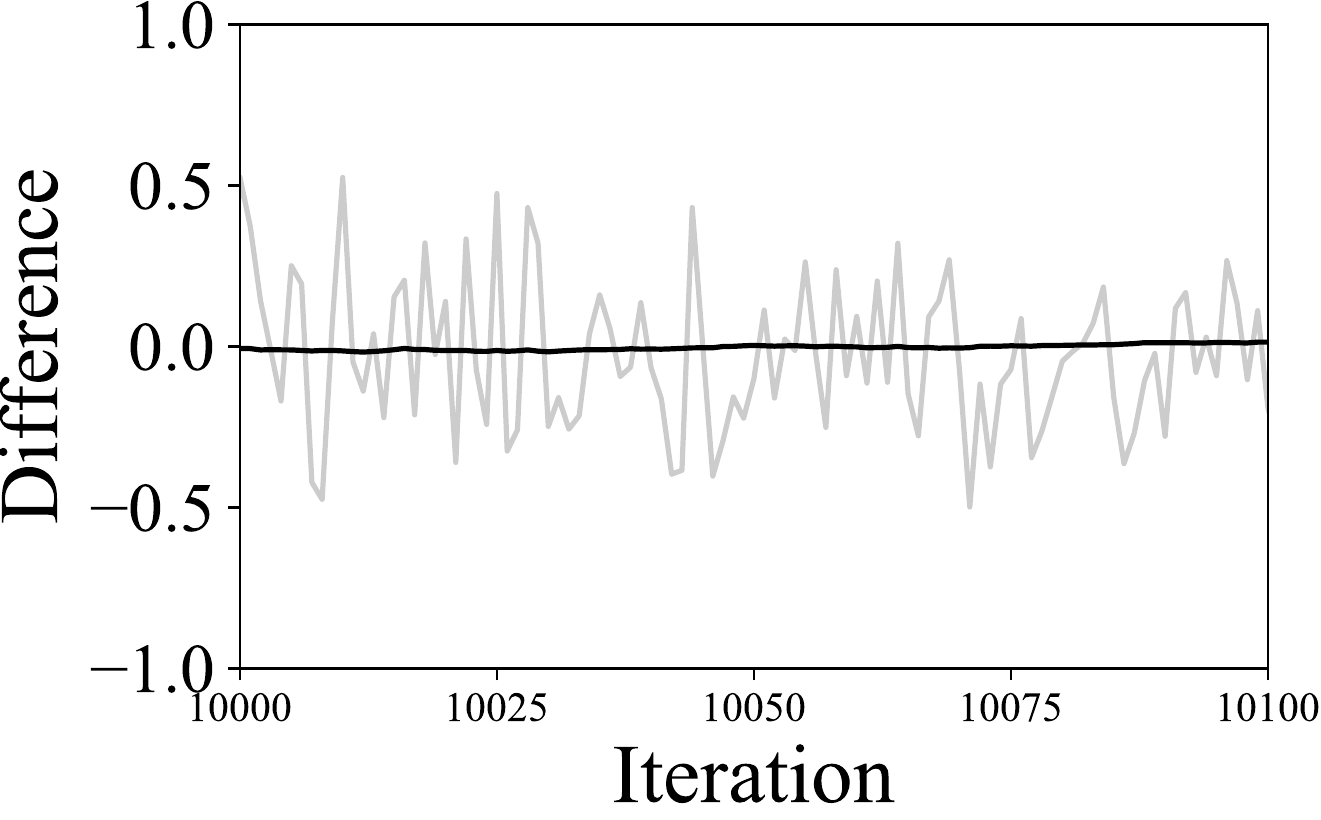} 
   \end{subfigure} 
  \caption{The two conditions \eqref{eqn:gkp-dk-correlation} (top) and \eqref{eqn:fdr1} (bottom) evaluated on a logistic regression model trained on MNIST, with $\alpha=1.0$. Left two columns: iteration 0-100; Right two columns: iteration 10000-10100. In the statistics plots, the red and black curves (dark) are the running estimates of the left-hand and right-hand side of each condition, respectively. The light curves show the raw value of each estimator at each iteration. Even when the number of iterations is very large, the statistics suggested by \eqref{eqn:gkp-dk-correlation} still do not match. On the other hand, the difference between the statistics in \eqref{eqn:fdr1} quickly converges to zero.}
  \label{fig:comparisonfig}
  \end{figure}

Figure \ref{fig:comparisonfig} evaluates the two stationary conditions \eqref{eqn:gkp-dk-correlation} and \eqref{eqn:fdr1} by training an L2-regularized logistic regression model on MNIST and logging the estimators for both sides of each relation. The top row shows that even when the number of iterations grows large, there is still non-negligible error in Pflug's condition even though the function is strongly convex. Contrastingly, the statistics in Yaida's relation, shown in the bottom row, quickly become almost indistinguishable, as predicted by \eqref{eqn:fdr1} and \eqref{eqn:mean-conv}. Together with the difficulty of estimating its right-hand-side, this inaccuracy makes the Pflug condition unattractive for quantitative applications such as ours, which require a precise relationship to hold. However, the \emph{qualitative} intuition given by such quadratic stationary formulae has proven useful \citep{MandtHoffmanBlei2017}.

\section{Additional SASA discussion}
\label{apdx:more-discussion}
\subsection{The missing step to derive the stationary condition \eqref{eqn:fdr1}}
\label{app:derivation}
Assuming the existence of a stationary condition $\pi(d,x,g)$ for the SGM dynamics~\eqref{eqn:sgm-fixed}, \cite{yaida2018fluctuation} showed 
\begin{equation}
\label{eqn:fdr1yaida}
\E_{\pi}[\langle x, \nabla F(x)\rangle ] = \frac{\alpha}{2}\frac{1 + \beta}{1-\beta}\E_{\pi}[ \langle d, d\rangle ].
\end{equation}
Using history average to estimate the left hand side needs the full gradient of $F$, which is not available (or expensive to compute) during training. Instead, both \cite{yaida2018fluctuation} and SASA use \eqref{eqn:fdr1} in practice, i.e.,
\begin{equation}
\label{eqn:fdr1app}
\E_{\pi}[\langle x, g \rangle ] = \frac{\alpha}{2}\frac{1 + \beta}{1-\beta}\E_{\pi}[ \langle d, d\rangle ],
\end{equation}
where the left hand side can be estimated with nearly no computational overhead. Here, we provide the missing step from \eqref{eqn:fdr1yaida} to \eqref{eqn:fdr1app}.

By the law of total probability, we have
$$\E_{\pi}[\langle x, g \rangle ] = \E_{\pi}\left[ \E_{\pi}[\langle x, g \rangle | x,d] \right] = \E_{\pi}\left[ \langle x, \E_{\pi}[ g | x,d]\rangle \right].$$
We denote the time-independent transition kernel from $(\xk,\dk,\gk)$ to $(\xkp,\dkp,\gkp)$ in \eqref{eqn:sgm-fixed} as $\mathbf{T}$. Then since $\pi$ is the stationary distribution, we have the pushforward measure of $\pi$ under $\mathbf{T}$ is still $\pi$, i.e., $\mathbf{T}^{\sharp} \pi = \pi$. Then we have
$$\E_{\pi}[ g | x,d] = \E_{\mathbf{T}^{\sharp} \pi}[ g | x,d] \stackrel{(*)}{=} \E_{\mathbf{T}^{\sharp} \pi}[ g(x) | x,d] = \nabla F(x),$$
where the definition of the transition kernel~\eqref{eqn:sgm-fixed} is used in the step (*) and the unbiasedness of the stochastic gradient, see Eqn.~\eqref{eqn:gmarkovian}, in the last step.

\subsection{Discussion on the multiple-test problem}
Although SASA performs sequential hypothesis testing, it does not seem to suffer from the issue of inflated false discovery rate~\cite{mcdonald2009handbook}. That is, we do not observe that the test fires earlier than it ``should'' in our numerical experiments. From Figure~\ref{fig:cifar_stats}, we can see that the statistic $\bar{z}_N$ is either monotonically decreasing to 0 or first decreasing and then increasing to 0, leading to high positive correlation among the tests. This high correlation may prevent proportional inflated false discovery rates; see, e.g., \cite{benjamini1995controlling,blanchard2009adaptive,lindquist2015zen}.

\end{document}